\documentclass[numbers,10pt]{arxiv}

\usepackage[utf8]{inputenc}

\usepackage{tabularx}
\usepackage{nicefrac}
\usepackage{graphicx, subcaption}
\usepackage{enumitem}
\usepackage{array, tabularx}
\newcolumntype{L}{>{\raggedright\arraybackslash}X}
\usepackage{fvextra}
\usepackage[most]{tcolorbox}
\usepackage{xspace}
\usepackage{bxcoloremoji}

\usepackage{graphicx}
\usepackage{enumitem}
\usepackage{caption}
\usepackage{xspace}
\usepackage{fvextra}
\usepackage{cleveref}
\usepackage{float} 
\usepackage{subcaption}
\usepackage[most]{tcolorbox}
\usepackage{booktabs}
\usepackage{adjustbox}
\usepackage{multirow}
\usepackage{siunitx}

\makeatletter
\DeclareRobustCommand\onedot{\futurelet\@let@token\@onedot}
\def\@onedot{\ifx\@let@token.\else.\null\fi\xspace}

\makeatother

\renewcommand{\paragraph}[1]{\textbf{#1}}

\usepackage{xcolor}
\usepackage{amsmath}
\usepackage{wrapfig}

\definecolor{promptblue}{RGB}{25, 84, 166}
\definecolor{thinkgreen}{RGB}{34, 139, 34}
\definecolor{thinktag}{RGB}{110, 65, 30}
\definecolor{thinkbg}{RGB}{252, 248, 243}
\definecolor{boxbg}{RGB}{252, 252, 252}

\tcbset{
    mainbox/.style={
        colback=boxbg,
        colframe=gray!40,
        arc=2mm,
        boxrule=0.8pt,
        left=12pt,
        right=12pt,
        top=12pt,
        bottom=12pt,
        fonttitle=\bfseries\large,
        coltitle=white,
        colbacktitle=gray!60
    }
}

\crefname{appendix}{Appendix}{Appendices}
\Crefname{appendix}{Appendix}{Appendices}

\usepackage[T1]{fontenc}
\usepackage{amsfonts}
\usepackage{amssymb}
\usepackage{amsthm}
\usepackage{mathtools}
\usepackage{longtable}
\usepackage{rotating}
\usepackage{xfrac}
\usepackage{url}
\usepackage{colortbl}
\usepackage{pifont}
\usepackage{fontawesome5}
\usepackage{etoc}

\renewcommand{\paragraph}[1]{\sansbf{#1}}

\newcommand{\tokcount}{400B\xspace}

\newcommand{\names}{LCLMs\xspace}
\newcommand{\name}{LCLM\xspace}
\newcommand{\fullnames}{Latent Context Language Models\xspace}
\newcommand{\fullname}{Latent Context Language Model\xspace}

\newcommand{\nyu}{\ensuremath{\textcolor[HTML]{57068C}{\spadesuit}}}
\newcommand{\umd}{\mbox{\textcolor[HTML]{E21833}{\scalebox{0.6}{\faHeart}}}}
\newcommand{\princeton}{\ensuremath{\textcolor[HTML]{E77500}{\clubsuit}}}
\newcommand{\columbia}{\ensuremath{\textcolor[HTML]{009EFF}{\blacklozenge}}}
\newcommand{\modal}{\ensuremath{\textcolor[HTML]{00A86B}{\scalebox{1.0}{$\blacktriangle$}}}}
\newcommand{\meta}{\ensuremath{\textcolor[HTML]{0064E0}{\scalebox{0.7}{$\blacksquare$}}}}
\newcommand{\llnl}{\mbox{\textcolor[HTML]{003C71}{\raisebox{-0.15ex}{\scalebox{1.25}{$\bullet$}}}}}
\newcommand{\harvard}{\ensuremath{\textcolor[HTML]{A51C30}{\bigstar}}}

\title{End-to-End Context Compression at Scale}

\author[\nyu\,\modal\,\star]{Ang Li}
\author[\umd\,\star]{Sean McLeish}
\author[\princeton\,\star]{Haozhe Chen}
\author[\columbia]{Nimit Kalra}
\authorbigbreak
\author[\columbia]{Zaiqian Chen}
\author[\harvard]{Artem Gazizov}
\author[\columbia]{Venkata Anoop Suhas Kumar Morisetty}
\authorbreak
\author[\llnl]{Bhavya Kailkhura}
\author[\llnl]{Harshitha Menon}
\author[\princeton]{Zhuang Liu}
\author[\llnl]{Brian R. Bartoldson}
\authorbigbreak
\author[\umd]{Tom Goldstein}
\author[\meta\,1]{Sanae Lotfi}
\author[\columbia\,\ddagger]{Micah Goldblum}
\author[\nyu\,\ddagger]{Pavel Izmailov}

\contribution[\ensuremath{\star}]{Equal contribution, correspondence to \texttt{al6843@nyu.edu, smcleish@umd.edu, hc5019@princeton.edu}}
\contribution[\ensuremath{\ddagger}]{Equal advising, correspondence to \texttt{micah.g@columbia.edu, pi390@nyu.edu}}

\affiliation[\nyu]{\textcolor[HTML]{57068C}{New York University}}
\affiliation[\modal]{\textcolor[HTML]{00A86B}{Modal Labs}}
\affiliation[\umd]{\textcolor[HTML]{E21833}{University of Maryland}}
\affiliation[\princeton]{\textcolor[HTML]{E77500}{Princeton University}}
\affiliation[\columbia]{\textcolor[HTML]{009EFF}{Columbia University}}
\affiliation[\harvard]{\textcolor[HTML]{A51C30}{Harvard University}}
\affiliation[\llnl]{\textcolor[HTML]{003C71}{Lawrence Livermore National Laboratory}}
\affiliation[\meta]{\textcolor[HTML]{0064E0}{FAIR at Meta}}

\abstract{
Long-context language model inference is bottlenecked by memory, as the KV cache grows with context length. Recent techniques to compress the KV cache fall short: they either degrade model quality substantially or require considerable time and compute to compress a single long prompt. Furthermore, many methods require the input to fit within the target model's context window, and are generally incompatible with modern production inference engines. Encoder-decoder compressors, which map a long token sequence to a shorter sequence of latent embeddings consumed by a decoder, are an appealing alternative in principle. However, existing approaches are not competitive with KV cache compression on the accuracy-efficiency frontier. In this work, we revisit encoder-decoder compression and close this gap. We first perform an architecture search, pre-training many variants from scratch to determine how best to design and train encoder-decoder compressors. Guided by our findings, we continually pre-train a family of 0.6B-encoder, 4B-decoder models on nearly \tokcount tokens each, at compression ratios of 1:4, 1:8, and 1:16. We introduce \textit{Latent Context Language Models} (LCLMs), a family of compressors that improve the Pareto frontier across general-task performance, compression speed, and peak memory usage. LCLMs also unlock new efficient downstream applications: they serve as backbones for long-horizon agents that skim a compressed long context and adaptively expand relevant segments on demand, and they can compact the model's own generated reasoning into latent memory during inference, which is nearly lossless on the full MATH test set while being highly efficient, reducing peak KV usage by up to \(7.9\times\).

\coloremojicode{1F917} \textbf{Models}: \href{https://huggingface.co/latent-context}{huggingface.co/latent-context} \\
\faGithub\ \textbf{Code}: \href{https://github.com/LeonLixyz/LCLM}{github.com/LeonLixyz/LCLM}
}

\begin{document}
\etocdepthtag.toc{mtmain}
\maketitle

\blankfootnote{$^{1}$Served solely in an advisory role for this paper. All model training and data processing were conducted entirely outside of Meta.}
\setcounter{footnote}{0}

\section{Introduction}\label{sec:intro}
Reasoning over long contexts is a crucial capability for state-of-the-art Large Language Models (LLMs), as it enables them to parse long documents, engage in multi-turn conversations, and perform long-horizon agentic tasks. However, in production systems, the input context, working horizon, and memory can grow to millions of tokens, making inference increasingly constrained by memory and latency due to the growth of the KV cache~\citep{hooper2024kvquant}.
Even when inputs fit within the model's maximum context window, models often struggle to reliably use information distributed across long contexts~\citep{liu2024lost,an2025why}. As a result, context and memory management have emerged as critical systems and modeling challenges for deploying long-horizon LLM systems.

A natural response is to reduce the size of the KV cache directly.
KV cache compression methods reduce the memory footprint of cached activations by evicting cache entries \citep{li2024snapkv,devoto2025expected,kim2025kvzip,zweiger2026fast} or training a compact cache offline \citep{cartridges}.
However, they have important limitations for general long-context inference.
Many methods still require the full context to be prefilled before compression, and some require substantial additional time or memory beyond the original prefill.
Query-dependent methods produce caches that are difficult to reuse across turns \citep{li2024scbench}, while methods that evict caches non-uniformly across attention heads and layers \citep{kim2025kvzip} are difficult to integrate with modern inference engines that assume a shared sequence length across the cache.

\begin{figure*}[t]
  \centering
  \includegraphics[width=\textwidth]{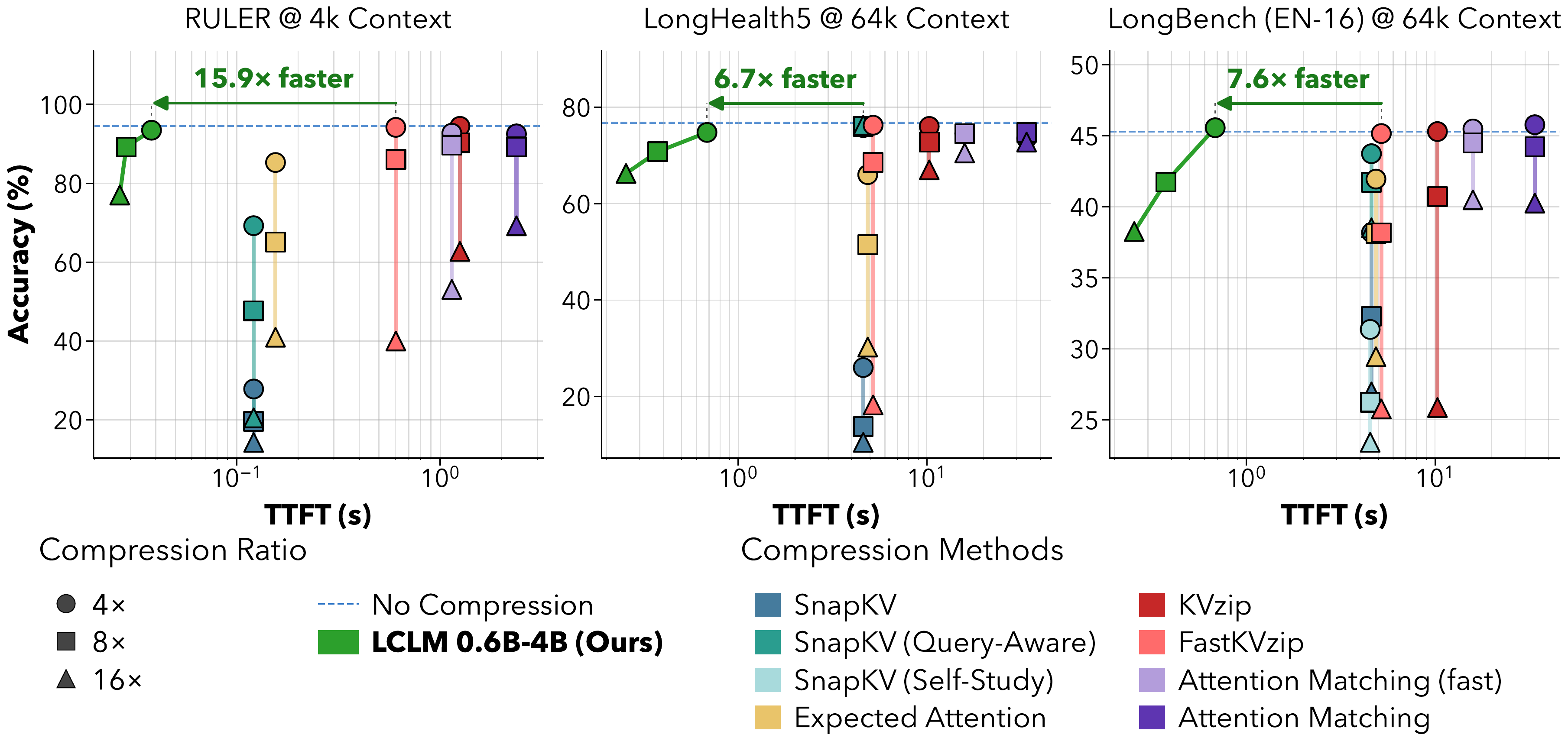}
  \caption{\textbf{Our \fullnames achieve high quality compression while being fast and memory efficient.}
  We show accuracy vs.\ TTFT (time to first token) per sample on a single H200 for RULER~\citep{hsieh2024ruler} at 4k tokens, and for LongHealth5~\citep{adams2025longhealth} and LongBench~\citep{bai2024longbench} at 64k tokens. All methods are based on the same decoder \texttt{Qwen3-4B-Instruct-2507}~\citep{qwen3}. We see that our models lie on a new Pareto frontier, compressing contexts much faster while maintaining high accuracy, especially for high compression ratios.
  KV cache compression baselines appear as vertical lines because the context is prefilled the same way regardless of compression ratio, and the eviction operation is much faster than the prefill time. Arrows mark the speedup of the \name at \(4\times\) compression over the fastest KV cache method at \(4\times\) compression with accuracy within 1 point of it, so the matched baseline, and hence the speedup, can differ between benchmarks of the same length.
}
  \label{fig:cover}
\end{figure*}

Encoder-decoder soft-token methods~\citep[e.g.][]{refrag,in-context-autoencoder,auto-compressor,e2llm,cepe} offer a promising way to address the limitations of KV cache compression. Rather than manipulating the KV cache directly, these models encode the raw input tokens into a much shorter sequence of continuous embeddings, which are then provided to the LLM 
in place of the original context.
In principle, this approach is highly attractive: compression is parallelizable, supported by standard LLM inference engines, and 
works with standard LLM decoding while extending the decoder far beyond its native context length.

However, existing soft-token compression methods typically face a trade-off: they either substantially degrade the capabilities of the base language model or depend on domain-specific training to remain effective.
This motivates a fundamental question: \textbf{can we train a general, task-agnostic compressor that preserves a model's original capabilities?} To this end, we introduce \textit{\fullnames} (\names), a family of encoder-decoder compressors trained end-to-end at scale.
We initialize both the encoder and decoder with pre-trained models and jointly train them on a total of nearly \tokcount tokens for the combined encoder-decoder. 
Crucially, we demonstrate that \names can preserve decoder performance while reducing memory used for input context.

Our contributions are summarized as follows: 
\begin{enumerate}
\item \textbf{Training recipe:} We develop a fine-grained training recipe for encoder-decoder context compression. Accordingly, we curate continual pre-training and finetuning data for training compressors and optimize training details across multiple training stages. See \Cref{sec:training}.

\item \textbf{Comprehensive architectural search:} We conduct a large-scale architecture search to identify optimal design choices for encoder-decoder compression models in a controlled setting. See \Cref{sec:arch} and \Cref{fig:arch_sweep}.

\item \textbf{New memory/speed-performance frontier:} We train a family of models across different compression ratios at scale. Our experiments identify a Pareto frontier that effectively balances memory, latency, and downstream accuracy. We open-source this suite of models and all code. See \Cref{sec:results} and \Cref{fig:cover,fig:pareto_optimal,fig:lm_pareto}.

\item \textbf{New capabilities from compression:} We create an agentic system with a high compression ratio, where the agent can select which compressed chunk to expand, substantially enhancing the model's performance on challenging needle-in-a-haystack tasks. We also use the encoder to compress the model's own generated tokens during decoding, so long chains of thought are kept in a compact latent memory; on MATH, this is nearly lossless while reducing KV usage several-fold. See \Cref{sec:capabilities} and \Cref{fig:agentic_bar_plot,fig:online_compaction}.
\end{enumerate}

\section{Related Work}\label{sec:related}
Context compression approaches generally fall into three categories: hard-token, soft-token, and KV cache compression.
We defer the discussion of hard-token approaches to \Cref{app-sec:extended-rel-work}, since they generally underperform compared to the latter two.

\textbf{KV cache compression.}
KV cache compression methods evict entries from the KV cache. Most methods rely on hand-crafted or learned heuristics to select which entries to drop, while others train the model to merge or sparsify its own cache \citep{pmlr-v235-nawrot24a,lancucki2026inference}.

Prompt-agnostic methods \citep{kim2025kvzip,devoto2025expected,zweiger2026fast} prune the context without knowledge of the query, whereas prompt-dependent methods such as SnapKV \citep{li2024snapkv} require explicit context-prompt pairs and yield query-specific caches that limit multi-turn applicability \citep{li2024scbench,kim2025kvzip}.
An alternative is to anticipate future queries at compression time: Cartridges \citep{cartridges} distills a fixed-size KV cache per corpus.
However, this comes at a substantial training cost per corpus, with an ICL-quality cache requiring approximately 30 minutes on an 8$\times$H100 node for an 8B model.

Although KV cache compression methods are well-studied, they are not widely adopted in inference engines such as vLLM \citep{kwon2023efficient} or SGLang \citep{zheng2024sglang}. Compression methods such as KVzip allocate eviction budgets non-uniformly across attention heads and layers; therefore, they cannot reduce the sequence-length dimension of the KV cache. Instead, KV cache compression methods mask evicted positions in the attention computation, forfeiting the memory and throughput benefits of these paged-attention engines. Some methods further require prefill passes that are two to three times longer than the original input \citep{kim2025kvzip}.
We note that soft token compression methods are fully compatible with these open-source inference frameworks.

\textbf{Soft-token compression.}
Despite promising results, most prior soft-token methods rely on offline preprocessing and do not convincingly preserve the base model's broad in-context behavior.
Prior work is mostly evaluated on domain- or task-specific finetuning that is not transferable across tasks \citep{GMSA, mean-pooling, dai2025pretraining, refrag, xrag}.
To the best of our knowledge, these methods are not evaluated on long-context benchmarks with information-dense tasks that require fine-grained details throughout the whole context, such as RULER \citep{hsieh2024ruler}, to demonstrate complex long-context understanding. 

In \Cref{sec:arch}, we discuss the architectural decisions of prior work on soft-token compression as part of our architectural search and analysis.
Our work shows that, with large-scale staged training, a single online soft-token compressor can handle heterogeneous inputs while closely matching uncompressed behavior.

\section{The \fullname Architecture}
An \name consists of an encoder with a pooling method that maps token chunks to soft tokens, an adapter to project to the decoder's embedding dimension, and a decoder that consumes the soft tokens as latent context. Given a sequence of input tokens $x_{1:T}$ and a compression ratio $N$, the encoder maps each contiguous block of $N$ input tokens to a single latent token. 
Let \(W\) denote the encoder window size, i.e., the number of input tokens processed by the encoder in one forward pass.
We split the input sequence into $I=\lceil T/W\rceil$ encoder windows, each containing at most $W$ tokens:
\begin{equation}
w_i
=
x_{
    (i-1)W+1
    \; : \;
    \min(iW,T)
},
\qquad
i = 1,\dots,I.
\end{equation}
For each window $w_i$, the encoder computes the final hidden states $h^{(i)}_{1:|w_i|}$ for each input token in the window.
A pooling operator then aggregates these hidden states into $M_i = \left\lceil \frac{|w_i|}{N} \right\rceil$ latent tokens $z^{(i)}_{1:M_i}$. When $W=N$, each encoder window contains a single compression block, yielding one latent token per forward pass. When $W=T$, the encoder processes the entire input sequence in one forward pass and produces $\lceil T/N\rceil$ latent tokens. We concatenate the latent tokens from all encoder windows to obtain the full compressed latent sequence $z_{1:M}$, where $M$ is the total number of latent tokens after compression.

Because the encoder and decoder may have different hidden dimensions, an adapter $a(\cdot)$ projects the compressed latent sequence $z_{1:M}$ from the encoder hidden dimension into the decoder hidden dimension, producing latent tokens $s_{1:M}$. The decoder then consumes these projected latent tokens in place of the corresponding input tokens. 

\section{Training Recipe}\label{sec:training}

We now describe our multi-stage recipe and data for training.
In contrast to prior context-compression work that trains specialized models on small-scale in-domain datasets~\citep{GMSA,mean-pooling,arc-encoder,xrag,e2llm}, our goal is to preserve the strong performance of a powerful LLM across downstream tasks. To this end, we curate three types of data: continual pre-training data, Supervised Fine-Tuning (SFT) data, and auxiliary reconstruction data.

\begin{figure*}[t]
  \centering
  \includegraphics[width=1.0\textwidth]{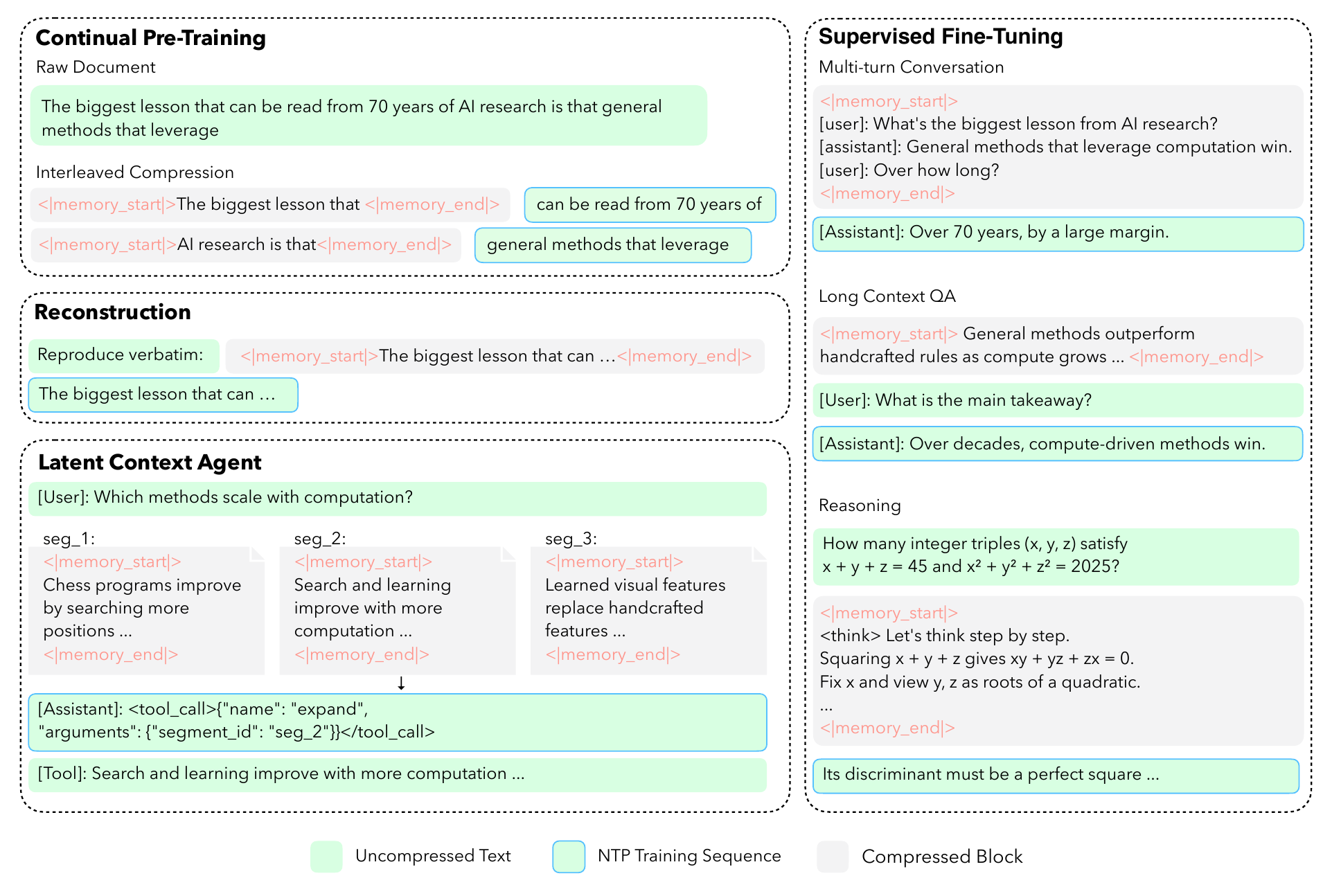}
    \caption{\textbf{Examples of the data used to train \names.} Continual pre-training interleaves compressed and uncompressed spans, and reconstruction recovers source text from latents. SFT includes conversations, long-context QA, reasoning, and agentic trajectories. Reasoning questions remain raw while selected spans of the reference chain of thought are compressed; the remaining continuation is supervised. Native tool trajectories remain raw, while expansion examples compress source observations and retain raw tool calls and returned text. Blue outlines mark supervised decoder tokens.}
    \label{fig:data_example}
\end{figure*}

\subsection{Training Data}

\paragraph{Continual pre-training dataset.}
We construct a carefully curated high-quality pre-training mixture covering Common Crawl text, code, science and reasoning, long-context data, and instruction-style data.
We partition each sequence into multiple segments and alternate between compressed segments and standard token segments.
We then compute the next-token prediction loss on the uncompressed tokens \textit{only}. This interleaved format differs from the simple first-half compressed, second-half trainable split used in prior work~\citep{refrag,in-context-autoencoder,auto-compressor,cepe}. By distributing compressed spans throughout the sequence, the model learns to condition on latent context at multiple positions rather than only at the beginning of the context. We provide a visual example of this data format in \Cref{fig:data_example}. The exact pre-training mixture is described in \Cref{app-subsec:cont-pre-train-data}, with additional details on compression formatting in \Cref{compression-data-details}.

\paragraph{SFT data.}
To recover performance after continual pre-training~\citep{gao2025train} and teach the model to use latent context, we post-train with supervised fine-tuning on compressed inputs. Our SFT mixture spans four components: (i) reasoning, (ii) long-context instruction following, (iii) general instruction following and multi-turn conversation, and (iv) agentic trajectories. For long-context and instruction data, we compress the documents and prompts. For reasoning, we instead keep the question uncompressed and compress spans of the reference chain of thought, so the model learns to continue reasoning from compressed spans. To preserve tool use under compressed inputs, we also include public tool-calling, search, and terminal trajectories~\citep{nvidia_nemotron_nano_v3_2025,raoof2026openthoughts}, kept uncompressed, and 0.15M synthetic trajectories (under 1\% of the mixture) in which the model calls an \texttt{expand} tool to read back compressed legal and contract documents; these teach the tool interface of \Cref{sec:agentic} but share no data with its RULER tasks. Because some original completions were generated by outdated models, we relabel a subset of the SFT data using \texttt{Qwen3-30B-A3B-Instruct-2507} and \texttt{Qwen3-235B-A22B-Instruct-2507}, released under the Apache-2.0 license. The exact mixture of 21.5M examples is reported in \Cref{app-subsec:posttrain-data}. 

\paragraph{Auxiliary reconstruction data.}
To accelerate training and encourage the latent representations to preserve fine-grained information, we introduce an auxiliary reconstruction task. Given a sampled document, we compress the text and ask the model to repeat the original content. To increase diversity, we include a variety of sources spanning code, text, long documents, math, and \LaTeX, as described in \Cref{app-subsec:aux-recon-data}. To reduce prompt overfitting, we use a bank of \(100\) prompt templates per source. This auxiliary task encourages the compressor not only to support semantic understanding but also to retain fine-grained details needed for tasks such as exact retrieval. We include auxiliary reconstruction data in both continual pre-training and SFT.

\subsection{Training Stages}

We adopt a multi-stage training recipe to improve stability, preserve the original model's capabilities, and mitigate catastrophic forgetting. The stages progressively increase the number of trainable components, starting with adapter and encoder warm up, followed by end-to-end continual pre-training and SFT.

\begin{itemize}[leftmargin=3em, itemsep=0pt, topsep=0pt]
    \item \textbf{Stage 0: Adapter warmup.} We freeze both the encoder and decoder and update only the adapter.
    \item \textbf{Stage 1: Encoder training.} We unfreeze the encoder while keeping the decoder frozen.
    \item \textbf{Stage 2: End-to-end continual pre-training.} We unfreeze the decoder with a small learning rate.
    \item \textbf{Stage 3: Supervised fine-tuning.} We train on the SFT mixture and increase the decoder learning rate.
\end{itemize}

This staged procedure is analogous to Vision-Language Model (VLM) training pipelines that first align pre-trained representations across modalities~\citep{llava, tong2024cambrian} and then fine-tune for downstream tasks: stages 0 and 1 play the role of projector alignment, stage 2 corresponds to interleaved continual pre-training, and stage 3 to instruction tuning. For stages 0, 1, and 2, we uniformly sample from the continual pre-training mixture; for stage 3, we train on the SFT mixture. In early experiments, we consider directly training the full model end-to-end from the beginning, but find that this approach underperforms the multi-stage recipe. At the beginning of training, the decoder is not accustomed to taking in the output of the embedding model, so both encoder and decoder parameter gradients can be large, leading to model degradation.  The staged approach, where we first train the adapter before unfreezing the encoder and decoder, keeps gradients smooth and training more stable, preventing this degradation.  We therefore adopt the multi-stage recipe for all scaled experiments. We also consider parameter-efficient variants that keep the decoder fully frozen \citep{in-context-autoencoder, xrag, cepe} or use LoRA \citep{e2llm, mean-pooling}, but we find them to substantially underperform compared to full-parameter training in early experiments; we therefore focus on end-to-end training. With compression disabled, the trained decoders show minimal drift from the original model (\Cref{app-subsec:retention}). Details of the continual pre-training and SFT data mixtures are provided in \Cref{app-tab:data-mixture-pre-training} and \Cref{app-tab:SFT-mixture}, respectively. All other training details, including token budgets, learning rates, and sequence lengths, are provided in \Cref{tab:training_recipe}.

\section{Architectural Design Space of Latent Encoders}\label{sec:arch}

We conduct a controlled architecture search over encoder-decoder context compressors by pre-training \name variants from scratch. Our cleanroom setting lets us isolate the effect of architectural choices without confounding from pre-trained initialization. This is particularly important for choices such as the encoder attention mask and pooling operators: for example, the pre-trained \texttt{Qwen3-Embedding-0.6B} encoder uses causal attention and EOS pooling, which could bias the architectural search. Specifically, we use \texttt{Qwen3-0.6B} as the base architecture for both the encoder and decoder, randomly initializing all weights with the same seed, and training all components end-to-end with no frozen parameters. Each variant is trained on the pre-training mixture described in \Cref{sec:training} for 38B tokens at a compression ratio $N=16$. This comprehensive search covers nearly all major architectural choices explored in prior soft-token compression work, along with additional underexplored choices. Prior soft-token compressors can be decomposed into a small set of design choices: the pooling operator, encoding granularity, encoder attention mask, adapter, encoder initialization, and decoder adaptation. Each prior method is a point in this design space (\Cref{app-tab:soft-token} in \Cref{app-subsec:soft-token}), so sweeping each axis under matched data, scale, and decoder compares the choices behind prior methods without the confounding differences between their original setups. We describe each design axis and its empirical effect below.

\begin{figure*}[t]
  \centering
  \includegraphics[width=0.95\textwidth]{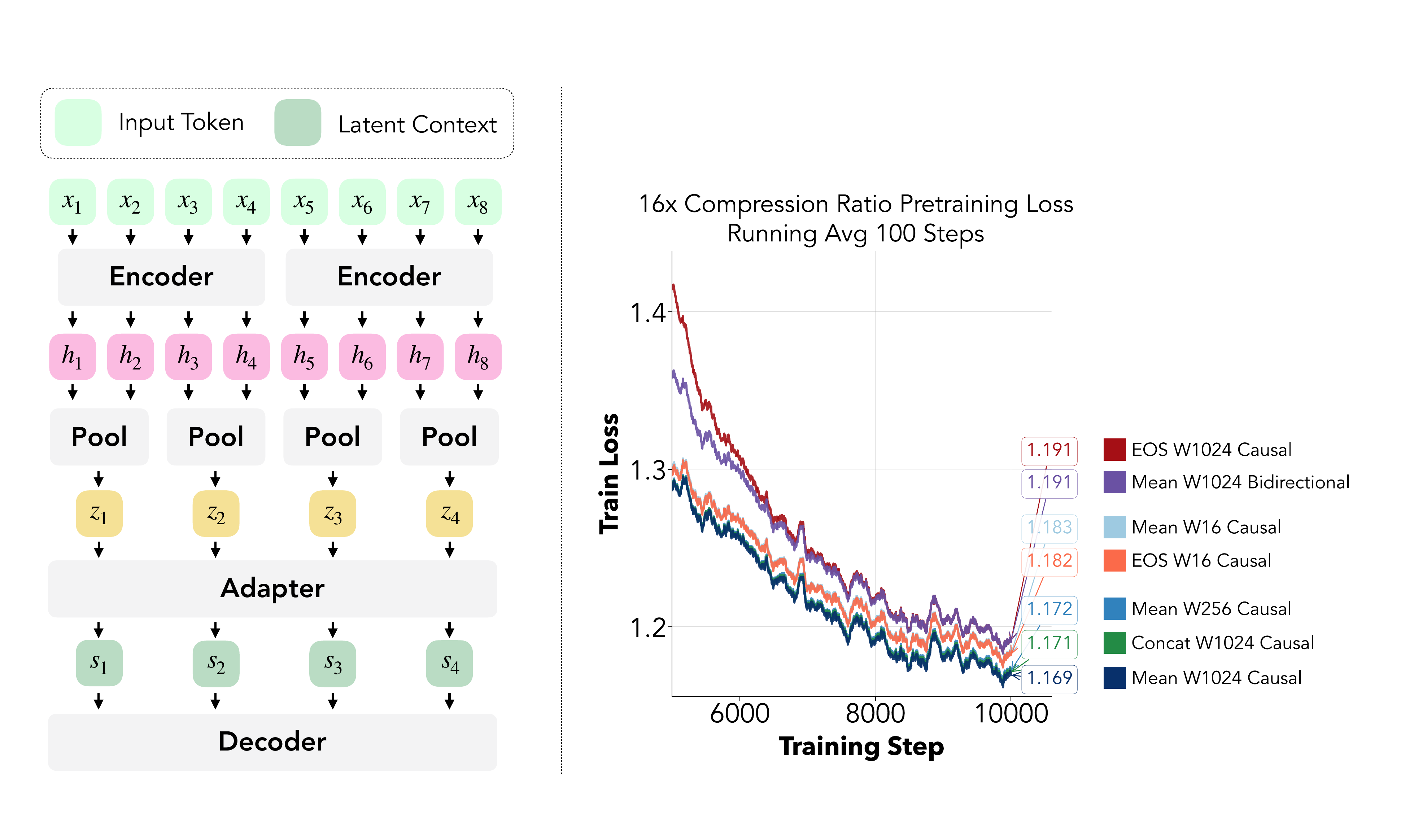}  \caption{\textbf{A from-scratch pre-training sweep identifies the best encoder-decoder compressor architecture.} \textbf{Left:} We explore how to best train an encoder LM to compress raw text into latent vectors and a decoder LM that uses the vectors as latent context. \textbf{Right:} We sweep many variants of encoder-decoder compressors end-to-end with 38B training tokens at a compression ratio of \(16\times\) and compare pre-training loss to find the best architecture. Since loss is computed only on the uncompressed tokens, the pre-training loss is substantially lower than a standard next-token prediction loss over all tokens. Extended experiments are in \Cref{app-sec:arch}.}
  \label{fig:arch_sweep}
\end{figure*}

\paragraph{Pooling operator.}
Prior work on context compression has primarily focused on two pooling strategies. 
Token-based pooling appends or prepends designated tokens, such as \texttt{EOS} or \texttt{CLS}, and uses their final hidden states as compressed representations~\citep{auto-compressor,in-context-autoencoder}.
Mean pooling averages encoder hidden states over each block of $N$ input tokens~\citep{GMSA,mean-pooling}, a design that is common in vision and multimodal models~\citep{team2026gemma}. 
We additionally consider concatenation, a widely used alternative in vision-style architectures but less explored for context compression \citep{bai2025qwen3, tong2024cambrian}. Concatenation preserves the $N$ sequential encoder hidden states by stacking them into a single wide vector, after which the adapter projects the resulting representation back to the LLM hidden dimension. The formal definitions of pooling operators can be found in \Cref{app-subsec:encoder-defs}.

\Cref{fig:arch_sweep} compares the three pooling operators in our pre-training sweep. Mean pooling matches or improves pre-training loss over token-based pooling, and performs on par with concatenation. This agrees with prior evidence that mean pooling can outperform token-based pooling for compression~\citep{mean-pooling}, and with findings that mean pooling can be surprisingly effective despite being a lossy operation in text embedding models~\citep{hara2026mean}. It is also consistent with related observations in vision architectures~\citep{chu2021conditional}. While concatenation can, in principle, preserve distinct representations that mean loses, and has been found to outperform mean pooling in other representation-learning settings~\citep{chen2025comodo}, we do not observe this advantage in our experiments. We compare these two operators at scale in \Cref{app-fig:pool-abl-at-scale,app-subsubsec:pooling-operator}, where concatenation outperforms mean pooling at low compression ratios and mean pooling outperforms concatenation at high compression ratios.

\paragraph{Encoding granularity.}
Ideally, one would set the encoder window size to the full context length, allowing the encoder to jointly contextualize all input tokens, analogous to ViT-style encoders \citep{dosovitskiy2020image} that process the entire image at once. For language inputs, however, the context length can be substantially larger, making full-context encoding prohibitive in both memory and time.

To study this trade-off, we vary the encoder window size over $W \in \{N, 256, 1024\}$ for mean pooling in \Cref{fig:arch_sweep}. We find that increasing $W$ from $N\;(16)$ \citep{refrag, e2llm} to $256$ leads to a large improvement, while increasing $W$ further from $256$ to $1024$ yields an additional smaller gain. Since $W=1024$ does not introduce significant memory or runtime overhead in our setting, we adopt $W=1024$ as the default. This setting enables the encoder to attend over a larger local context and produce richer compressed representations. One limitation of using $W < T$ is that the input must be chunked into multiple windows, so information crossing a boundary is split across two encoder forward passes. To address this, we also design a boundary-overlap variant that allows each window to see neighboring tokens during encoding. However, this approach does not consistently improve pre-training loss while increasing encoder compute, so we use no overlap by default and defer details to \Cref{app-subsec:arch-overlap}. 

\paragraph{Encoder attention mask.}
Unlike standard decoder language modeling, where causal masking is required to match autoregressive inference, context compression is applied to the prompt before decoding. The encoder can therefore use either causal or bidirectional attention, as in prefix-LMs~\citep{raffel2020exploring}. Prior compressors use both: some encode with a causal LLM~\citep{in-context-autoencoder,GMSA}, while others remove the causal mask~\citep{mean-pooling}. In \Cref{fig:arch_sweep,app-fig:adapter-ov-loss}, we compare these two masking choices and find that causal masking consistently achieves lower pre-training loss.

\paragraph{Adapter design.}
The adapter projects encoder latents from dimension \(d_{\mathrm{enc}}\) to the decoder embedding dimension \(d_{\mathrm{dec}}\). Prior encoder-decoder compression work often assumes matching encoder and decoder hidden sizes, but a wide range of encoder-decoder models pair smaller encoders with larger decoders, making the adapter a non-trivial design choice. Prior work uses either a small MLP projector~\citep{xrag,refrag} or attention-based layers~\citep{GMSA}. We compare a lightweight MLP projection against an attention-based adapter that adds a single self-attention layer over the latent sequence before the MLP projection. In contrast to prior findings \citep{GMSA}, we observe in \Cref{app-fig:adapter-ov-loss} that the MLP-only adapter achieves lower pre-training loss while using less compute, and we adopt it as the default. Formal definitions are provided in \Cref{app-subsec:adapter}.

\begin{figure*}[t]
  \centering
  \includegraphics[width=1.0\textwidth]{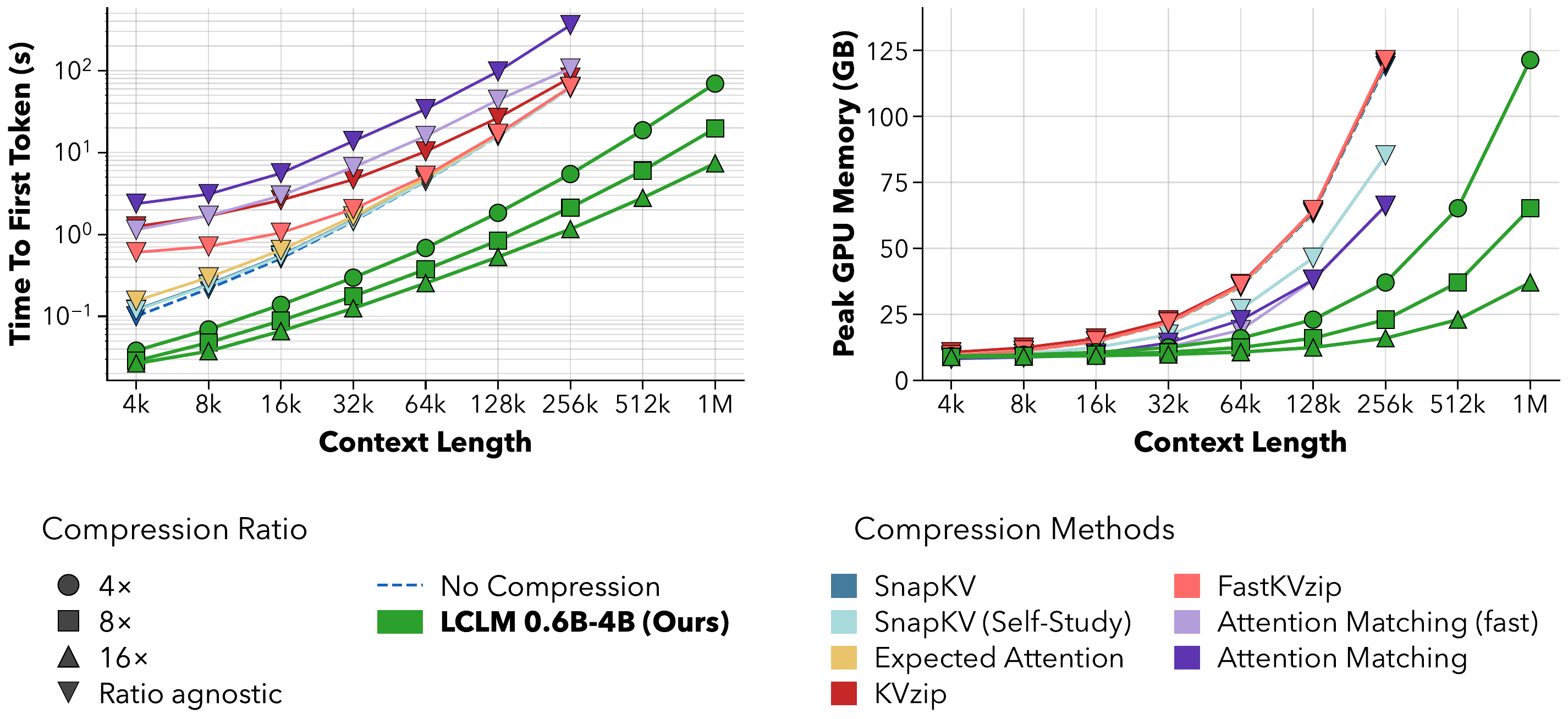}
    \caption{\textbf{\fullnames have lower TTFT and peak GPU memory as context length increases.}
    We plot time to first token (left) and peak GPU memory (right) for a single sample. We note the baselines use the same time and memory regardless of compression ratio; hence, there is only one line per baseline method.
    We record all measurements on the same H200 GPU, and truncate lines when the method cannot be performed within the 141GB of memory. For peak GPU memory, our encoder's activations stay small, so memory grows only with the decoder prefill over the compressed latent sequence.
    }
    \label{fig:pareto_optimal}
\end{figure*}

\paragraph{Optimal architecture at scale.}
\label{par:optimal-arch-at-scale}
Overall, our from-scratch pre-training experiments in \Cref{fig:arch_sweep} serve as a lightweight architecture sweep, allowing us to narrow the search space to a smaller set of promising design choices. We then scale these candidates using the full training pipeline, both to identify the strongest architecture at scale and to validate that the small-scale pre-training findings remain predictive under large-scale training. For our main experiments, we use \texttt{Qwen3-Embedding-0.6B} as the encoder with \texttt{Qwen3-4B-Instruct-2507} as the decoder.
Training at scale confirms the choices from the small-scale sweep: both the continual pre-training losses (\Cref{app-fig:pool-abl-at-scale,app-fig:adapter-ov-loss,app-fig:large-scale-sweep}) and the downstream ablations (\Cref{app-subsec:arch-sweep}) again favor an encoder window \(W=1024\) and causal masking. At scale, the attention-based adapter is competitive with the MLP adapter; we keep the MLP adapter, which adds less computation.
The choice of pooling operator depends on the compression ratio: mean pooling has a small edge over concatenation at \(16\times\) compression, while concatenation slightly outperforms mean pooling at \(4\times\). Since we primarily focus on \(16\times\) compression in our analysis, we adopt mean pooling as the default in all subsequent experiments. Full results and per-task breakdowns are reported in \Cref{app-subsec:arch-sweep}. 

We also study alternative model choices, including whether to initialize the encoder from an embedding model~\citep{e2llm,xrag} or a language model~\citep{in-context-autoencoder,GMSA,mean-pooling}, in \Cref{app-subsec:encoder_lm_or_emb}. In \Cref{app-sec:scaling model size}, we analyze the performance of scaling the sizes of the encoder and decoder.

\section{Results: Improving Latency / Memory-Performance Tradeoffs}
\label{sec:results}

We predominantly focus on long-context benchmarks: we evaluate on RULER~\citep{hsieh2024ruler}, LongBench~\citep{bai2024longbench}, and LongHealth~\citep{adams2025longhealth}.
For the long-context evaluation suite, we maintain instructions as uncompressed tokens, while we compress long-context segments. 
We report which parts of each benchmark are compressed and which are left uncompressed in \Cref{app-tab:eval-benchmarks}.

We measure two efficiency axes: peak memory during compression, and compression time.
Peak memory includes both the encoder and decoder for a single sample at each context length.
For every method, we report the fastest of 20 timed runs after 30 warmup runs on the same H200 GPU.
We define compression time as time-to-first-token (TTFT), i.e., the time required for a method to reach the point at which it can generate the first token.
We adapt the KVPress~\citep{devoto2025expected} timing code\footnote{\url{https://github.com/NVIDIA/kvpress/blob/main/notebooks/speed_and_memory.ipynb}} to support our method.
Because the KV cache compression baselines are implemented in Hugging Face Transformers~\citep{wolf2019huggingface}, we benchmark every method in their comparisons with it, including our decoder; \Cref{app-subsec:timing} details the implementation of every timing setup.

\begin{figure*}[t]
  \centering
  \includegraphics[width=1.0\textwidth]{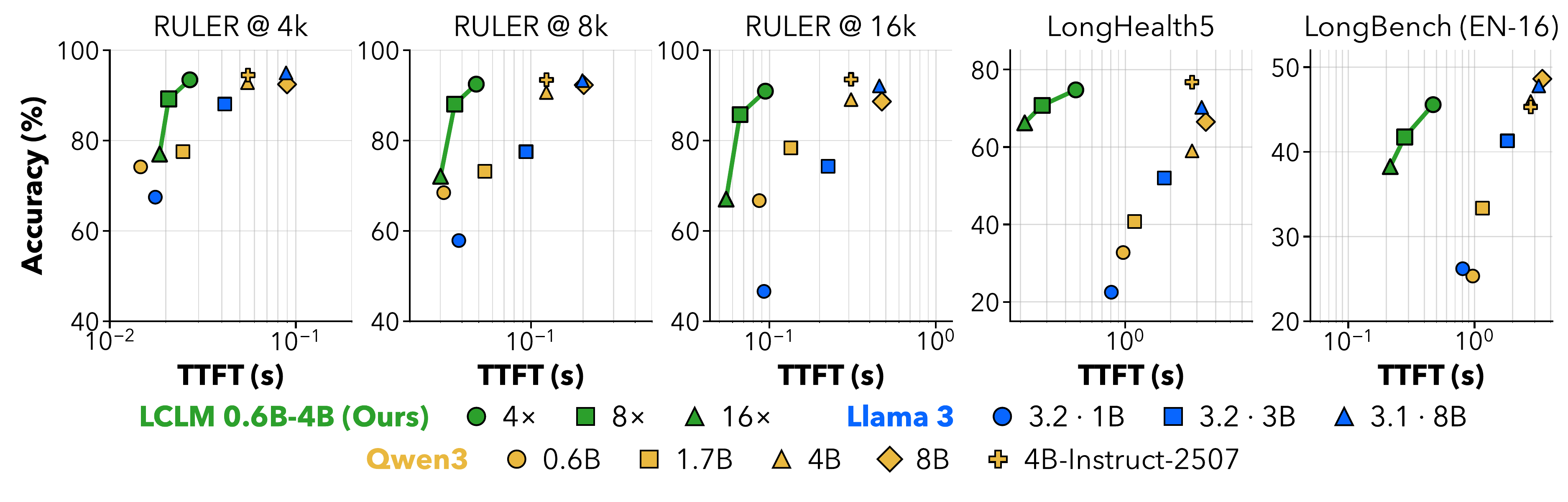}
  \caption{\textbf{\names are faster and more accurate than smaller uncompressed language models.} Accuracy vs.\ TTFT for \names and uncompressed Qwen3 and Llama~3 models from 0.6B to 8B parameters. All models are optimized for inference, with every decoder served by vLLM on a single H200, timed at fixed context lengths (4k, 8k, and 16k tokens for RULER; 64k for LongHealth5 and LongBench), taking the fastest of 20 trials after 30 warmup runs; \name TTFT includes encoding.}
  \label{fig:lm_pareto}
\end{figure*}

Since \names process input tokens in encoder-window chunks, we can parallelize compression by batching encoder forward passes before passing the resulting soft tokens to the decoder.
For all figures in this paper, we use an encoder batch size of \(32\); with encoder window size \(W=1024\), this corresponds to processing \(32{,}768\) input tokens per batched encoder pass.
We find this batch size provides a good trade-off between compression speed and encoder memory usage, improving GPU utilization without exceeding memory limits for ultra-long contexts.
Moreover, because our method preserves the standard KV cache structure, prefill and generation can be further accelerated with standard inference frameworks such as vLLM~\citep{kwon2023efficient} and SGLang~\citep{zheng2024sglang}, as well as efficient attention implementations~\citep{linearattention, kimilinear, deepseekv2, xu2026deepseek}.
We use this in \Cref{fig:lm_pareto}: that comparison involves only \names and uncompressed models, whose decoders all run in vLLM, so we serve every decoder there with vLLM and add our encoder's time for \names. The Hugging Face measurements in \Cref{fig:cover,fig:pareto_optimal} are therefore conservative relative to what optimized serving implementations could achieve.

\subsection{Baselines}
We consider a range of KV cache compression baselines, including Expected Attention~\citep{devoto2025expected}, Attention Matching~\citep{zweiger2026fast}, KVzip~\citep{kim2025kvzip}, FastKVzip~\citep{kim2026fast}, and SnapKV~\citep{li2024snapkv}. Each method has different native support for query-aware, query-agnostic, and self-study compression configurations, which we describe in \Cref{app-subsec:baselines} together with the difference between algorithmic and systems-realizable cache reduction.

We use the KVPress \citep{devoto2025expected} implementation for SnapKV, KVzip, FastKVzip, and Expected Attention. We use the official implementation for Attention Matching.

\begin{figure*}[t]
  \centering
  \includegraphics[width=1.0\textwidth]{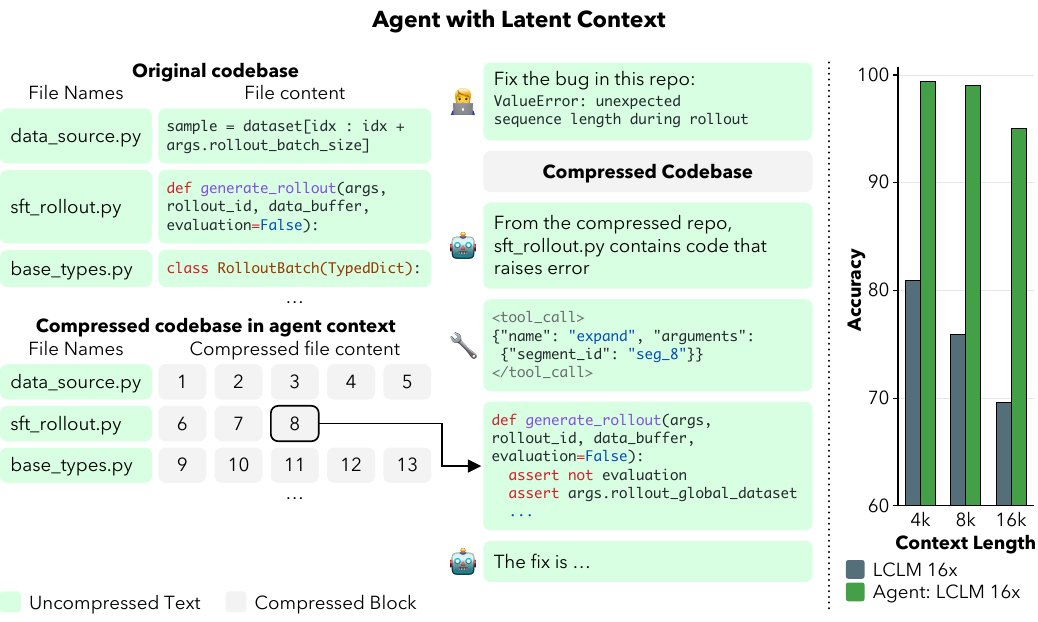}
  \caption{\textbf{\names can use tools to retrieve compressed context and improve retrieval accuracy.} \textbf{Left:} We equip \names with a tool that retrieves compressed context and inserts it into their working context uncompressed. \textbf{Right:} RULER NIAH scores at \(16\times\) compression, with and without expansion, averaged over eight tasks with 50 matched questions per task and context length. We cap the agent at 16 expansion calls at 4k and 8k and 32 at 16k; these caps were chosen on three of the eight tasks and kept fixed for the rest.
  }
  \label{fig:agentic_bar_plot}
\end{figure*}

\begin{figure*}[t]
  \centering
  \includegraphics[width=1.0\textwidth]{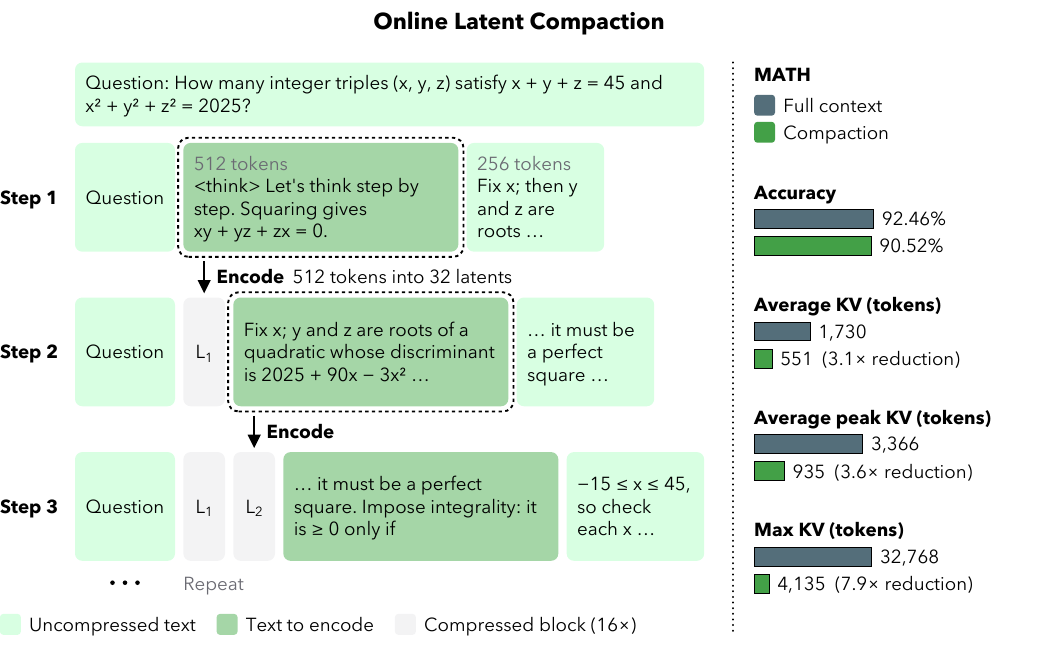}
  \caption{\textbf{Online latent compaction.} \textbf{Left:} During generation, every 512 uncompressed tokens are encoded into 32 latent tokens (\(16\times\) compression), while the prompt and the latest 256 tokens stay uncompressed. We iteratively compress and generate from the compacted context and the latest raw tokens. \textbf{Right:} Accuracy and decoder context length on the full MATH test set. Average KV and peak KV are computed per problem and averaged over problems; max KV is the largest per-problem peak, where the full-context limit is 32,768 tokens.}
  \label{fig:online_compaction}
\end{figure*}

\subsection{A New Pareto Frontier} 

We train \names with compression ratios of \(16\times\), \(8\times\), and \(4\times\). In \Cref{fig:cover}, \names establish a new Pareto frontier in compression time and accuracy over KV cache compression baselines on RULER, LongBench, and LongHealth. KV cache compression methods first materialize the full KV cache and then evict, mask, or compact it, so their compression time barely depends on the target ratio. In contrast, \names reduce the sequence length before decoder prefill, so higher compression ratios directly reduce decoder-side computation and memory.

\textbf{Comparison with smaller uncompressed models.} A simple alternative to compression is to run a smaller model on the full context. In \Cref{fig:lm_pareto}, we compare \names with uncompressed Qwen3~\citep{qwen3} and Llama~3~\citep{grattafiori2024llama} models from 0.6B to 8B parameters, with every model optimized for inference. At 64k tokens, the \(16\times\) \name reaches the first token faster than the smallest uncompressed models (0.21\,s, compared with 0.81\,s for Llama~3.2~1B and 0.97\,s for Qwen3-0.6B) while being far more accurate on LongHealth5 (66.25\% vs.\ 22.50\% and 32.75\%). The \(4\times\) \name exceeds the uncompressed \texttt{Qwen3-4B-Instruct-2507} on LongBench with a \(6\times\) lower TTFT.

\textbf{Time and memory scaling with context length.}
In \Cref{fig:pareto_optimal}, we visualize compression time and peak GPU memory from 4K to 1M tokens. \names achieve the fastest compression time at every context length and substantially lower peak memory at long contexts; all baselines run out of memory at 512K tokens, except Attention Matching, which fails at 512K due to numerical instability in its linear solver and runs out of memory at 1M. Because the encoder is small and processes at most 32K input tokens per batched forward pass, peak memory is driven by the decoder prefill over the compressed latent sequence: the \(16\times\) model grows from 9\,GB at 4K tokens to only 23\,GB at 512K, and at 1M tokens the \(16\times\), \(8\times\), and \(4\times\) models peak at 37, 65, and 121\,GB.

\section{New Capabilities from Compression}\label{sec:capabilities}
Because the decoder reads latent tokens in place of the original text, \names enable uses beyond compressing a fixed input. We study two: an agent that reads a compressed context and expands segments on demand (\Cref{sec:agentic}), and online latent compaction of the model's own generated reasoning (\Cref{sec:compaction}).

\subsection{Agent Workflow With Latent Context}\label{sec:agentic}
Agentic systems are commonly used to handle large amounts of information \citep{zhang2025rlm}.
However, agents usually rely on lexical or semantic search to locate the information needed. These information retrieval methods can fail when the search keyword is not obvious before reading through the information. For example, in a large codebase, a bug reported in the ``dashboard login flow'' may actually originate in an indirectly called entitlement module that never mentions ``dashboard'' or ``login''. Ideally, an agent should read all information at once before acting, but context limits of existing models prevent this from happening. 

LCLMs' more efficient long-context capacity makes it possible to place, for example, an entire code repository in an agent's context. We further enhance the agent's capability to locate precise information by allowing the \name agent to expand compressed segments into raw text on demand. We segment the input into fixed-size chunks of 1,024 encoder tokens, compress each chunk, and assign it a segment identifier. The model receives the entire compressed sequence in a single prompt, together with one \texttt{expand} tool. At each turn, the model can issue a native tool call to \texttt{expand} with a segment identifier (\Cref{fig:agentic_bar_plot}), and the tool returns the original text of that segment.

We use the needle-in-a-haystack tasks from RULER as a testbed for evaluating this agentic retrieval mechanism with latent context. As shown in \Cref{fig:agentic_bar_plot}, our agent substantially improves performance over the same \name with \(16\times\) compressed context and no expansion. These results suggest that compressed latent context can provide broad corpus-level visibility, while tool-based expansion supplies the exact fine-grained information needed.  The marked success of this initial agentic attempt suggests the promise of adaptive expansion.  LCLMs can skim a large quantity of context globally before deciding which relevant subset to zoom in on and read more carefully.  Future work may explore how to choose which context to expand, or learn this behavior end-to-end.

\subsection{Online Latent Compaction}\label{sec:compaction}
For long-horizon tasks, long CoT and agent trajectories can come to dominate the context and eventually fill the context window, requiring context compaction~\citep{claude_code,codex}. Here, we investigate whether we can use LCLMs to compact context into the latent space iteratively. Because the \name encoder can compress any span of text, we can also apply it to the model's own output during generation, which we call online latent compaction (\Cref{fig:online_compaction}). After the first 768 generated tokens, we encode the oldest 512 generated tokens into latent memory tokens at a \(16\times\) compression ratio, resulting in 32 latent tokens, and keep the most recent 256 tokens uncompressed. Generation then continues from the compacted context and latest raw tokens, and the procedure repeats after every further 512 generated tokens. Earlier latent blocks are reused thereafter.

We evaluate on all 5,000 problems of the MATH test set with the \(16\times\) \name, comparing generation with and without compaction under the same prompts, 32,768-token context limit, and scoring. Online latent compaction is nearly lossless, reaching 90.52\% accuracy compared with 92.46\% for full-context generation, while being highly efficient: it reduces the average KV length by \(3.1\times\), the average per-problem peak by \(3.6\times\), and the maximum peak by \(7.9\times\): the longest full-context trajectories reach the 32,768-token limit, whereas no compacted trajectory exceeds 4,135 tokens. Lengths count active decoder positions, including latents, not allocated memory.

\section{Conclusions}
\label{sec:discuss}
Through a controlled architecture search and end-to-end training at scale, our family of \fullnames shows that encoder-decoder compression can preserve strong in-context capabilities while compressing faster and using less memory than KV cache compression. The same models support agents that expand compressed context on demand and online compaction of the model's own reasoning, making them a practical building block for efficient long-context systems.

\fullnames open a broad design space for future work and are naturally compatible with agentic frameworks such as Recursive Language Models~\citep{zhang2025rlm}. 
More generally, \names provide a promising substrate for long-horizon agents with large, persistent working memories by dramatically reducing the size of the inputs at scale. 
Future iterations of \names could further improve the quality-efficiency trade-off by compressing inputs at multiple granularities and dynamically allocating capacity based on information density or input perplexity. 
Such adaptive compression could allow models to preserve fine-grained details where needed while maintaining a compact global context, reducing reliance on explicit expansion tools.  Another promising direction is to extend online latent compaction beyond chain-of-thought to tool observations and an agent's accumulated working history, which can grow to dominate the context budget in long-horizon tasks.

\section*{Acknowledgments} 
MG and ZC were supported by IBM, the Google Research Award, and the NVIDIA Academic Grant Program for Researchers.
SM and TG are supported by DOE Office of Science's ASCR AI for Science initiative, the NSF TRAILS Institute (2229885), and Coefficient Giving. NK is supported by the NSF Graduate Research Fellowship.

Computing resources for this project were supported in part by the Swiss National Supercomputing Center (CSCS). Prepared in collaboration with LLNL under Contract DE-AC52-07NA27344 and supported by the LLNL-LDRD Program under Project No. 24-ERD-010 (LLNL-CONF-2019739). This manuscript has been authored by Lawrence Livermore National Security, LLC under Contract No. DE-AC52-07NA27344 with the U.S. Department of Energy. The United States Government retains, and the publisher, by accepting the article for publication, acknowledges that the United States Government retains a non-exclusive, paid-up, irrevocable, world-wide license to publish or reproduce the published form of this manuscript, or allow others to do so, for United States Government purposes.
This material is based upon work supported by the U.S. Department of Energy, Office of Science, Office of Advanced Scientific Computing Research, through solicitation DE-FOA-0003264, ``Advancements in Artificial Intelligence for Science,'' under Award Number DE-SC0025598.

Special thanks to Shengbang Tong for extensive discussions on architecture design and encoder-decoder model training. We would like to thank Preston Zhou and Eitan Borgnia from Relace AI for their early discussions and support for the project. We also thank Vadim Bereznyuk, John Kirchenbauer, and Neel Jain for providing feedback on the draft.

\bibliography{refs}

\begin{thebibliography}{102}
\providecommand{\natexlab}[1]{#1}
\providecommand{\url}[1]{\texttt{#1}}
\expandafter\ifx\csname urlstyle\endcsname\relax
  \providecommand{\doi}[1]{doi: #1}\else
  \providecommand{\doi}{doi: \begingroup \urlstyle{rm}\Url}\fi

\bibitem[Adams et~al.(2025)Adams, Busch, Han, Excoffier, Ortala, L{\"o}ser,
  Aerts, Kather, Truhn, and Bressem]{adams2025longhealth}
Lisa Adams, Felix Busch, Tianyu Han, Jean-Baptiste Excoffier, Matthieu Ortala,
  Alexander L{\"o}ser, Hugo~JWL Aerts, Jakob~Nikolas Kather, Daniel Truhn, and
  Keno Bressem.
\newblock Longhealth: A question answering benchmark with long clinical
  documents.
\newblock \emph{Journal of Healthcare Informatics Research}, 9\penalty0
  (3):\penalty0 280--296, 2025.

\bibitem[Agarwal et~al.(2025)Agarwal, Ahmad, Ai, Altman, Applebaum, Arbus,
  Arora, Bai, Baker, Bao, et~al.]{agarwal2025gpt}
Sandhini Agarwal, Lama Ahmad, Jason Ai, Sam Altman, Andy Applebaum, Edwin
  Arbus, Rahul~K Arora, Yu~Bai, Bowen Baker, Haiming Bao, et~al.
\newblock gpt-oss-120b \& gpt-oss-20b model card.
\newblock \emph{arXiv preprint arXiv:2508.10925}, 2025.

\bibitem[Aggarwal et~al.(2025)Aggarwal, Singh, Awasthi, Kanade, and
  Natarajan]{aggarwal2025nextcoder}
Tushar Aggarwal, Swayam Singh, Abhijeet Awasthi, Aditya Kanade, and Nagarajan
  Natarajan.
\newblock Nextcoder: Robust adaptation of code {LM}s to diverse code edits.
\newblock In \emph{Forty-second International Conference on Machine Learning},
  2025.
\newblock URL \url{https://openreview.net/forum?id=3B6fF1PxYD}.

\bibitem[An et~al.(2025)An, Zhang, Zhong, Li, Gong, Luo, Xu, and
  Kong]{an2025why}
Chenxin An, Jun Zhang, Ming Zhong, Lei Li, Shansan Gong, Yao Luo, Jingjing Xu,
  and Lingpeng Kong.
\newblock Why does the effective context length of {LLM}s fall short?
\newblock In \emph{The Thirteenth International Conference on Learning
  Representations}, 2025.
\newblock URL \url{https://openreview.net/forum?id=eoln5WgrPx}.

\bibitem[{Anthropic}(2025)]{claude_code}
{Anthropic}.
\newblock {Claude Code}, 2025.
\newblock URL \url{https://github.com/anthropics/claude-code}.

\bibitem[Bai et~al.(2025)Bai, Cai, Chen, Chen, Chen, Cheng, Deng, Ding, Gao,
  Ge, et~al.]{bai2025qwen3}
Shuai Bai, Yuxuan Cai, Ruizhe Chen, Keqin Chen, Xionghui Chen, Zesen Cheng,
  Lianghao Deng, Wei Ding, Chang Gao, Chunjiang Ge, et~al.
\newblock {Qwen3-VL} technical report.
\newblock \emph{arXiv preprint arXiv:2511.21631}, 2025.

\bibitem[Bai et~al.(2024)Bai, Lv, Zhang, Lyu, Tang, Huang, Du, Liu, Zeng, Hou,
  Dong, Tang, and Li]{bai2024longbench}
Yushi Bai, Xin Lv, Jiajie Zhang, Hongchang Lyu, Jiankai Tang, Zhidian Huang,
  Zhengxiao Du, Xiao Liu, Aohan Zeng, Lei Hou, Yuxiao Dong, Jie Tang, and
  Juanzi Li.
\newblock {L}ong{B}ench: A bilingual, multitask benchmark for long context
  understanding.
\newblock In \emph{Proceedings of the 62nd Annual Meeting of the Association
  for Computational Linguistics (Volume 1: Long Papers)}, pages 3119--3137,
  Bangkok, Thailand, August 2024. Association for Computational Linguistics.
\newblock \doi{10.18653/v1/2024.acl-long.172}.
\newblock URL \url{https://aclanthology.org/2024.acl-long.172}.

\bibitem[Blakeman et~al.(2025)Blakeman, Grattafiori, Basant, Gupta, Khattar,
  Renduchintala, Vavre, Shukla, Bercovich, Ficek,
  et~al.]{nvidia_nemotron_nano_v3_2025}
Aaron Blakeman, Aaron Grattafiori, Aarti Basant, Abhibha Gupta, Abhinav
  Khattar, Adi Renduchintala, Aditya Vavre, Akanksha Shukla, Akhiad Bercovich,
  Aleksander Ficek, et~al.
\newblock Nemotron 3 nano: Open, efficient mixture-of-experts hybrid
  mamba-transformer model for agentic reasoning.
\newblock \emph{arXiv preprint arXiv:2512.20848}, 2025.

\bibitem[Chen et~al.(2025)Chen, Wongso, Li, Khaokaew, Xue, and
  Salim]{chen2025comodo}
Baiyu Chen, Wilson Wongso, Zechen Li, Yonchanok Khaokaew, Hao Xue, and Flora
  Salim.
\newblock Comodo: Cross-modal video-to-imu distillation for efficient
  egocentric human activity recognition.
\newblock \emph{arXiv preprint arXiv:2503.07259}, 2025.

\bibitem[Chen(2023)]{October2001_AwesomeKVCacheCompression_2025}
Longze Chen.
\newblock Awesome-kv-cache-compression.
\newblock GitHub repository, 2023.
\newblock URL
  \url{https://github.com/October2001/Awesome-KV-Cache-Compression}.

\bibitem[Cheng et~al.(2024)Cheng, Wang, Zhang, Ge, Chen, Wei, Zhang, and
  Zhao]{xrag}
Xin Cheng, Xun Wang, Xingxing Zhang, Tao Ge, Si-Qing Chen, Furu Wei, Huishuai
  Zhang, and Dongyan Zhao.
\newblock xrag: Extreme context compression for retrieval-augmented generation
  with one token.
\newblock \emph{Advances in Neural Information Processing Systems},
  37:\penalty0 109487--109516, 2024.

\bibitem[Chevalier et~al.(2023)Chevalier, Wettig, Ajith, and
  Chen]{auto-compressor}
Alexis Chevalier, Alexander Wettig, Anirudh Ajith, and Danqi Chen.
\newblock Adapting language models to compress contexts.
\newblock \emph{arXiv preprint arXiv:2305.14788}, 2023.

\bibitem[Chu et~al.(2021)Chu, Tian, Zhang, Wang, and Shen]{chu2021conditional}
Xiangxiang Chu, Zhi Tian, Bo~Zhang, Xinlong Wang, and Chunhua Shen.
\newblock Conditional positional encodings for vision transformers.
\newblock \emph{arXiv preprint arXiv:2102.10882}, 2021.

\bibitem[Chuang et~al.(2024)Chuang, Xing, Chang, Liu, Chen, and
  Hu]{chuang2024learning}
Yu-Neng Chuang, Tianwei Xing, Chia-Yuan Chang, Zirui Liu, Xun Chen, and Xia Hu.
\newblock Learning to compress prompt in natural language formats.
\newblock \emph{arXiv preprint arXiv:2402.18700}, 2024.

\bibitem[Cobbe et~al.(2021)Cobbe, Kosaraju, Bavarian, Chen, Jun, Kaiser,
  Plappert, Tworek, Hilton, Nakano, et~al.]{cobbe2021training}
Karl Cobbe, Vineet Kosaraju, Mohammad Bavarian, Mark Chen, Heewoo Jun, Lukasz
  Kaiser, Matthias Plappert, Jerry Tworek, Jacob Hilton, Reiichiro Nakano,
  et~al.
\newblock Training verifiers to solve math word problems.
\newblock \emph{arXiv preprint arXiv:2110.14168}, 2021.

\bibitem[Cotroneo et~al.(2025)Cotroneo, De~Rosa, and Liguori]{11052783}
Domenico Cotroneo, Giuseppe De~Rosa, and Pietro Liguori.
\newblock Pyresbugs: A dataset of residual python bugs for natural
  language-driven fault injection.
\newblock In \emph{2025 IEEE/ACM Second International Conference on AI
  Foundation Models and Software Engineering (Forge)}, pages 146--150, 2025.
\newblock \doi{10.1109/Forge66646.2025.00024}.

\bibitem[Dai et~al.(2025)Dai, Lian, Huang, Zhang, Zhou, Wu, Xie, and
  Liao]{dai2025pretraining}
Yuhong Dai, Jianxun Lian, Yitian Huang, Wei Zhang, Mingyang Zhou, Mingqi Wu,
  Xing Xie, and Hao Liao.
\newblock Pretraining context compressor for large language models with
  embedding-based memory.
\newblock In \emph{Proceedings of the 63rd Annual Meeting of the Association
  for Computational Linguistics (Volume 1: Long Papers)}, pages 28715--28732,
  2025.

\bibitem[Dai et~al.(2019)Dai, Yang, Yang, Carbonell, Le, and
  Salakhutdinov]{transformerxl}
Zihang Dai, Zhilin Yang, Yiming Yang, Jaime~G Carbonell, Quoc Le, and Ruslan
  Salakhutdinov.
\newblock Transformer-xl: Attentive language models beyond a fixed-length
  context.
\newblock In \emph{Proceedings of the 57th annual meeting of the association
  for computational linguistics}, pages 2978--2988, 2019.

\bibitem[Dao(2023)]{dao2023flashattention}
Tri Dao.
\newblock Flashattention-2: Faster attention with better parallelism and work
  partitioning.
\newblock \emph{arXiv preprint arXiv:2307.08691}, 2023.

\bibitem[Devoto et~al.(2025)Devoto, Jeblick, and J{\'e}gou]{devoto2025expected}
Alessio Devoto, Maximilian Jeblick, and Simon J{\'e}gou.
\newblock Expected attention: Kv cache compression by estimating attention from
  future queries distribution.
\newblock \emph{arXiv preprint arXiv:2510.00636}, 2025.

\bibitem[Dong et~al.(2024)Dong, Feng, Guessous, Liang, and He]{dong2024flex}
Juechu Dong, Boyuan Feng, Driss Guessous, Yanbo Liang, and Horace He.
\newblock Flex attention: A programming model for generating optimized
  attention kernels.
\newblock \emph{arXiv preprint arXiv:2412.05496}, 2\penalty0 (3):\penalty0 4,
  2024.

\bibitem[Dosovitskiy et~al.(2020)Dosovitskiy, Beyer, Kolesnikov, Weissenborn,
  Zhai, Unterthiner, Dehghani, Minderer, Heigold, Gelly,
  et~al.]{dosovitskiy2020image}
Alexey Dosovitskiy, Lucas Beyer, Alexander Kolesnikov, Dirk Weissenborn,
  Xiaohua Zhai, Thomas Unterthiner, Mostafa Dehghani, Matthias Minderer, Georg
  Heigold, Sylvain Gelly, et~al.
\newblock An image is worth 16x16 words: Transformers for image recognition at
  scale.
\newblock \emph{arXiv preprint arXiv:2010.11929}, 2020.

\bibitem[Eyuboglu et~al.(2025)Eyuboglu, Ehrlich, Arora, Guha, Zinsley, Liu,
  Tennien, Rudra, Zou, Mirhoseini, et~al.]{cartridges}
Sabri Eyuboglu, Ryan Ehrlich, Simran Arora, Neel Guha, Dylan Zinsley, Emily
  Liu, Will Tennien, Atri Rudra, James Zou, Azalia Mirhoseini, et~al.
\newblock Cartridges: Lightweight and general-purpose long context
  representations via self-study.
\newblock \emph{arXiv preprint arXiv:2506.06266}, 2025.

\bibitem[Feldman and Artzi(2025)]{mean-pooling}
Yair Feldman and Yoav Artzi.
\newblock Simple context compression: Mean-pooling and multi-ratio training.
\newblock \emph{arXiv preprint arXiv:2510.20797}, 2025.

\bibitem[Gao et~al.(2025)Gao, Wettig, Yen, and Chen]{gao2025train}
Tianyu Gao, Alexander Wettig, Howard Yen, and Danqi Chen.
\newblock How to train long-context language models (effectively).
\newblock In \emph{Proceedings of the 63rd Annual Meeting of the Association
  for Computational Linguistics (Volume 1: Long Papers)}, pages 7376--7399,
  2025.

\bibitem[Ge et~al.(2023)Ge, Hu, Wang, Wang, Chen, and
  Wei]{in-context-autoencoder}
Tao Ge, Jing Hu, Lei Wang, Xun Wang, Si-Qing Chen, and Furu Wei.
\newblock In-context autoencoder for context compression in a large language
  model.
\newblock \emph{arXiv preprint arXiv:2307.06945}, 2023.

\bibitem[{Gemma Team} et~al.(2026){Gemma Team}, Abd, Aggarwal, Algayres,
  Andreev, Bachem, Ballantyne, Brick, C{\u{a}}rbune, Casbon,
  et~al.]{team2026gemma}
{Gemma Team}, Sherif~El Abd, Vaibhav Aggarwal, Robin Algayres, Alek Andreev,
  Olivier Bachem, Ian Ballantyne, Cormac Brick, Victor C{\u{a}}rbune, Michelle
  Casbon, et~al.
\newblock {Gemma 4} technical report.
\newblock \emph{arXiv preprint arXiv:2607.02770}, 2026.

\bibitem[Grattafiori et~al.(2024)Grattafiori, Dubey, Jauhri, Pandey, Kadian,
  Al-Dahle, Letman, Mathur, Schelten, Vaughan, et~al.]{grattafiori2024llama}
Aaron Grattafiori, Abhimanyu Dubey, Abhinav Jauhri, Abhinav Pandey, Abhishek
  Kadian, Ahmad Al-Dahle, Aiesha Letman, Akhil Mathur, Alan Schelten, Alex
  Vaughan, et~al.
\newblock The {Llama} 3 herd of models.
\newblock \emph{arXiv preprint arXiv:2407.21783}, 2024.

\bibitem[Gu and Dao(2024)]{mamba}
Albert Gu and Tri Dao.
\newblock Mamba: Linear-time sequence modeling with selective state spaces.
\newblock In \emph{First conference on language modeling}, 2024.

\bibitem[Gu et~al.(2021)Gu, Goel, and R{\'e}]{s4}
Albert Gu, Karan Goel, and Christopher R{\'e}.
\newblock Efficiently modeling long sequences with structured state spaces.
\newblock \emph{arXiv preprint arXiv:2111.00396}, 2021.

\bibitem[Guo et~al.(2025)Guo, Yang, Zhang, Song, Wang, Zhu, Xu, Zhang, Ma, Bi,
  et~al.]{guo2025deepseek}
Daya Guo, Dejian Yang, Haowei Zhang, Junxiao Song, Peiyi Wang, Qihao Zhu,
  Runxin Xu, Ruoyu Zhang, Shirong Ma, Xiao Bi, et~al.
\newblock Deepseek-r1: Incentivizing reasoning capability in llms via
  reinforcement learning.
\newblock \emph{arXiv preprint arXiv:2501.12948}, 2025.

\bibitem[Hara et~al.(2026)Hara, Kurita, Imaizumi, Inui, and
  Yokoi]{hara2026mean}
Tomomasa Hara, Hiroto Kurita, Masaaki Imaizumi, Kentaro Inui, and Sho Yokoi.
\newblock Why mean pooling works: Quantifying second-order collapse in text
  embeddings.
\newblock \emph{arXiv preprint arXiv:2604.27398}, 2026.

\bibitem[He et~al.(2025)He, Wang, Weber, Zhu, Athiwaratkun, and
  Zhang]{he2025scaling}
Linda He, Jue Wang, Maurice Weber, Shang Zhu, Ben Athiwaratkun, and Ce~Zhang.
\newblock Scaling instruction-tuned llms to million-token contexts via
  hierarchical synthetic data generation.
\newblock \emph{arXiv preprint arXiv:2504.12637}, 2025.

\bibitem[Hendrycks and Gimpel(2016)]{hendrycks2016gaussian}
Dan Hendrycks and Kevin Gimpel.
\newblock Gaussian error linear units (gelus).
\newblock \emph{arXiv preprint arXiv:1606.08415}, 2016.

\bibitem[Hooper et~al.(2024)Hooper, Kim, Mohammadzadeh, Mahoney, Shao, Keutzer,
  and Gholami]{hooper2024kvquant}
Coleman Hooper, Sehoon Kim, Hiva Mohammadzadeh, Michael~W Mahoney, Yakun~S
  Shao, Kurt Keutzer, and Amir Gholami.
\newblock Kvquant: Towards 10 million context length llm inference with kv
  cache quantization.
\newblock \emph{Advances in Neural Information Processing Systems},
  37:\penalty0 1270--1303, 2024.

\bibitem[Hsieh et~al.(2024)Hsieh, Sun, Kriman, Acharya, Rekesh, Jia, Zhang, and
  Ginsburg]{hsieh2024ruler}
Cheng-Ping Hsieh, Simeng Sun, Samuel Kriman, Shantanu Acharya, Dima Rekesh, Fei
  Jia, Yang Zhang, and Boris Ginsburg.
\newblock Ruler: What's the real context size of your long-context language
  models?
\newblock \emph{arXiv preprint arXiv:2404.06654}, 2024.

\bibitem[Jiang et~al.(2023)Jiang, Wu, Lin, Yang, and Qiu]{jiang2023llmlingua}
Huiqiang Jiang, Qianhui Wu, Chin-Yew Lin, Yuqing Yang, and Lili Qiu.
\newblock Llmlingua: Compressing prompts for accelerated inference of large
  language models.
\newblock \emph{arXiv preprint arXiv:2310.05736}, 2023.

\bibitem[Jiang et~al.(2024)Jiang, Wu, Luo, Li, Lin, Yang, and
  Qiu]{jiang2024longllmlingua}
Huiqiang Jiang, Qianhui Wu, Xufang Luo, Dongsheng Li, Chin-Yew Lin, Yuqing
  Yang, and Lili Qiu.
\newblock Longllmlingua: Accelerating and enhancing llms in long context
  scenarios via prompt compression.
\newblock In \emph{Proceedings of the 62nd Annual Meeting of the Association
  for Computational Linguistics (Volume 1: Long Papers)}, pages 1658--1677,
  2024.

\bibitem[Jin et~al.(2019)Jin, Dhingra, Liu, Cohen, and Lu]{jin2019pubmedqa}
Qiao Jin, Bhuwan Dhingra, Zhengping Liu, William Cohen, and Xinghua Lu.
\newblock Pubmedqa: A dataset for biomedical research question answering.
\newblock In \emph{Proceedings of the 2019 conference on empirical methods in
  natural language processing and the 9th international joint conference on
  natural language processing (EMNLP-IJCNLP)}, pages 2567--2577, 2019.

\bibitem[Joshi et~al.(2017)Joshi, Choi, Weld, and
  Zettlemoyer]{joshi2017triviaqa}
Mandar Joshi, Eunsol Choi, Daniel~S Weld, and Luke Zettlemoyer.
\newblock Triviaqa: A large scale distantly supervised challenge dataset for
  reading comprehension.
\newblock \emph{arXiv preprint arXiv:1705.03551}, 2017.

\bibitem[Katharopoulos et~al.(2020)Katharopoulos, Vyas, Pappas, and
  Fleuret]{linearattention}
Angelos Katharopoulos, Apoorv Vyas, Nikolaos Pappas, and Fran{\c{c}}ois
  Fleuret.
\newblock Transformers are rnns: Fast autoregressive transformers with linear
  attention.
\newblock In \emph{International conference on machine learning}, pages
  5156--5165. PMLR, 2020.

\bibitem[Kim et~al.(2025)Kim, Kim, Kwon, Lee, Yun, and Song]{kim2025kvzip}
Jang-Hyun Kim, Jinuk Kim, Sangwoo Kwon, Jae~W Lee, Sangdoo Yun, and Hyun~Oh
  Song.
\newblock Kvzip: Query-agnostic kv cache compression with context
  reconstruction.
\newblock \emph{arXiv preprint arXiv:2505.23416}, 2025.

\bibitem[Kim et~al.(2026)Kim, Han, and Yun]{kim2026fast}
Jang-Hyun Kim, Dongyoon Han, and Sangdoo Yun.
\newblock Fast kvzip: Efficient and accurate llm inference with gated kv
  eviction.
\newblock \emph{arXiv preprint arXiv:2601.17668}, 2026.

\bibitem[{Kimi Team} et~al.(2025){Kimi Team}, Zhang, Lin, Yao, Hu, Meng, Liu,
  Men, Yang, Li, et~al.]{kimilinear}
{Kimi Team}, Yu~Zhang, Zongyu Lin, Xingcheng Yao, Jiaxi Hu, Fanqing Meng,
  Chengyin Liu, Xin Men, Songlin Yang, Zhiyuan Li, et~al.
\newblock Kimi linear: An expressive, efficient attention architecture.
\newblock \emph{arXiv preprint arXiv:2510.26692}, 2025.

\bibitem[Kwon et~al.(2023)Kwon, Li, Zhuang, Sheng, Zheng, Yu, Gonzalez, Zhang,
  and Stoica]{kwon2023efficient}
Woosuk Kwon, Zhuohan Li, Siyuan Zhuang, Ying Sheng, Lianmin Zheng, Cody~Hao Yu,
  Joseph Gonzalez, Hao Zhang, and Ion Stoica.
\newblock Efficient memory management for large language model serving with
  pagedattention.
\newblock In \emph{Proceedings of the 29th symposium on operating systems
  principles}, pages 611--626, 2023.

\bibitem[Lambert et~al.(2024)Lambert, Morrison, Pyatkin, Huang, Ivison,
  Brahman, Miranda, Liu, Dziri, Lyu, Gu, Malik, Graf, Hwang, Yang, Bras,
  Tafjord, Wilhelm, Soldaini, Smith, Wang, Dasigi, and
  Hajishirzi]{lambert2024tulu3}
Nathan Lambert, Jacob Morrison, Valentina Pyatkin, Shengyi Huang, Hamish
  Ivison, Faeze Brahman, Lester James~V. Miranda, Alisa Liu, Nouha Dziri, Shane
  Lyu, Yuling Gu, Saumya Malik, Victoria Graf, Jena~D. Hwang, Jiangjiang Yang,
  Ronan~Le Bras, Oyvind Tafjord, Chris Wilhelm, Luca Soldaini, Noah~A. Smith,
  Yizhong Wang, Pradeep Dasigi, and Hannaneh Hajishirzi.
\newblock Tülu 3: Pushing frontiers in open language model post-training,
  2024.

\bibitem[{\L}a{\'n}cucki et~al.(2026){\L}a{\'n}cucki, Staniszewski, Nawrot, and
  Ponti]{lancucki2026inference}
Adrian {\L}a{\'n}cucki, Konrad Staniszewski, Piotr Nawrot, and Edoardo~Maria
  Ponti.
\newblock Inference-time hyper-scaling with {KV} cache compression.
\newblock \emph{Advances in Neural Information Processing Systems},
  38:\penalty0 9365--9397, 2026.

\bibitem[Li et~al.(2024{\natexlab{a}})Li, Ding, Fang, and
  Tao]{li2024revisiting}
Hongyu Li, Liang Ding, Meng Fang, and Dacheng Tao.
\newblock Revisiting catastrophic forgetting in large language model tuning.
\newblock In \emph{Findings of the association for computational linguistics:
  EMNLP 2024}, pages 4297--4308, 2024{\natexlab{a}}.

\bibitem[Li and Liang(2021)]{li2021prefix}
Xiang~Lisa Li and Percy Liang.
\newblock Prefix-tuning: Optimizing continuous prompts for generation.
\newblock \emph{arXiv preprint arXiv:2101.00190}, 2021.

\bibitem[Li et~al.(2023)Li, Dong, Guerin, and Lin]{li2023compressing}
Yucheng Li, Bo~Dong, Frank Guerin, and Chenghua Lin.
\newblock Compressing context to enhance inference efficiency of large language
  models.
\newblock In \emph{Proceedings of the 2023 conference on empirical methods in
  natural language processing}, pages 6342--6353, 2023.

\bibitem[Li et~al.(2024{\natexlab{b}})Li, Jiang, Wu, Luo, Ahn, Zhang, Abdi, Li,
  Gao, Yang, et~al.]{li2024scbench}
Yucheng Li, Huiqiang Jiang, Qianhui Wu, Xufang Luo, Surin Ahn, Chengruidong
  Zhang, Amir~H Abdi, Dongsheng Li, Jianfeng Gao, Yuqing Yang, et~al.
\newblock Scbench: A kv cache-centric analysis of long-context methods.
\newblock \emph{arXiv preprint arXiv:2412.10319}, 2024{\natexlab{b}}.

\bibitem[Li et~al.(2024{\natexlab{c}})Li, Huang, Yang, Venkitesh, Locatelli,
  Ye, Cai, Lewis, and Chen]{li2024snapkv}
Yuhong Li, Yingbing Huang, Bowen Yang, Bharat Venkitesh, Acyr Locatelli,
  Hanchen Ye, Tianle Cai, Patrick Lewis, and Deming Chen.
\newblock Snapkv: Llm knows what you are looking for before generation.
\newblock \emph{Advances in Neural Information Processing Systems},
  37:\penalty0 22947--22970, 2024{\natexlab{c}}.

\bibitem[Li et~al.(2025)Li, Su, and Collier]{li2025500xcompressor}
Zongqian Li, Yixuan Su, and Nigel Collier.
\newblock 500xcompressor: Generalized prompt compression for large language
  models.
\newblock In \emph{Proceedings of the 63rd Annual Meeting of the Association
  for Computational Linguistics (Volume 1: Long Papers)}, pages 25081--25091,
  2025.

\bibitem[Liao et~al.(2025)Liao, Wang, Yu, Wei, Li, and Zhang]{e2llm}
Zihan Liao, Jun Wang, Hang Yu, Lingxiao Wei, Jianguo Li, and Wei Zhang.
\newblock E2llm: Encoder elongated large language models for long-context
  understanding and reasoning.
\newblock In \emph{Proceedings of the 2025 Conference on Empirical Methods in
  Natural Language Processing}, pages 19212--19241, 2025.

\bibitem[Lin et~al.(2025)Lin, Ghosh, Low, Shrivastava, and Mohan]{refrag}
Xiaoqiang Lin, Aritra Ghosh, Bryan Kian~Hsiang Low, Anshumali Shrivastava, and
  Vijai Mohan.
\newblock Refrag: Rethinking rag based decoding.
\newblock \emph{arXiv preprint arXiv:2509.01092}, 2025.

\bibitem[Liu et~al.(2024{\natexlab{a}})Liu, Feng, Wang, Wang, Liu, Zhao, Dengr,
  Ruan, Dai, Guo, et~al.]{deepseekv2}
Aixin Liu, Bei Feng, Bin Wang, Bingxuan Wang, Bo~Liu, Chenggang Zhao, Chengqi
  Dengr, Chong Ruan, Damai Dai, Daya Guo, et~al.
\newblock Deepseek-v2: A strong, economical, and efficient mixture-of-experts
  language model.
\newblock \emph{arXiv preprint arXiv:2405.04434}, 2024{\natexlab{a}}.

\bibitem[Liu et~al.(2023{\natexlab{a}})Liu, Li, Wu, and Lee]{llava}
Haotian Liu, Chunyuan Li, Qingyang Wu, and Yong~Jae Lee.
\newblock Visual instruction tuning.
\newblock \emph{Advances in neural information processing systems},
  36:\penalty0 34892--34916, 2023{\natexlab{a}}.

\bibitem[Liu et~al.(2024{\natexlab{b}})Liu, Lin, Hewitt, Paranjape, Bevilacqua,
  Petroni, and Liang]{liu2024lost}
Nelson~F Liu, Kevin Lin, John Hewitt, Ashwin Paranjape, Michele Bevilacqua,
  Fabio Petroni, and Percy Liang.
\newblock Lost in the middle: How language models use long contexts.
\newblock \emph{Transactions of the association for computational linguistics},
  12:\penalty0 157--173, 2024{\natexlab{b}}.

\bibitem[Liu et~al.(2023{\natexlab{b}})Liu, Xu, and McAuley]{liu2023repobench}
Tianyang Liu, Canwen Xu, and Julian McAuley.
\newblock Repobench: Benchmarking repository-level code auto-completion
  systems.
\newblock \emph{arXiv preprint arXiv:2306.03091}, 2023{\natexlab{b}}.

\bibitem[Liu et~al.(2024{\natexlab{c}})Liu, Chen, Ji, Zhou, Chen, and
  Wang]{liu2024raginstructboostingllmsdiverse}
Wanlong Liu, Junying Chen, Ke~Ji, Li~Zhou, Wenyu Chen, and Benyou Wang.
\newblock Rag-instruct: Boosting llms with diverse retrieval-augmented
  instructions, 2024{\natexlab{c}}.
\newblock URL \url{https://arxiv.org/abs/2501.00353}.

\bibitem[Liu et~al.(2024{\natexlab{d}})Liu, Ping, Roy, Xu, Lee, Shoeybi, and
  Catanzaro]{liu2024chatqa}
Zihan Liu, Wei Ping, Rajarshi Roy, Peng Xu, Chankyu Lee, Mohammad Shoeybi, and
  Bryan Catanzaro.
\newblock Chatqa: Building gpt-4 level conversational qa models.
\newblock \emph{CoRR}, 2024{\natexlab{d}}.

\bibitem[Lozhkov et~al.(2024)Lozhkov, Li, Allal, Cassano, Lamy-Poirier, Tazi,
  Tang, Pykhtar, Liu, Wei, Liu, Tian, Kocetkov, Zucker, Belkada, Wang, Liu,
  Abulkhanov, Paul, Li, Li, Risdal, Li, Zhu, Zhuo, Zheltonozhskii, Dade, Yu,
  Krauß, Jain, Su, He, Dey, Abati, Chai, Muennighoff, Tang, Oblokulov, Akiki,
  Marone, Mou, Mishra, Gu, Hui, Dao, Zebaze, Dehaene, Patry, Xu, McAuley, Hu,
  Scholak, Paquet, Robinson, Anderson, Chapados, Patwary, Tajbakhsh, Jernite,
  Ferrandis, Zhang, Hughes, Wolf, Guha, von Werra, and
  de~Vries]{lozhkov2024starcoder}
Anton Lozhkov, Raymond Li, Loubna~Ben Allal, Federico Cassano, Joel
  Lamy-Poirier, Nouamane Tazi, Ao~Tang, Dmytro Pykhtar, Jiawei Liu, Yuxiang
  Wei, Tianyang Liu, Max Tian, Denis Kocetkov, Arthur Zucker, Younes Belkada,
  Zijian Wang, Qian Liu, Dmitry Abulkhanov, Indraneil Paul, Zhuang Li, Wen-Ding
  Li, Megan Risdal, Jia Li, Jian Zhu, Terry~Yue Zhuo, Evgenii Zheltonozhskii,
  Nii Osae~Osae Dade, Wenhao Yu, Lucas Krauß, Naman Jain, Yixuan Su, Xuanli
  He, Manan Dey, Edoardo Abati, Yekun Chai, Niklas Muennighoff, Xiangru Tang,
  Muhtasham Oblokulov, Christopher Akiki, Marc Marone, Chenghao Mou, Mayank
  Mishra, Alex Gu, Binyuan Hui, Tri Dao, Armel Zebaze, Olivier Dehaene, Nicolas
  Patry, Canwen Xu, Julian McAuley, Han Hu, Torsten Scholak, Sebastien Paquet,
  Jennifer Robinson, Carolyn~Jane Anderson, Nicolas Chapados, Mostofa Patwary,
  Nima Tajbakhsh, Yacine Jernite, Carlos~Muñoz Ferrandis, Lingming Zhang, Sean
  Hughes, Thomas Wolf, Arjun Guha, Leandro von Werra, and Harm de~Vries.
\newblock Starcoder 2 and the stack v2: The next generation, 2024.

\bibitem[Luo et~al.(2025)Luo, Yang, Meng, Li, Zhou, and
  Zhang]{luo2025empirical}
Yun Luo, Zhen Yang, Fandong Meng, Yafu Li, Jie Zhou, and Yue Zhang.
\newblock An empirical study of catastrophic forgetting in large language
  models during continual fine-tuning.
\newblock \emph{IEEE Transactions on Audio, Speech and Language Processing},
  2025.

\bibitem[Mu et~al.(2023)Mu, Li, and Goodman]{gist}
Jesse Mu, Xiang Li, and Noah Goodman.
\newblock Learning to compress prompts with gist tokens.
\newblock \emph{Advances in Neural Information Processing Systems},
  36:\penalty0 19327--19352, 2023.

\bibitem[Muennighoff et~al.(2023)Muennighoff, Liu, Zebaze, Zheng, Hui, Zhuo,
  Singh, Tang, von Werra, and Longpre]{muennighoff2023octopack}
Niklas Muennighoff, Qian Liu, Armel Zebaze, Qinkai Zheng, Binyuan Hui,
  Terry~Yue Zhuo, Swayam Singh, Xiangru Tang, Leandro von Werra, and Shayne
  Longpre.
\newblock Octopack: Instruction tuning code large language models.
\newblock \emph{arXiv preprint arXiv:2308.07124}, 2023.

\bibitem[Nathawani et~al.(2025{\natexlab{a}})Nathawani, Ding, Lavrukhin,
  Gitman, Majumdar, Bakhturina, Ginsburg, and
  Polak~Scowcroft]{NemotronPostTrainingDatasetV2}
Dhruv Nathawani, Shuoyang Ding, Vitaly Lavrukhin, Igor Gitman, Somshubra
  Majumdar, Evelina Bakhturina, Boris Ginsburg, and Jane Polak~Scowcroft.
\newblock {Nemotron-Post-Training-Dataset-v2}, aug 2025{\natexlab{a}}.
\newblock URL
  \url{https://huggingface.co/datasets/nvidia/Nemotron-Post-Training-Dataset-v2}.

\bibitem[Nathawani et~al.(2025{\natexlab{b}})Nathawani, Gitman, Majumdar,
  Bakhturina, Sunil~Mahabaleshwarkar, Zhang, and
  Polak~Scowcroft]{NemotronPostTrainingDatasetV1}
Dhruv Nathawani, Igor Gitman, Somshubra Majumdar, Evelina Bakhturina, Ameya
  Sunil~Mahabaleshwarkar, Jian Zhang, and Jane Polak~Scowcroft.
\newblock {Nemotron-Post-Training-Dataset-v1}, July 2025{\natexlab{b}}.
\newblock URL
  \url{https://huggingface.co/datasets/nvidia/Nemotron-Post-Training-Dataset-v1}.

\bibitem[Nawrot et~al.(2024)Nawrot, {\L}a\'{n}cucki, Chochowski, Tarjan, and
  Ponti]{pmlr-v235-nawrot24a}
Piotr Nawrot, Adrian {\L}a\'{n}cucki, Marcin Chochowski, David Tarjan, and
  Edoardo Ponti.
\newblock Dynamic memory compression: Retrofitting {LLM}s for accelerated
  inference.
\newblock In Ruslan Salakhutdinov, Zico Kolter, Katherine Heller, Adrian
  Weller, Nuria Oliver, Jonathan Scarlett, and Felix Berkenkamp, editors,
  \emph{Proceedings of the 41st International Conference on Machine Learning},
  volume 235 of \emph{Proceedings of Machine Learning Research}, pages
  37396--37412. PMLR, 21--27 Jul 2024.
\newblock URL \url{https://proceedings.mlr.press/v235/nawrot24a.html}.

\bibitem[NVIDIA et~al.(2025)NVIDIA, Basant, Khairnar, Paithankar, Khattar,
  Renduchintala, Malte, Bercovich, Hazare, Rico, Ficek, Kondratenko,
  Shaposhnikov, Bukharin, Taghibakhshi, Barton, Mahabaleshwarkar, Shen, Tao,
  Guan, Shors, Mandarwal, Mehta, Venkatesan, Sharabiani, Aithal, Poojary,
  Dattagupta, Buddharaju, Zhu, Simkin, Kartal, Rouhani, Chen, Ginsburg, Norick,
  Yu, Catanzaro, Wang, Truong, Mungekar, Patel, Alexiuk, Munley, Parisien, Su,
  Afrimi, Korzekwa, Rohrer, Gitman, Mosallanezhad, Narayanan, Rekesh, Yared,
  Pykhtar, Ahn, Riach, Long, Ning, Chung, Galinkin, Bakhturina, Prasad, Shen,
  Qian, Elisha, Sharma, Ross, Ngo, Sahota, Wang, Shin, Huang, Cunningham,
  Gitman, Moshkov, Jung, Kautz, Scowcroft, Casper, Zhang, Zeng, Zhang, Xue,
  Huang, Conway, Kamalu, Cohen, Jennings, Vialard, Yi, Parmar, Briski, Cheung,
  Luna, Wyss, Santhanam, Kong, Pawelec, Anik, Li, Ahmadian, McAfee, Sleiman,
  Derczynski, Vega, de~Melo, Sreedhar, Chochowski, Cai, Kliegl,
  Stepniewska-Dziubinska, Novikov, Samadi, Price, Boubdir, Boone, Evans, Bien,
  Zawalski, Martinez, Chrzanowski, Shoeybi, Patwary, Dhameja, Assaf, Habibi,
  Bhatia, Pope, Tajbakhsh, Juluru, Rybakov, Hrinchuk, Kuchaiev, Olabiyi,
  Ribalta, Subramanian, Chadha, Molchanov, Dykas, Jin, Bialecki, Januszewski,
  Thalasta, Gaikwad, Varshney, Gundecha, Tredak, Mahabadi, Patel, El-Yaniv,
  Rajan, Cheruvu, Shahbazyan, Borkar, Gala, Waleffe, Zhang, Hewett, Prenger,
  Jain, Kriman, Satheesh, Kaji, Yurick, Muralidharan, Narenthiran, Bak, Sameni,
  Han, Ramasamy, Ghosh, Sreenivas, Thomas, Diao, Gopal, Prabhumoye, Toshniwal,
  Ding, Singh, Jain, Majumdar, Singhal, Alborghetti, Akter, Kong, Moon,
  Hliwiak, Asida, Wang, Konuk, Vashishth, Poon, Karpas, Noroozi, Srinivasan,
  Korthikanti, Fugro, Kalluru, Kurin, Lavrukhin, Ahmad, Du, Byeon, Lu, Dong,
  Karnati, Choi, Zhang, Lin, Fu, Suhara, Dong, Li, Zhu, and
  Chen]{nvidia2025nvidianemotronnano2}
NVIDIA, Aarti Basant, Abhijit Khairnar, Abhijit Paithankar, Abhinav Khattar,
  Adithya Renduchintala, Aditya Malte, Akhiad Bercovich, Akshay Hazare,
  Alejandra Rico, Aleksander Ficek, Alex Kondratenko, Alex Shaposhnikov,
  Alexander Bukharin, Ali Taghibakhshi, Amelia Barton, Ameya~Sunil
  Mahabaleshwarkar, Amy Shen, Andrew Tao, Ann Guan, Anna Shors, Anubhav
  Mandarwal, Arham Mehta, Arun Venkatesan, Ashton Sharabiani, Ashwath Aithal,
  Ashwin Poojary, Ayush Dattagupta, Balaram Buddharaju, Banghua Zhu, Barnaby
  Simkin, Bilal Kartal, Bita~Darvish Rouhani, Bobby Chen, Boris Ginsburg,
  Brandon Norick, Brian Yu, Bryan Catanzaro, Charles Wang, Charlie Truong,
  Chetan Mungekar, Chintan Patel, Chris Alexiuk, Christian Munley, Christopher
  Parisien, Dan Su, Daniel Afrimi, Daniel Korzekwa, Daniel Rohrer, Daria
  Gitman, David Mosallanezhad, Deepak Narayanan, Dima Rekesh, Dina Yared,
  Dmytro Pykhtar, Dong Ahn, Duncan Riach, Eileen Long, Elliott Ning, Eric
  Chung, Erick Galinkin, Evelina Bakhturina, Gargi Prasad, Gerald Shen, Haifeng
  Qian, Haim Elisha, Harsh Sharma, Hayley Ross, Helen Ngo, Herman Sahota, Hexin
  Wang, Hoo~Chang Shin, Hua Huang, Iain Cunningham, Igor Gitman, Ivan Moshkov,
  Jaehun Jung, Jan Kautz, Jane~Polak Scowcroft, Jared Casper, Jian Zhang, Jiaqi
  Zeng, Jimmy Zhang, Jinze Xue, Jocelyn Huang, Joey Conway, John Kamalu,
  Jonathan Cohen, Joseph Jennings, Julien~Veron Vialard, Junkeun Yi, Jupinder
  Parmar, Kari Briski, Katherine Cheung, Katherine Luna, Keith Wyss, Keshav
  Santhanam, Kezhi Kong, Krzysztof Pawelec, Kumar Anik, Kunlun Li, Kushan
  Ahmadian, Lawrence McAfee, Laya Sleiman, Leon Derczynski, Luis Vega,
  Maer~Rodrigues de~Melo, Makesh~Narsimhan Sreedhar, Marcin Chochowski, Mark
  Cai, Markus Kliegl, Marta Stepniewska-Dziubinska, Matvei Novikov, Mehrzad
  Samadi, Meredith Price, Meriem Boubdir, Michael Boone, Michael Evans, Michal
  Bien, Michal Zawalski, Miguel Martinez, Mike Chrzanowski, Mohammad Shoeybi,
  Mostofa Patwary, Namit Dhameja, Nave Assaf, Negar Habibi, Nidhi Bhatia, Nikki
  Pope, Nima Tajbakhsh, Nirmal~Kumar Juluru, Oleg Rybakov, Oleksii Hrinchuk,
  Oleksii Kuchaiev, Oluwatobi Olabiyi, Pablo Ribalta, Padmavathy Subramanian,
  Parth Chadha, Pavlo Molchanov, Peter Dykas, Peter Jin, Piotr Bialecki, Piotr
  Januszewski, Pradeep Thalasta, Prashant Gaikwad, Prasoon Varshney, Pritam
  Gundecha, Przemek Tredak, Rabeeh~Karimi Mahabadi, Rajen Patel, Ran El-Yaniv,
  Ranjit Rajan, Ria Cheruvu, Rima Shahbazyan, Ritika Borkar, Ritu Gala, Roger
  Waleffe, Ruoxi Zhang, Russell~J. Hewett, Ryan Prenger, Sahil Jain, Samuel
  Kriman, Sanjeev Satheesh, Saori Kaji, Sarah Yurick, Saurav Muralidharan, Sean
  Narenthiran, Seonmyeong Bak, Sepehr Sameni, Seungju Han, Shanmugam Ramasamy,
  Shaona Ghosh, Sharath~Turuvekere Sreenivas, Shelby Thomas, Shizhe Diao,
  Shreya Gopal, Shrimai Prabhumoye, Shubham Toshniwal, Shuoyang Ding, Siddharth
  Singh, Siddhartha Jain, Somshubra Majumdar, Soumye Singhal, Stefania
  Alborghetti, Syeda~Nahida Akter, Terry Kong, Tim Moon, Tomasz Hliwiak, Tomer
  Asida, Tony Wang, Tugrul Konuk, Twinkle Vashishth, Tyler Poon, Udi Karpas,
  Vahid Noroozi, Venkat Srinivasan, Vijay Korthikanti, Vikram Fugro, Vineeth
  Kalluru, Vitaly Kurin, Vitaly Lavrukhin, Wasi~Uddin Ahmad, Wei Du, Wonmin
  Byeon, Ximing Lu, Xin Dong, Yashaswi Karnati, Yejin Choi, Yian Zhang, Ying
  Lin, Yonggan Fu, Yoshi Suhara, Zhen Dong, Zhiyu Li, Zhongbo Zhu, and Zijia
  Chen.
\newblock Nvidia nemotron nano 2: An accurate and efficient hybrid
  mamba-transformer reasoning model, 2025.
\newblock URL \url{https://arxiv.org/abs/2508.14444}.

\bibitem[{OLMo Team} et~al.(2025){OLMo Team}, Ettinger, Bertsch, Kuehl, Graham,
  Heineman, Groeneveld, Brahman, Timbers, Ivison, Morrison, Poznanski, Lo,
  Soldaini, Jordan, Chen, Noukhovitch, Lambert, Walsh, Dasigi, Berry, Malik,
  Shah, Geng, Arora, Gupta, Anderson, Xiao, Murray, Romero, Graf, Asai, Bhagia,
  Wettig, Liu, Rangapur, Anastasiades, Huang, Schwenk, Trivedi, Magnusson,
  Lochner, Liu, Miranda, Sap, Morgan, Schmitz, Guerquin, Wilson, Huff, Bras,
  Xin, Shao, Skjonsberg, Shen, Li, Wilde, Pyatkin, Merrill, Chang, Gu, Zeng,
  Sabharwal, Zettlemoyer, Koh, Farhadi, Smith, and Hajishirzi]{olmo2025olmo3}
{OLMo Team}, Allyson Ettinger, Amanda Bertsch, Bailey Kuehl, David Graham,
  David Heineman, Dirk Groeneveld, Faeze Brahman, Finbarr Timbers, Hamish
  Ivison, Jacob Morrison, Jake Poznanski, Kyle Lo, Luca Soldaini, Matt Jordan,
  Mayee Chen, Michael Noukhovitch, Nathan Lambert, Pete Walsh, Pradeep Dasigi,
  Robert Berry, Saumya Malik, Saurabh Shah, Scott Geng, Shane Arora, Shashank
  Gupta, Taira Anderson, Teng Xiao, Tyler Murray, Tyler Romero, Victoria Graf,
  Akari Asai, Akshita Bhagia, Alexander Wettig, Alisa Liu, Aman Rangapur, Chloe
  Anastasiades, Costa Huang, Dustin Schwenk, Harsh Trivedi, Ian Magnusson,
  Jaron Lochner, Jiacheng Liu, Lester James~V. Miranda, Maarten Sap, Malia
  Morgan, Michael Schmitz, Michal Guerquin, Michael Wilson, Regan Huff,
  Ronan~Le Bras, Rui Xin, Rulin Shao, Sam Skjonsberg, Shannon~Zejiang Shen,
  Shuyue~Stella Li, Tucker Wilde, Valentina Pyatkin, Will Merrill, Yapei Chang,
  Yuling Gu, Zhiyuan Zeng, Ashish Sabharwal, Luke Zettlemoyer, Pang~Wei Koh,
  Ali Farhadi, Noah~A. Smith, and Hannaneh Hajishirzi.
\newblock Olmo 3, 2025.
\newblock URL \url{https://arxiv.org/abs/2512.13961}.

\bibitem[{OpenAI}(2025)]{codex}
{OpenAI}.
\newblock Introducing codex, May 2025.
\newblock URL \url{https://openai.com/index/introducing-codex/}.
\newblock Accessed: 2026-01-09.

\bibitem[Packer et~al.(2023)Packer, Wooders, Lin, Fang, Patil, Stoica, and
  Gonzalez]{packer2023memgpt}
Charles Packer, Sarah Wooders, Kevin Lin, Vivian Fang, Shishir~G. Patil, Ion
  Stoica, and Joseph~E. Gonzalez.
\newblock Memgpt: Towards {LLM}s as operating systems, 2023.
\newblock URL \url{https://arxiv.org/abs/2310.08560}.

\bibitem[Penedo(2025)]{penedo2025finewiki}
Guilherme Penedo.
\newblock Finewiki, 2025.
\newblock URL \url{https://huggingface.co/datasets/HuggingFaceFW/finewiki}.
\newblock Source: Wikimedia Enterprise Snapshot API
  (https://api.enterprise.wikimedia.com/v2/snapshots). Text licensed under CC
  BY-SA 4.0 with attribution to Wikipedia contributors.

\bibitem[Peng et~al.(2023)Peng, Quesnelle, Fan, and Shippole]{yarn}
Bowen Peng, Jeffrey Quesnelle, Honglu Fan, and Enrico Shippole.
\newblock Yarn: Efficient context window extension of large language models.
\newblock \emph{arXiv preprint arXiv:2309.00071}, 2023.

\bibitem[Pilchen et~al.(2025)Pilchen, Grave, and P{\'e}rez]{arc-encoder}
Hippolyte Pilchen, Edouard Grave, and Patrick P{\'e}rez.
\newblock Arc-encoder: learning compressed text representations for large
  language models.
\newblock \emph{arXiv preprint arXiv:2510.20535}, 2025.

\bibitem[Rae et~al.(2019)Rae, Potapenko, Jayakumar, and
  Lillicrap]{rae2019compressive}
Jack~W Rae, Anna Potapenko, Siddhant~M Jayakumar, and Timothy~P Lillicrap.
\newblock Compressive transformers for long-range sequence modelling.
\newblock \emph{arXiv preprint arXiv:1911.05507}, 2019.

\bibitem[Raffel et~al.(2020)Raffel, Shazeer, Roberts, Lee, Narang, Matena,
  Zhou, Li, and Liu]{raffel2020exploring}
Colin Raffel, Noam Shazeer, Adam Roberts, Katherine Lee, Sharan Narang, Michael
  Matena, Yanqi Zhou, Wei Li, and Peter~J Liu.
\newblock Exploring the limits of transfer learning with a unified text-to-text
  transformer.
\newblock \emph{Journal of machine learning research}, 21\penalty0
  (140):\penalty0 1--67, 2020.

\bibitem[Raoof et~al.(2026)Raoof, Zhuang, Nezhurina, Guha, Tejaswi, Marten,
  Ruan, Griggs, Shaw, Bansal, et~al.]{raoof2026openthoughts}
Negin Raoof, Richard Zhuang, Marianna Nezhurina, Etash Guha, Atula Tejaswi,
  Ryan Marten, Charlie~F Ruan, Tyler Griggs, Alexander~Glenn Shaw, Hritik
  Bansal, et~al.
\newblock {OpenThoughts-Agent}: Data recipes for agentic models.
\newblock \emph{arXiv preprint arXiv:2606.24855}, 2026.

\bibitem[Su et~al.(2024)Su, Ahmed, Lu, Pan, Bo, and Liu]{rope}
Jianlin Su, Murtadha Ahmed, Yu~Lu, Shengfeng Pan, Wen Bo, and Yunfeng Liu.
\newblock Roformer: Enhanced transformer with rotary position embedding.
\newblock \emph{Neurocomputing}, 568:\penalty0 127063, 2024.

\bibitem[Synk et~al.(2025)Synk, Hoover, Kirchenbauer, Jain, Stein, Shu,
  Sanchez, Duraiswami, and Goldstein]{synk2025exploiting}
Ryan Synk, Monte Hoover, John Kirchenbauer, Neel Jain, Alex Stein, Manli Shu,
  Josue~Melendez Sanchez, Ramani Duraiswami, and Tom Goldstein.
\newblock Exploiting sparsity for long context inference: Million token
  contexts on commodity gpus.
\newblock \emph{arXiv preprint arXiv:2502.06766}, 2025.

\bibitem[Tan et~al.(2024)Tan, Li, Patil, Wu, Zhang, Keutzer, Gonzalez, and
  Popa]{lloco}
Sijun Tan, Xiuyu Li, Shishir~G Patil, Ziyang Wu, Tianjun Zhang, Kurt Keutzer,
  Joseph~E Gonzalez, and Raluca~Ada Popa.
\newblock Lloco: Learning long contexts offline.
\newblock In \emph{Proceedings of the 2024 Conference on Empirical Methods in
  Natural Language Processing}, pages 17605--17621, 2024.

\bibitem[Tang et~al.(2025)Tang, Zhang, Wu, Ye, Bai, Wang, Lu, Chen, Hai, Zheng,
  et~al.]{GMSA}
Jiwei Tang, Zhicheng Zhang, Shunlong Wu, Jingheng Ye, Lichen Bai, Zitai Wang,
  Tingwei Lu, Jiaqi Chen, Lin Hai, Hai-Tao Zheng, et~al.
\newblock Gmsa: Enhancing context compression via group merging and layer
  semantic alignment.
\newblock \emph{arXiv preprint arXiv:2505.12215}, 2025.

\bibitem[Tong et~al.(2024)Tong, Brown, Wu, Woo, Middepogu, Akula, Yang, Yang,
  Iyer, Pan, et~al.]{tong2024cambrian}
Shengbang Tong, Ellis Brown, Penghao Wu, Sanghyun Woo, Manoj Middepogu, Sai~C
  Akula, Jihan Yang, Shusheng Yang, Adithya Iyer, Xichen Pan, et~al.
\newblock Cambrian-1: A fully open, vision-centric exploration of multimodal
  llms.
\newblock \emph{Advances in Neural Information Processing Systems},
  37:\penalty0 87310--87356, 2024.

\bibitem[Wang et~al.(2023)Wang, Liang, Yang, Huang, Wu, Wu, Lu, Ma, and
  Li]{wang2023scm}
Bing Wang, Xinnian Liang, Jian Yang, Hui Huang, Shuangzhi Wu, Peihao Wu, Lu~Lu,
  Zejun Ma, and Zhoujun Li.
\newblock Enhancing large language model with self-controlled memory framework,
  2023.
\newblock URL \url{https://arxiv.org/abs/2304.13343}.

\bibitem[Weber et~al.(2024)Weber, Fu, Anthony, Oren, Adams, Alexandrov, Lyu,
  Nguyen, Yao, Adams, Athiwaratkun, Chalamala, Chen, Ryabinin, Dao, Liang, Ré,
  Rish, and Zhang]{together2023redpajama}
Maurice Weber, Daniel~Y. Fu, Quentin Anthony, Yonatan Oren, Shane Adams, Anton
  Alexandrov, Xiaozhong Lyu, Huu Nguyen, Xiaozhe Yao, Virginia Adams, Ben
  Athiwaratkun, Rahul Chalamala, Kezhen Chen, Max Ryabinin, Tri Dao, Percy
  Liang, Christopher Ré, Irina Rish, and Ce~Zhang.
\newblock Redpajama: an open dataset for training large language models.
\newblock \emph{NeurIPS Datasets and Benchmarks Track}, 2024.

\bibitem[Wolf et~al.(2019)Wolf, Debut, Sanh, Chaumond, Delangue, Moi, Cistac,
  Rault, Louf, Funtowicz, et~al.]{wolf2019huggingface}
Thomas Wolf, Lysandre Debut, Victor Sanh, Julien Chaumond, Clement Delangue,
  Anthony Moi, Pierric Cistac, Tim Rault, R{\'e}mi Louf, Morgan Funtowicz,
  et~al.
\newblock {HuggingFace's Transformers}: State-of-the-art natural language
  processing.
\newblock \emph{arXiv preprint arXiv:1910.03771}, 2019.

\bibitem[Xie et~al.(2025)Xie, Li, Gao, Du, Lam, Zou, and Chen]{xie2025swe}
Chengxing Xie, Bowen Li, Chang Gao, He~Du, Wai Lam, Difan Zou, and Kai Chen.
\newblock Swe-fixer: Training open-source llms for effective and efficient
  github issue resolution.
\newblock \emph{arXiv preprint arXiv:2501.05040}, 2025.

\bibitem[Xu et~al.(2026)Xu, Lin, Xue, Wang, Xu, Wu, Zhang, Lin, Dong, Ling,
  et~al.]{xu2026deepseek}
Anyi Xu, Bangcai Lin, Bing Xue, Bingxuan Wang, Bingzheng Xu, Bochao Wu, Bowei
  Zhang, Chaofan Lin, Chen Dong, Chenchen Ling, et~al.
\newblock {DeepSeek-V4}: Towards highly efficient million-token context
  intelligence.
\newblock \emph{arXiv preprint arXiv:2606.19348}, 2026.

\bibitem[Xu et~al.(2024)Xu, Ping, Wu, Xu, Liu, Shoeybi, and
  Catanzaro]{xu2024chatqa}
Peng Xu, Wei Ping, Xianchao Wu, Chejian Xu, Zihan Liu, Mohammad Shoeybi, and
  Bryan Catanzaro.
\newblock Chatqa 2: Bridging the gap to proprietary llms in long context and
  rag capabilities.
\newblock \emph{arXiv preprint arXiv:2407.14482}, 2024.

\bibitem[Xu et~al.(2025)Xu, Liang, Mei, Gao, Tan, and Zhang]{xu2025amem}
Wujiang Xu, Zujie Liang, Kai Mei, Hang Gao, Juntao Tan, and Yongfeng Zhang.
\newblock A-mem: Agentic memory for {LLM} agents, 2025.
\newblock URL \url{https://arxiv.org/abs/2502.12110}.

\bibitem[Yang et~al.(2025)Yang, Li, Yang, Zhang, Hui, Zheng, Yu, Gao, Huang,
  Lv, et~al.]{qwen3}
An~Yang, Anfeng Li, Baosong Yang, Beichen Zhang, Binyuan Hui, Bo~Zheng, Bowen
  Yu, Chang Gao, Chengen Huang, Chenxu Lv, et~al.
\newblock Qwen3 technical report.
\newblock \emph{arXiv preprint arXiv:2505.09388}, 2025.

\bibitem[Yang et~al.(2018)Yang, Qi, Zhang, Bengio, Cohen, Salakhutdinov, and
  Manning]{yang2018hotpotqa}
Zhilin Yang, Peng Qi, Saizheng Zhang, Yoshua Bengio, William Cohen, Ruslan
  Salakhutdinov, and Christopher~D Manning.
\newblock Hotpotqa: A dataset for diverse, explainable multi-hop question
  answering.
\newblock In \emph{Proceedings of the 2018 conference on empirical methods in
  natural language processing}, pages 2369--2380, 2018.

\bibitem[Yen et~al.(2024)Yen, Gao, and Chen]{cepe}
Howard Yen, Tianyu Gao, and Danqi Chen.
\newblock Long-context language modeling with parallel context encoding.
\newblock In \emph{Proceedings of the 62nd Annual Meeting of the Association
  for Computational Linguistics (Volume 1: Long Papers)}, pages 2588--2610,
  2024.

\bibitem[Yoon et~al.(2024)Yoon, Lee, Hwang, Jeong, and Kang]{yoon2024compact}
Chanwoong Yoon, Taewhoo Lee, Hyeon Hwang, Minbyul Jeong, and Jaewoo Kang.
\newblock Compact: Compressing retrieved documents actively for question
  answering.
\newblock \emph{arXiv preprint arXiv:2407.09014}, 2024.

\bibitem[Zhang et~al.(2025{\natexlab{a}})Zhang, Kraska, and
  Khattab]{zhang2025rlm}
Alex~L. Zhang, Tim Kraska, and Omar Khattab.
\newblock Recursive language models, 2025{\natexlab{a}}.
\newblock URL \url{https://arxiv.org/abs/2512.24601}.

\bibitem[Zhang and Sennrich(2019)]{zhang2019root}
Biao Zhang and Rico Sennrich.
\newblock Root mean square layer normalization.
\newblock \emph{Advances in neural information processing systems}, 32, 2019.

\bibitem[Zhang et~al.(2025{\natexlab{b}})Zhang, Li, Long, Zhang, Lin, Yang,
  Xie, Yang, Liu, Lin, et~al.]{qwen3-embedding}
Yanzhao Zhang, Mingxin Li, Dingkun Long, Xin Zhang, Huan Lin, Baosong Yang,
  Pengjun Xie, An~Yang, Dayiheng Liu, Junyang Lin, et~al.
\newblock Qwen3 embedding: Advancing text embedding and reranking through
  foundation models.
\newblock \emph{arXiv preprint arXiv:2506.05176}, 2025{\natexlab{b}}.

\bibitem[Zhao et~al.(2024)Zhao, Ren, Hessel, Cardie, Choi, and
  Deng]{zhao2024wildchat}
Wenting Zhao, Xiang Ren, Jack Hessel, Claire Cardie, Yejin Choi, and Yuntian
  Deng.
\newblock Wildchat: 1m chatgpt interaction logs in the wild.
\newblock \emph{arXiv preprint arXiv:2405.01470}, 2024.

\bibitem[Zheng et~al.(2023)Zheng, Chiang, Sheng, Li, Zhuang, Wu, Zhuang, Li,
  Lin, Xing, et~al.]{zheng2023lmsys}
Lianmin Zheng, Wei-Lin Chiang, Ying Sheng, Tianle Li, Siyuan Zhuang, Zhanghao
  Wu, Yonghao Zhuang, Zhuohan Li, Zi~Lin, Eric~P Xing, et~al.
\newblock Lmsys-chat-1m: A large-scale real-world llm conversation dataset.
\newblock \emph{arXiv preprint arXiv:2309.11998}, 2023.

\bibitem[Zheng et~al.(2024)Zheng, Yin, Xie, Sun, Huang, Yu, Cao, Kozyrakis,
  Stoica, Gonzalez, et~al.]{zheng2024sglang}
Lianmin Zheng, Liangsheng Yin, Zhiqiang Xie, Chuyue~Livia Sun, Jeff Huang,
  Cody~Hao Yu, Shiyi Cao, Christos Kozyrakis, Ion Stoica, Joseph~E Gonzalez,
  et~al.
\newblock Sglang: Efficient execution of structured language model programs.
\newblock \emph{Advances in neural information processing systems},
  37:\penalty0 62557--62583, 2024.

\bibitem[Zhong et~al.(2023)Zhong, Guo, Gao, Ye, and Wang]{zhong2023memorybank}
Wanjun Zhong, Lianghong Guo, Qiqi Gao, He~Ye, and Yanlin Wang.
\newblock Memorybank: Enhancing large language models with long-term memory,
  2023.
\newblock URL \url{https://arxiv.org/abs/2305.10250}.

\bibitem[Zweiger et~al.(2026)Zweiger, Fu, Guo, and Kim]{zweiger2026fast}
Adam Zweiger, Xinghong Fu, Han Guo, and Yoon Kim.
\newblock Fast kv compaction via attention matching.
\newblock \emph{arXiv preprint arXiv:2602.16284}, 2026.

\end{thebibliography}
\bibliographystyle{plainnat}

\newpage
\appendix
\onecolumn
\setcounter{topnumber}{2}
\crefalias{section}{appendix}
\crefalias{subsection}{appendix}
\crefalias{subsubsection}{appendix}
\crefname{appendix}{Appendix}{Appendices}
\Crefname{appendix}{Appendix}{Appendices}
\etocdepthtag.toc{mtappendix}
\etocsettagdepth{mtmain}{none}
\etocsettagdepth{mtappendix}{subsection}
\renewcommand{\contentsname}{Appendix Contents}
\tableofcontents
\newpage
\section{Extended Related Work}
\label{app-sec:extended-rel-work}

\paragraph{Hard-token compression.}
Hard-token methods achieve compression by deleting or rewriting input tokens.
Token pruning approaches remove tokens via importance heuristics \citep{li2023compressing,jiang2023llmlingua,jiang2024longllmlingua}, while summarization and rewriting methods compress by paraphrasing into fewer tokens \citep{chuang2024learning,yoon2024compact}.
While widely used in practice \citep[e.g., context compaction in Claude Code and Codex; ][]{claude_code, codex} and very effective for reducing prompt length, these methods are intrinsically lossy: once information is dropped or paraphrased in the discrete token space, exact lexical and structural detail is unrecoverable.
This limits their ability to support code fidelity, format-sensitive reasoning, and lookup-style in-context learning in long-horizon settings.

Relatedly, many agent systems externalize memory via summarization, retrieval, or structured stores.
Examples include hierarchical memory management \citep{packer2023memgpt}, long-term conversational memory \citep{zhong2023memorybank,wang2023scm}, and agentic memory frameworks \citep{xu2025amem}.
While effective in extending interaction horizons, these approaches typically rely on lossy summaries or task-specific memory representations that are external to the base model.

\paragraph{KV cache compression.}
KV cache compression reduces the size of the KV cache by evicting entries; most methods rely on a hand-crafted or learned heuristic to select which entries to drop.
SnapKV \citep{li2024snapkv} compresses the cache by removing entries with low aggregated query attention on a per-head basis.
SnapKV can be either prompt-agnostic, where pruning is performed without knowledge of the prompt, or prompt-dependent, where pruning is performed with knowledge of the prompt.
Prompt-dependent methods have the downside of requiring labeling of the context and prompt for inputs, and pruning is highly specific to each individual input, limiting multi-turn chat applicability \citep{li2024scbench}.
To overcome these limitations, self-study \citep{cartridges,zweiger2026fast} can be used to generate synthetic query signals from the context itself, enabling prompt-agnostic pruning without requiring labeled prompt-context pairs.

KVzip \citep{kim2025kvzip} performs query-agnostic KV cache compression by using teacher-forced context reconstruction as a proxy objective, assigning each KV pair an importance score from the maximum attention it receives during reconstruction, and pruning low-scoring entries.
Fast KV Compaction via Attention Matching \citep{zweiger2026fast} compresses the KV cache by fitting a smaller set of keys and values, together with per-entry bias terms, that match the original cache's attention outputs and attention mass on reference queries.
Expected Attention \citep{devoto2025expected} predicts KV importance by approximating future queries with a Gaussian distribution, computing each key's expected attention in closed form, and pruning entries with the lowest expected contribution.
There are many other KV cache compression techniques and we refer the reader to Awesome KV cache Compression GitHub \citep{October2001_AwesomeKVCacheCompression_2025} for a comprehensive review.

We note that KV cache compression methods also pool before eviction: both \citet{li2024snapkv} and \citet{zweiger2026fast} use max pooling, as does \citet{devoto2025expected}. \citet{li2024snapkv} further note that the choice of pooling operator does not significantly impact their performance.

It is also possible to do targeted KV cache compression offline for specific prompts.
This approach applies to the setting where it is logical to expend more compute to compress the cache once, as this cost will be amortized over many user queries.
\citet{synk2025exploiting} compute the full KV cache before applying top-k cache compression and reuse the smaller cache.
\citet{cartridges} use self-study to learn small KV caches per corpus.

\paragraph{Soft-token compression.}
A complementary direction encodes long contexts into fewer learned continuous tokens.
The most successful examples of soft-token approaches to compression require training on the target context, such as prefix tuning \citep{li2021prefix, lloco}.
These methods can achieve comparable in-context performance to the base model with large compression ratios for the specific contexts they are trained on.
Other soft-token approaches that do not require training on individual contexts are gist tokens \citep{gist}, in-context autoencoders \citep{in-context-autoencoder} and encoder-decoder compression frameworks \citep{auto-compressor,e2llm,dai2025pretraining,li2025500xcompressor}.

\paragraph{Efficient long sequence modeling.}
A complementary line of work targets the cost of long-context inference by modifying the underlying sequence model, the attention mechanism, or the positional encoding, rather than compressing the input.
Early work scales Transformers to long inputs through segment-level recurrence and external memory: Transformer-XL~\citep{transformerxl} caches hidden states from previous segments to extend the effective receptive field, and the Compressive Transformer~\citep{rae2019compressive} further compresses older memory entries into a smaller compressed memory.

A second direction replaces softmax attention with formulations that scale linearly in sequence length. Linear-attention Transformers~\citep{linearattention} cast attention as a kernel feature map to obtain $O(T)$ inference, state-space models such as S4~\citep{s4} and Mamba~\citep{mamba} use recurrent or selective-scan formulations with constant-size hidden state, and more recent hybrids such as Kimi Linear~\citep{kimilinear} interleave linear-attention blocks with sparse full-attention blocks to recover long-range recall while keeping most of the per-layer cost linear. These approaches lower the asymptotic cost of attention but typically trade off exact long-range recall on retrieval-style tasks.

Within the Transformer family, recent DeepSeek models compress the attention computation itself: Multi-head Latent Attention in DeepSeek-V2~\citep{deepseekv2} projects keys and values into a low-rank latent KV cache, and DeepSeek-V4~\citep{xu2026deepseek} introduces a hybrid attention architecture that interleaves Compressed Sparse Attention (CSA), which consolidates every $m$ tokens of the KV cache into a single entry and then applies sparse top-$k$ selection over those entries, with Heavily Compressed Attention (HCA), which uses a much larger compression rate $m' \gg m$ but retains dense attention over the resulting entries, jointly reducing both KV memory and per-token attention FLOPs at million-token context lengths.

A separate thread extends the usable context of pre-trained models through their positional encoding: rotary position embeddings~\citep{rope} provide relative attention with smooth extrapolation, and YaRN~\citep{yarn} rescales RoPE frequencies to extend the context window with minimal additional training.

Finally, parallel context encoding methods such as CEPE~\citep{cepe} take a different route: a small encoder produces dense representations of long-context chunks in parallel, and a frozen decoder consumes them through inserted cross-attention layers, exposing long contexts to the decoder without full self-attention over the raw tokens.

These approaches reduce the per-token cost of attention, the cost of positional indexing, or the effective sequence length seen by self-attention, rather than the number of tokens reaching the decoder; they are largely orthogonal to LCLMs and can in principle be composed.

\section{Dataset Details}
\label{app-sec:data}

\subsection{Continual Pre-training Dataset Curation}\label{app-subsec:cont-pre-train-data}
In early experiments, we only train on reconstruction data and hope the model can learn generalizable patterns. However, we find that training on such data alone allows the model to reconstruct the full sequence with negligible loss, but it generalizes poorly beyond reconstruction-style tasks. In fact, it cannot perform any other tasks, even when the LLM decoder is kept frozen. This might not seem surprising, as we are essentially performing a general version of prefix tuning. Thus, the learned representation, though general for different contexts, collapses to the task of reconstruction. On the other hand, training solely on next-token prediction enables the model to leverage compressed context for downstream generation and reasoning. However, the resulting representations often lack fine-grained fidelity (e.g., exact-string details), leading to weaker reconstruction and brittleness on tasks requiring precise recovery. 

Motivated by this trade-off, we train on a mixture of next-token prediction and reconstruction data. Our working hypothesis is that next-token prediction provides task-aligned learning signals that teach the decoder to operate on compressed context, while reconstruction acts as an auxiliary objective that encourages information preservation and can accelerate early-stage representation learning. Theoretically, sufficiently large-scale next-token training alone may recover fine-grained fidelity, but we find that in practice, the reconstruction data serves as an auxiliary data source that accelerates the training of the general compressor.

\begin{table*}[ht]
\centering
\small
\setlength{\tabcolsep}{5pt}
\renewcommand{\arraystretch}{1.12}
\caption{
Dataset mixture for continual pre-training. Pre/Post-Comp are token counts before/after compression.
We refer to the unique Nvidia license as ``Nvidia''; the license can be found at \url{https://huggingface.co/datasets/nvidia/Nemotron-CC-v2.1/blob/main/LICENSE.md}.
}
\label{app-tab:data-mixture-pre-training}
\resizebox{\linewidth}{!}{
\begin{tabular}{l l l r r r r r}
\toprule
\multirow{2}{*}{\textbf{Data Name}} &
\multirow{2}{*}{\textbf{License}} &
\multirow{2}{*}{\textbf{Data Description}} &
\multirow{2}{*}{\textbf{Num Samples}} &
\multicolumn{3}{c}{\textbf{Tokens}} &
\multirow{2}{*}{\textbf{\%}} \\
\cmidrule(lr){5-7}
& & & & \textbf{Pre-Comp} & \textbf{Post-Comp} & \textbf{Trainable} & \\
\midrule
Nemotron CC & Nvidia & Continual pre-training (CC Text)      & 84.84M  & 60.69B & 30.38B & 30.08B & 20.37 \\
Nemotron Code & Nvidia & Continual pre-training (Code)         & 46.85M  & 51.18B & 25.21B & 25.01B & 16.90 \\
Nemotron Reasoning & Nvidia & Continual pre-training (Reasoning)    & 16.62M  & 52.32B & 25.61B & 25.50B & 17.17 \\
Longmino & odc-by & Continual pre-training (Long-Context) & 0.548M  & 26.62B & 4.24B  & 4.23B  & 2.84  \\
SFT & cc-by-4.0 & SFT with Compressed Prompt           & 20.67M  & 42.21B & 38.02B & 37.54B & 25.49 \\
Reconstruction & Nvidia & Reconstruction from Compression      & 23.53M  & 50.75B & 25.69B & 25.18B & 17.23 \\
\midrule
\textbf{Total} & -- & & \textbf{193.06M} & \textbf{283.78B} & \textbf{149.16B} & \textbf{147.54B} & \textbf{100.00} \\
\bottomrule
\end{tabular}
}
\end{table*}

\paragraph{Curation.}
Our primary sources are NVIDIA's Nemotron pre-training datasets \citep{nvidia2025nvidianemotronnano2, nvidia_nemotron_nano_v3_2025}. We sample 50M data points from Nemotron's pre-training code and cc code, 50M examples from high-quality and synthetic high-quality cc text, and 50M examples from Nemotron specialized SFT mixtures that cover diverse reasoning traces.
A substantial portion of the synthetic data from Nemotron consists of rewrites or generations produced by strong teacher models (e.g., \texttt{Qwen3-30B-A3B}, \texttt{Qwen3-235B-A22B}~\citep{qwen3}, \texttt{GPT-oss-120B}~\citep{agarwal2025gpt}, and \texttt{DeepSeek-R1}~\citep{guo2025deepseek}), which alleviate the problem of forgetting. Furthermore, to expose the compressor to longer documents, we sample approximately 500K examples from the OLMo-3 longmino collection \citep{olmo2025olmo3}. Finally, to better preserve instruction-following and chat-template behavior, we also include Nemotron pre-training SFT data and convert it into a standardized instruction tuning and multi-turn chat format. Exact dataset composition is reported in \Cref{app-tab:data-mixture-pre-training}.

\subsection{Auxiliary Reconstruction Dataset Curation}
\label{app-subsec:aux-recon-data}
We construct reconstruction data spanning three broad categories: text, code, and \emph{\LaTeX}.
Text is sampled from Nemotron CC / math, FineWiki \citep{penedo2025finewiki}, and RedPajama ArXiv \citep{together2023redpajama}; code is sampled from RedPajama GitHub, The Stack v2 (we concatenate files from the repo into one document) \citep{lozhkov2024starcoder}, and CodeParrot; and \LaTeX\ is sampled from TexTeller-OCR. Our goal is to maximize data diversity so that the compressor retains sufficient coverage for downstream tasks that may involve any of these modalities.

\subsection{SFT Dataset Curation}\label{app-subsec:posttrain-data}
For reasoning, we curate math, code, and science supervision from the Nemotron SFT datasets and OLMo-3 Dolci-Think \citep{NemotronPostTrainingDatasetV2, NemotronPostTrainingDatasetV1, nvidia_nemotron_nano_v3_2025, olmo2025olmo3}. Unlike the other SFT data, where the prompt or document is compressed, reasoning examples keep the question uncompressed and compress spans of the reference chain of thought; the model is trained to continue reasoning from the compressed spans. About half of the reasoning examples (2.70M) contain compressed chain-of-thought spans, and the rest are left uncompressed. 

For long-context instruction tuning, we combine long-context and RAG-style QA datasets \citep{he2025scaling, jin2019pubmedqa, liu2024chatqa, xu2024chatqa, yang2018hotpotqa, liu2024raginstructboostingllmsdiverse, joshi2017triviaqa} with long-context code instruction datasets, including SWE-Fixer \citep{xie2025swe}, NextCoder \citep{aggarwal2025nextcoder}, PyResBugs \citep{11052783}, CommitPack \citep{muennighoff2023octopack}, and RepoBench \citep{liu2023repobench}. Since some public datasets contain outdated or low-quality responses, we regenerate targets by labeling them with \texttt{Qwen3-235B-A22B-Instruct-2507}. In addition, we generate three synthetic long-context corpora: (1) repository summarization, using summarization question templates as instructions and concatenated repository files from The Stack v2 \citep{lozhkov2024starcoder} as the document; (2) document summarization, using long documents from FineWiki \citep{penedo2025finewiki} and RedPajama-ArXiv \citep{together2023redpajama}; and (3) repository code generation, where we hold out one file and ask the model to reconstruct it given the remaining repository context. All synthetic completions are generated by \texttt{Qwen3-30B-A3B-Instruct-2507}. 

For instruction following, we include multi-turn conversation data from WildChat and LMSYS-Chat-1M \citep{zhao2024wildchat, zheng2023lmsys}, along with additional alignment-style data obtained by re-labeling with \texttt{Qwen3-30B-A3B-Instruct-2507}. We also add OLMo-3 instruction tuning, Nemotron instruction tuning, and the Tulu-3 mixture \citep{lambert2024tulu3}; since these datasets are already generated by strong teacher models, we keep their original responses unchanged. Finally, we retain a small amount of reconstruction data to stabilize training and mitigate drift when optimizing on compressed inputs.
For agentic behavior, we include tool-calling, interactive, search, and terminal trajectories from Nemotron-Agentic-v1, Nemotron-SFT-Agentic-v2, and OpenThoughts-Agent-SFT-100K \citep{nvidia_nemotron_nano_v3_2025,raoof2026openthoughts}, which are kept uncompressed. We also generate synthetic memory-expansion trajectories with Qwen3 models \citep{qwen3}, in which the source documents are compressed and the model requests individual segments with the \texttt{expand} tool before answering from the returned text.
Exact dataset composition is reported in \Cref{app-tab:SFT-mixture}, and token statistics after packing in \Cref{app-tab:SFT-tokens}.

\begin{table*}[t]
    \centering
    \small
    \setlength{\tabcolsep}{5pt}
    \renewcommand{\arraystretch}{1.12}
    \caption{SFT data mixture. Examples are counts after filtering; token statistics per packing pool are in \Cref{app-tab:SFT-tokens}. The same mixture is used for all compression ratios. Memory-expansion trajectories are generated with Qwen3 models (Apache-2.0); their source documents (e.g., LexGLUE, ACORD, ContractNLI) retain their original licenses.
    We refer to the unique Nvidia license as ``Nvidia''; the license can be found at \url{https://huggingface.co/datasets/nvidia/Nemotron-CC-v2.1/blob/main/LICENSE.md}.
    }
    \label{app-tab:SFT-mixture}
    \resizebox{\linewidth}{!}{
    \begin{tabular}{l l l r}
    \toprule
    \textbf{Data Name} & \textbf{License} & \textbf{Description} & \textbf{Examples} \\
    \midrule
    \textbf{Reasoning} & & & \textbf{5.46M} \\
    \quad Nemotron Reasoning & cc-by-4.0 & Nemotron reasoning (compressed CoT) & 4.23M \\
    \quad Dolci Think & odc-by & OLMo reasoning (compressed CoT) & 1.01M \\
    \quad Nemotron Science & cc-by-4.0 & Nemotron science (MCQ, reading QA) & 0.22M \\
    \midrule
    \textbf{Long-Context Instruction Tuning} & & & \textbf{7.32M} \\
    \quad Repo Summarization & Apache & Synthetic repo-level summarization & 1.35M \\
    \quad Document Summarization & Apache & Synthetic long-document summarization & 1.75M \\
    \quad Repo Code Generation & Apache & Synthetic repo-level code generation & 2.02M \\
    \quad Long-Context QA & Apache & Long-context / RAG QA & 0.84M \\
    \quad Long-Context Code Instruct & Apache & Long-context code instruction tuning & 1.36M \\
    \midrule
    \textbf{General Instruction Following} & & & \textbf{5.26M} \\
    \quad Multi-turn Conversation & odc-by & WildChat + LMSYS Chat 1M & 2.48M \\
    \quad Dolci Instruct & odc-by & OLMo instruction tuning & 1.46M \\
    \quad Tulu3 SFT Mixture & odc-by & Tulu3 instruction tuning & 0.84M \\
    \quad Nemotron Instruct & Nvidia & Nemotron instruction tuning & 0.35M \\
    \quad HelpSteer & cc-by-4.0 & Alignment & 0.13M \\
    \midrule
    \textbf{Reconstruction Data} & & & \textbf{2.08M} \\
    \quad Reconstruction & Nvidia & Small reconstruction mixture & 2.08M \\
    \midrule
    \textbf{Agent} & & & \textbf{1.35M} \\
    \quad Agent Trajectories & cc-by-4.0 / Apache / MIT & Nemotron-Agentic, OpenThoughts-Agent (uncompressed) & 1.19M \\
    \quad Memory Expansion & Apache & Synthetic segment-expansion trajectories (Qwen3) & 0.15M \\
    \midrule
    \textbf{Total} & & & \textbf{21.46M} \\
    \bottomrule
    \end{tabular}
    }
\end{table*}

\begin{table}[t]
    \centering
    \small
    \setlength{\tabcolsep}{5pt}
    \caption{SFT token statistics after packing into sequences of at most 16,384 decoder positions at \(16\times\) compression. Decoder positions include latent positions and memory boundary tags; trainable tokens receive next-token prediction loss. Nemotron Science is packed in the other pool.}
    \label{app-tab:SFT-tokens}
    \begin{tabular}{lrrrrr}
    \toprule
    Pool & Examples & Sequences & Decoder positions & Trainable tokens & Share \\
    \midrule
    Agent & 1.35M & 0.27M & 4.36B & 1.10B & 11.7\% \\
    Reasoning & 5.24M & 0.71M & 11.66B & 9.44B & 31.4\% \\
    Other & 14.88M & 1.29M & 21.16B & 17.24B & 56.9\% \\
    \midrule
    Total & 21.46M & 2.28M & 37.19B & 27.78B & 100\% \\
    \bottomrule
    \end{tabular}
\end{table}

\subsection{Compression Data Format}
\label{compression-data-details}
\paragraph{Continual pre-training.}
For the continual pre-training data, given a sequence, we split it into multiple segments, wrap every other segment with the special tokens \texttt{<|memory\_start|>} and \texttt{<|memory\_end|>}, and treat the wrapped spans as compressed content while leaving the remaining segments uncompressed. 

For raw pre-training documents, we construct a fixed alternating pattern of wrapped and unwrapped segments, always beginning with a wrapped segment and ending with an unwrapped segment. Documents longer than 16,384 tokens retain 8,192 trainable (unwrapped) tokens, while shorter documents retain half of their tokens as trainable. The number of segment pairs depends on document length: documents shorter than 2,048 tokens receive 1 pair; documents between 2,048 and 4,096 tokens receive 1-2 pairs; documents between 4,096 and 8,192 tokens receive 2-4 pairs; documents between 8,192 and 16,384 tokens receive 4-6 pairs; and documents longer than 16,384 tokens receive 4-8 pairs. Within each sequence, the first wrapped segment contains 30\%-40\% of the total wrapped tokens, the last unwrapped segment contains 40\%-50\% of the total trainable tokens, and the remaining segments are allocated with a $\pm 20\%$ jitter. No segment is allowed to be shorter than 128 tokens.

\paragraph{Instruction tuning.}
For conversational data, we apply wrapping independently to each message. With 50\% probability, the entire message content is enclosed within a single pair of tags. Otherwise, with the remaining 50\% probability, we insert 1-3 wrapped regions inside the message and leave the remaining tokens outside the tags. The number of wrapped regions depends on message length: messages shorter than 32 tokens are always fully wrapped; messages between 32 and 200 tokens receive 1 region; messages between 200 and 600 tokens receive 1-2 regions; and messages of 600 tokens or more receive 1-3 regions. Wrapped regions cover 80\%-100\% of the message tokens in aggregate, and we enforce a minimum 5\% unwrapped gap whenever feasible. If the realized wrapped coverage falls below 70\%, we fall back to fully wrapping the message. Since this decision is made independently for each message, multi-turn conversations naturally contain a mixture of fully wrapped turns and turns with small unwrapped spans, with a bias toward high overall coverage.

\section{Extended Architecture Search}\label{app-sec:arch}

\subsection{Prior Soft-Token Methods in Our Design Space}\label{app-subsec:soft-token}
\Cref{app-tab:soft-token} describes representative soft-token compressors by the design axes we sweep. Each column corresponds to one of our controlled comparisons: the readout to the pooling operator and encoding granularity (\Cref{fig:arch_sweep}, \Cref{app-subsec:encoder-defs}), the encoder to the attention mask (\Cref{fig:arch_sweep}) and initialization (\Cref{app-subsec:encoder_lm_or_emb}), the adapter to the adapter comparison (\Cref{app-subsec:adapter}), and decoder training to the frozen, LoRA, and full-parameter settings (\Cref{sec:training}). These comparisons isolate individual design choices under matched data, scale, and decoder; they do not reproduce the complete prior systems, which also differ in training data, scale, and target tasks.

\begin{table}[!ht]
\centering
\caption{\textbf{Prior soft-token compressors described by our design axes.}}
\label{app-tab:soft-token}
\small
\setlength{\tabcolsep}{4pt}
\resizebox{\linewidth}{!}{%
\begin{tabular}{lllll}
\toprule
Method & Readout & Encoder & Adapter & Decoder training \\
\midrule
AutoCompressor~\citep{auto-compressor} & learned summary tokens & the decoder LM itself & -- & full (OPT) / LoRA (Llama-2) \\
ICAE~\citep{in-context-autoencoder} & learned memory tokens & LLM with LoRA & -- & frozen \\
REFRAG~\citep{refrag} & one embedding per chunk ($W=N$) & RoBERTa encoder & projection layer & full, after alignment \\
E2LLM~\citep{e2llm} & attention pooling per chunk & pretrained embedding model & adapter & LoRA \\
GMSA~\citep{GMSA} & mean over token groups & LLM with LoRA & attention layers & full, in fine-tuning \\
Simple Context Compression~\citep{mean-pooling} & mean pooling & LLM, bidirectional, LoRA & -- & LoRA \\
xRAG~\citep{xrag} & one retrieval embedding & frozen retriever & 2-layer MLP & frozen \\
CEPE~\citep{cepe} & cross-attention to encoder states & small context encoder & cross-attention & frozen \\
\midrule
\name (ours) & mean pooling, $W=1024$ & embedding model, causal & MLP & full, staged \\
\bottomrule
\end{tabular}}
\end{table}

\begin{figure*}[t!]
  \centering

  \includegraphics[width=1.0\textwidth]{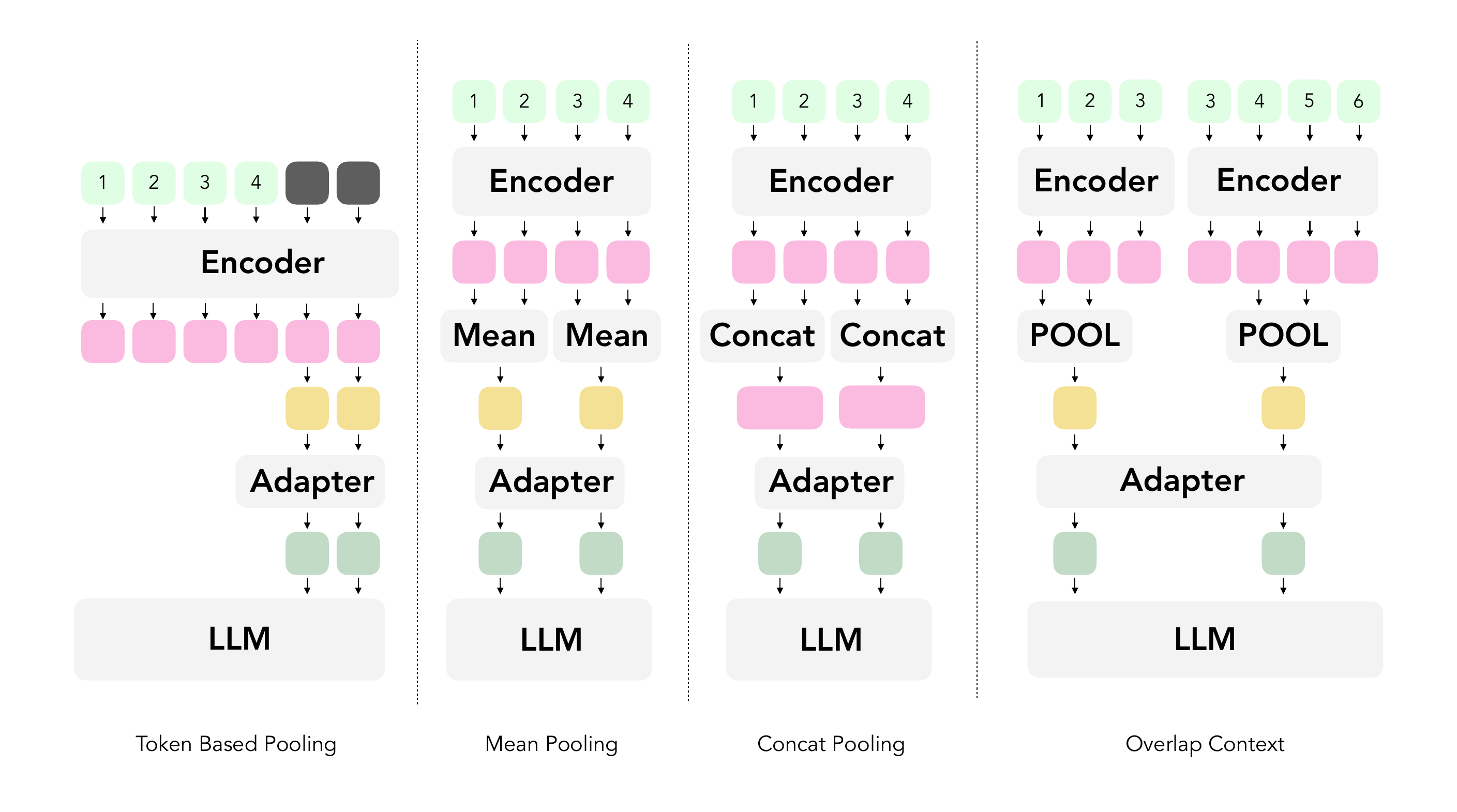}
  \caption{\textbf{Full architecture design choices for the compressor.} From left to right, we visualize token-based pooling, mean pooling, concat pooling, and overlap context.}
  \label{app-fig:arch_sweep}

  \vspace{0.5em}

  \includegraphics[width=0.8\textwidth]{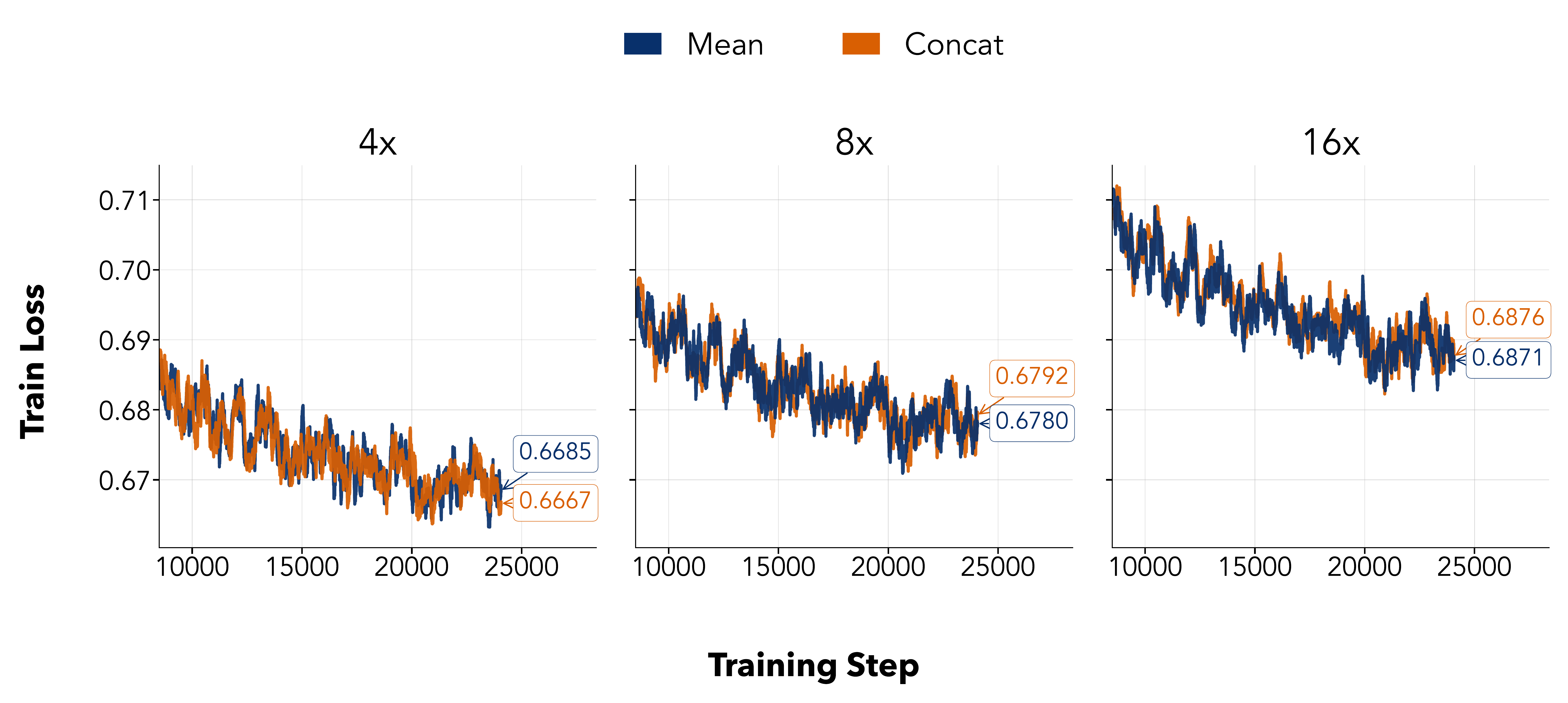}
  \caption{\textbf{Comparison between mean and concat at scale for all ratios.}
  From \Cref{fig:arch_sweep}, we identify mean and concat as the most promising pooling operators, with mean slightly outperforming concat in training loss. We train at larger scale and find that mean remains marginally better than concat at higher compression ratios, but loses to concat at lower compression ratios.}
  \label{app-fig:pool-abl-at-scale}
\end{figure*}

\subsection{Formal Definitions of Pooling Operators}\label{app-subsec:encoder-defs}
We now give the formal definitions of pooling operators, and illustrate them visually in \Cref{app-fig:arch_sweep}.

\paragraph{Token-based pooling.}
The most common method is appending one or more learned special pooling tokens per latent output to each encoder window.
Let \(p_{i,1},\dots,p_{i,M_i}\) denote these pooling tokens. 
We encode
\begin{equation}
(\tilde h^{(i)}_1,\dots,\tilde h^{(i)}_{|w_i|+M_i})
=
\mathrm{Enc}_\phi([w_i;p_{i,1};\dots;p_{i,M_i}]),
\end{equation}
and use the hidden states at the pooling-token positions:
\begin{equation}
z^{(i)}_k
=
\tilde h^{(i)}_{|w_i|+k},
\qquad
k=1,\dots,M_i.
\end{equation}
This generalizes the common EOS-pooling strategy: when \(W=N\), a single pooling token is appended for each compression block; when \(W>N\), several pooling tokens summarize different regions of the same encoder window.

\paragraph{Mean pooling.}
For each window \(w_i\), we partition its hidden states into consecutive groups of size \(N\). 
For latent index \(k=1,\dots,M_i\), define
\begin{equation}
G_{i,k}
=
\{(k-1)N+1,\dots,\min(kN,|w_i|)\}.
\end{equation}
The corresponding latent vector is the average hidden state over this group:
\begin{equation}
z^{(i)}_k
=
\frac{1}{|G_{i,k}|}
\sum_{j\in G_{i,k}} h^{(i)}_j.
\end{equation}
Mean pooling directly aggregates the token-level representations assigned to each compression block. 
Across our architecture search, mean pooling consistently outperforms EOS-style token pooling, matching observations from prior text-compression and vision-encoder work.

\paragraph{Concat pooling.}
Rather than averaging or selecting a designated token, concat pooling preserves the full per-token representation by concatenating consecutive encoder hidden states within each compression block.
For window \(w_i\) and latent index \(k=1,\dots,M_i\), let \(G_{i,k}=\{(k-1)N+1,\dots,\min(kN,|w_i|)\}\) as in mean pooling. The concatenated latent is
\begin{equation}
z^{(i)}_k
=
\bigl[\,h^{(i)}_{j}\,\bigr]_{j\in G_{i,k}}\in\mathbb{R}^{N\,d_{\mathrm{enc}}},
\end{equation}
i.e., a vector of dimension \(N\,d_{\mathrm{enc}}\) formed by stacking the \(N\) encoder hidden states.
The adapter is widened accordingly so that its first MLP layer takes inputs of dimension \(N\,d_{\mathrm{enc}}\) and projects back to the decoder hidden dimension \(d_{\mathrm{dec}}\), analogous to the patch-merger / spatial-concat scheme used in \citet{bai2025qwen3}.

\begin{figure*}[t]
  \centering
  \includegraphics[width=\textwidth]{figs/adapter_overlap.pdf}
  \caption{\textbf{MLP adapters, causal masking, and no boundary overlap give the lowest pre-training loss across the architecture sweep.} We sweep combinations of adapter, boundary-overlap, and encoder mask, and find that (1) causal masking is strictly better than bidirectional masking, (2) boundary overlap does not consistently improve over non-overlapping windows while substantially increasing compute, and (3) the MLP adapter outperforms attention-based adapters in the from-scratch runs. At scale (plot d), causal masking and no overlap remain best, while the two adapters are close.}
  \label{app-fig:adapter-ov-loss}
\end{figure*}

\subsection{Adapter Design}
\label{app-subsec:adapter}
The adapter \(a(\cdot)\) maps the compressed latent sequence \(z_{1:M}\in\mathbb{R}^{M\times d_{\mathrm{enc}}}\) into a sequence \(s_{1:M}\in\mathbb{R}^{M\times d_{\mathrm{dec}}}\) in the decoder embedding space.
Prior encoder-decoder compression work often assumes \(d_{\mathrm{enc}}=d_{\mathrm{dec}}\), but a wide range of encoder-decoder models pair smaller encoders with larger decoders, making the adapter an important design choice.
We formally define the two adapter families in our sweep below.

\paragraph{MLP adapter.}
The MLP adapter is a standard two-layer feed-forward network with a GELU non-linearity~\citep{hendrycks2016gaussian} and an RMSNorm~\citep{zhang2019root} pre-normalization, following the projection design used in LLaVA-style VLMs~\citep{llava}:
\begin{align}
\hat z_m \;&=\; \mathrm{RMSNorm}(z_m), \\
s_m \;&=\; W_2\,\mathrm{GELU}\!\bigl(W_1 \hat z_m + b_1\bigr) + b_2,
\qquad m = 1,\dots,M,
\end{align}
where \(W_1\in\mathbb{R}^{d_h\times d_{\mathrm{enc}}}\) and \(W_2\in\mathbb{R}^{d_{\mathrm{dec}}\times d_h}\) with hidden dimension \(d_h\). Each latent token is projected independently and the adapter does not mix information across the latent sequence. When the encoder uses concat pooling (\Cref{app-subsec:encoder-defs}), each \(z_m\) has dimension \(N\,d_{\mathrm{enc}}\) instead of \(d_{\mathrm{enc}}\), so the input dimension of \(W_1\) is widened to \(N\,d_{\mathrm{enc}}\) and the first MLP layer is correspondingly \(N\times\) larger.

\paragraph{Attention-based adapter.}
The attention-based adapter adds a single multi-head self-attention layer over the latent sequence before the MLP projection, with an RMSNorm pre-normalization on the latents:
\begin{align}
\hat z_m \;&=\; \mathrm{RMSNorm}(z_m), \\
\tilde z_{1:M} \;&=\; \mathrm{MHSA}\bigl(\hat z_{1:M}\bigr), \\
s_m \;&=\; W_2\,\mathrm{GELU}\!\bigl(W_1 \tilde z_m + b_1\bigr) + b_2,
\qquad m = 1,\dots,M.
\end{align}
The self-attention layer operates in the encoder hidden dimension \(d_{\mathrm{enc}}\) and lets information mix across latent tokens before projection, at the cost of additional parameters and a compute term quadratic in \(M\).

\paragraph{Comparison.}
Across the from-scratch runs of the adapter sweep in \Cref{app-fig:adapter-ov-loss}, the MLP adapter achieves lower pre-training loss than the attention-based variant while adding less computation, in contrast to prior findings \citep{GMSA} that attention-based adapters perform better. We therefore use the MLP adapter as the default in all subsequent experiments.

\subsection{Boundary Context}\label{app-subsec:arch-overlap}
A potential limitation of fixed encoder windows is that information crossing a window boundary is split across two encoder forward passes.
Increasing the window size \(W\) reduces the frequency of such boundaries, but also increases compression-time cost.
As an alternative, we implement boundary context overlapping, where each encoder window is extended with \(O\) neighboring tokens while the number of latent tokens is kept unchanged.

For a target window
\[
w_i = x_{(i-1)W+1:\min(iW,T)},
\]
we construct an overlapped encoder input \(\tilde w_i\) by adding neighboring context tokens.
For bidirectional attention, we include both left and right overlap when available:
\begin{equation}
\tilde w_i
=
x_{\max(1,(i-1)W-O+1):\min(iW+O,T)}.
\end{equation}
We provide a visualization in the rightmost figure in \Cref{app-fig:arch_sweep}. For causal attention, future context is unavailable, so we include only left overlap:
\begin{equation}
\tilde w_i
=
x_{\max(1,(i-1)W-O+1):\min(iW,T)}.
\end{equation}

After encoding \(\tilde w_i\), we apply pooling only to hidden states corresponding to the original non-overlapped region \(w_i\), and discard pooled outputs that correspond to overlap tokens.
Thus, overlap changes the encoder context available to each window but does not change the number of latent tokens passed to the decoder.

We sweep \(O\in\{0,32,256\}\) for \(W\in\{256,1024\}\) in \Cref{app-fig:adapter-ov-loss}.
Overlap does not noticeably improve pre-training loss in most settings, while increasing training cost because boundary tokens are encoded multiple times.
For example, with \(W=1024\) and \(O=256\), each interior bidirectional window processes \(1536\) tokens instead of \(1024\).
We therefore do not use overlap in the default architecture.

\section{Full-Scale Training Recipe Sweeps}\label{app-sec:training-recipe-sweeps}
\begin{table*}[t!]
    \small
    \centering
    \setlength{\tabcolsep}{8pt}
    \caption{\textbf{Training recipe for compression model for four different stages.} We report, for each stage, which module(s) are optimized (adapter / encoder / decoder), peak learning rates, learning-rate schedule, optimizer, global batch size (in tokens), sequence length, training-token budget, and the data source used.}
    \label{tab:training_recipe}
    \resizebox{\linewidth}{!}{
    \begin{tabular}{l|cccc}
    \toprule
     & \textbf{Stage 0} & \textbf{Stage 1} & \textbf{Stage 2} & \textbf{Stage 3} \\
     & \textbf{Adapter Training} & \textbf{Encoder Training} & \textbf{LLM Training} & \textbf{SFT} \\
    \hline
    \\[-4pt]
    \textbf{Training Config} \\
    Adapter Peak LR  & $1.0 \times 10^{-3}$ & $6.0 \times 10^{-5}$ & $6.0 \times 10^{-5}$ & $3.0 \times 10^{-5}$ \\
    Encoder Peak LR & NA & $6.0 \times 10^{-5}$ & $6.0 \times 10^{-5}$ & $3.0 \times 10^{-5}$ \\
    LLM Peak LR & NA & NA & $1.0 \times 10^{-5}$ & $3.0 \times 10^{-5}$ \\
    LR scheduler & \multicolumn{4}{c}{5\% warmup steps with cosine decay to $1.0 \times 10^{-6}$} \\
    Optimizer & \multicolumn{4}{c}{AdamW ($\beta_1=0.9$, $\beta_2=0.95$)} \\
    \midrule
    \textbf{Data} \\
    LLM Batch size (tokens) & \multicolumn{4}{c}{4 Million} \\
    LLM Sequence Length & \multicolumn{4}{c}{16384} \\
    Encoder Tokens & $17.54$ B & $35.07$ B & $82.43$ B & $60.35$ B \\
    LLM Tokens (16x) & $21.29$ B & $42.58$ B & $100.08$ B & $37.19$ B \\
    Total Tokens (16x) & $38.83$ B & $77.65$ B & $182.51$ B & $97.54$ B \\
    Data Source & pre-training Mixture & pre-training Mixture & pre-training Mixture & SFT \\
    \bottomrule
    \end{tabular}
    }
\end{table*}

\subsection{Pre-trained Encoder Representations}\label{app-subsec:encoder_lm_or_emb}
For the encoder initialization, to determine whether an embedding model's representation \citep{refrag,e2llm,cepe} is better than that of a standard language model \citep{GMSA, auto-compressor, mean-pooling}, we conduct an experiment with the EOS pooling token.
We can see from \Cref{app-fig:large-scale-sweep} (d) that using the embedding model as the encoder outperforms using an LLM as the encoder.
Note the embedding model was initialized from the LLM \citep{qwen3-embedding}.
This shows that embedding model training provides a better representation for the encoder, though the gap gradually narrows with longer training. We also show downstream evaluation performance in \Cref{tab:abl-encoder}.

\begin{figure*}[t]
  \centering
  \includegraphics[width=1.0\textwidth]{figs/cpt_ablation.pdf}
    \caption{\textbf{Architecture choices selected by the small-scale sweep continue to hold at scale.} Continual pre-training loss for the full-pipeline runs at compression ratio \(16\times\). Across (a) encoder window size, (b) encoder attention mask, (c) stage-2 learning rate, and (d) encoder initialization (LM vs. embedding), the configuration chosen from the from-scratch sweep (\(W=1024\), causal masking, embedding-initialized encoder) yields the lowest pre-training loss.}
    \label{app-fig:large-scale-sweep}
\end{figure*}

\subsection{LR Sweep}
\label{app-sec:continunal pre-training lr sweep}
Language models are prone to catastrophic forgetting during continual pre-training and SFT \citep{luo2025empirical,li2024revisiting}. To actively avoid forgetting, we sweep the peak learning rate for stages 2 and 3 of training.
To test for catastrophic forgetting, we then benchmark the language model without compression after SFT and check that its performance is comparable to the benchmark performance of the language model before training.
For stage 2, we sweep from $1\times10^{-6}$ to $1\times10^{-5}$ and find that a larger learning rate improves downstream benchmarks with minimal drift from the original model (\Cref{app-subsec:retention}).
For stage 3, we sweep from $2\times10^{-5}$ to $5\times10^{-5}$. We do not observe significant differences in those SFT LRs across different models; we report the detailed numbers on downstream benchmarks in \Cref{app-subsec:lr-sweep}.

\subsection{Decoder Retention Without Compression}
\label{app-subsec:retention}
To measure how much training changes the decoder itself, we evaluate our \(4\times\), \(8\times\), and \(16\times\) models with compression disabled, so that the decoder reads the full context as ordinary tokens, and compare them with the original \texttt{Qwen3-4B-Instruct-2507}. All four models use identical prompts, output budgets, and scoring; the base model is re-evaluated under this protocol. As shown in \Cref{app-tab:retention}, the trained decoders stay within 1.7 points of the original model on RULER at 4k, within 2.8 points at 8k, and within 4.5 points at 16k, and slightly exceed it on LongBench.

\begin{table}[!ht]
\centering
\caption{\textbf{Accuracy with compression disabled.} The decoders of our \(4\times\), \(8\times\), and \(16\times\) models read the uncompressed context and are compared with the original model under the same protocol (RULER: 13 tasks, 6,500 questions per context length; LongBench EN16: macro average over 16 English and code tasks). Differences from the original model are in parentheses.}
\label{app-tab:retention}
\small
\resizebox{\linewidth}{!}{%
\begin{tabular}{lcccc}
\toprule
 & \texttt{Qwen3-4B-Instruct-2507} & LCLM \(4\times\) decoder & LCLM \(8\times\) decoder & LCLM \(16\times\) decoder \\
\midrule
RULER 4k & 94.49 & 94.27 (\(-0.22\)) & 92.86 (\(-1.63\)) & 93.41 (\(-1.08\)) \\
RULER 8k & 93.44 & 91.55 (\(-1.89\)) & 90.73 (\(-2.71\)) & 90.98 (\(-2.46\)) \\
RULER 16k & 93.53 & 91.54 (\(-1.99\)) & 89.46 (\(-4.07\)) & 89.10 (\(-4.43\)) \\
LongBench EN16 & 45.30 & 47.13 (\(+1.83\)) & 45.98 (\(+0.68\)) & 46.91 (\(+1.61\)) \\
\bottomrule
\end{tabular}}
\end{table}

\begin{figure*}[t!]
\centering
\includegraphics[width=0.5\linewidth]{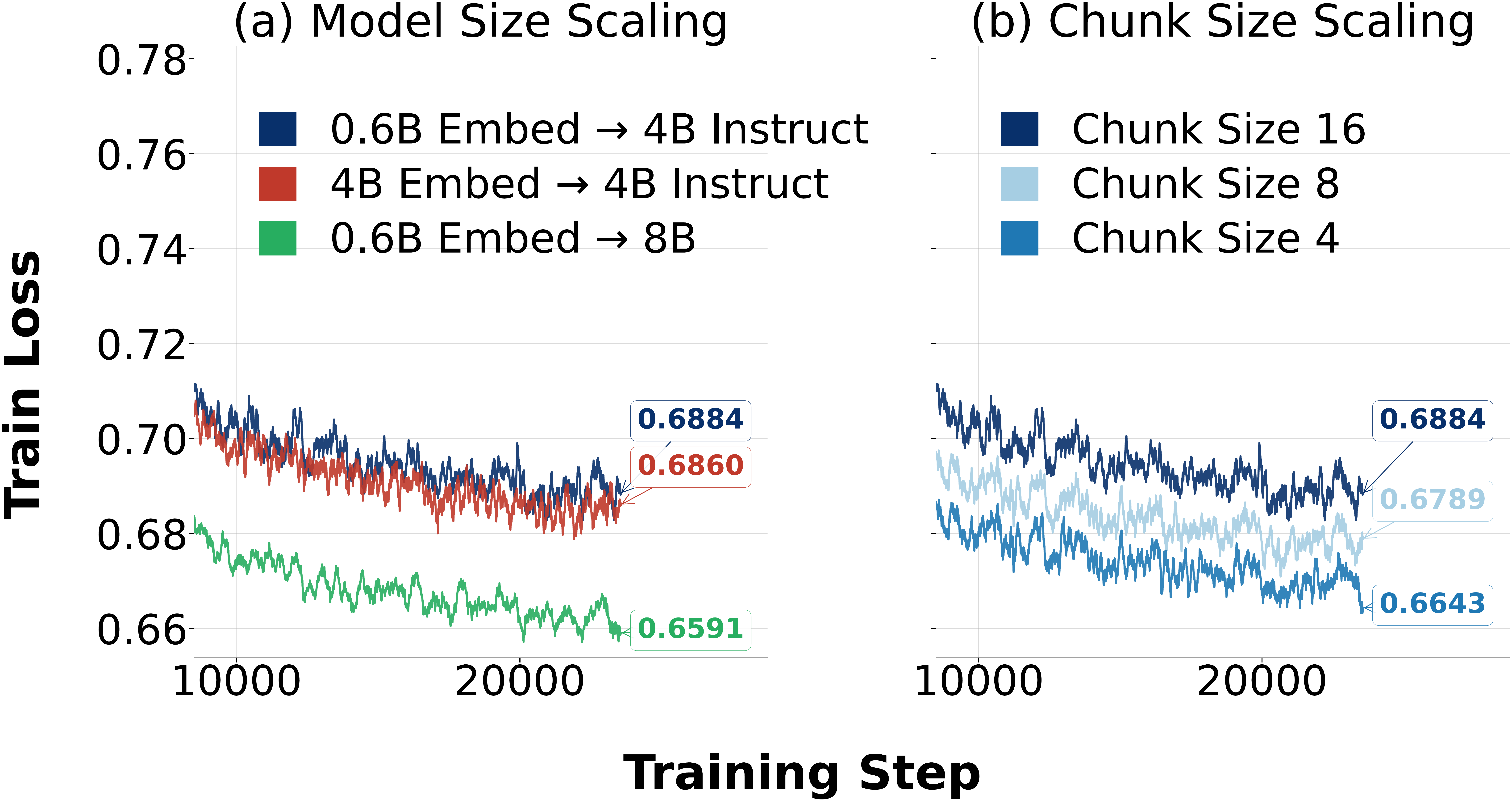}
  \caption{\textbf{Scaling the decoder helps more than scaling the encoder, and lower compression ratios yield lower pre-training loss.} \textbf{Left:} pre-training loss at the end of stage 2 with compression ratio \(16\times\) across encoder/decoder size combinations; increasing the decoder size produces a much larger drop in loss than increasing the encoder size. \textbf{Right:} pre-training loss at the end of stage 2 for our 0.6B-4B models across compression ratios; lower compression ratios yield lower pre-training loss. Note that the 4B-encoder and 8B-decoder runs use 256 ranks rather than 128 ranks. Because our stateful Parquet dataloader assigns work by rank, these runs complete slightly fewer optimization steps. For a clean comparison, we therefore truncate all loss curves at the same optimization step in the plot above.}
  \label{fig:stage_2_loss}
\end{figure*}

\subsection{Scaling Up Model Size}
\label{app-sec:scaling model size}
To study scaling behavior, we ask two questions: (i) does a larger encoder improve compression quality, and (ii) does a larger decoder, which has a wider hidden dimension that could compress more information per token, improve performance under compression? 
Accordingly, we add two model configurations: (1) using the larger \texttt{Qwen3-Embedding-4B} as the encoder with the same size \texttt{Qwen3-4B-Instruct-2507} decoder; and (2) using the same size \texttt{Qwen3-Embedding-0.6B} encoder and the larger \texttt{Qwen3-8B} as the decoder. In \Cref{fig:stage_2_loss} (a), we see that increasing the size of the decoder is much more beneficial in terms of pre-training loss than increasing the size of the encoder.

However, the scaling results in \Cref{app-tab:scaling} are mixed. 
The \(0.6\)B encoder performs best across the RULER tasks, while the \(4\)B encoder performs best on the remaining evaluations. 
In contrast, the \(8\)B decoder achieves substantially lower pre-training loss but does not yield the downstream gains we expected. 
We suspect this is partly due to a mismatch between the training mixture and the decoder initialization: our data curation and recipe were tuned around the \(4\)B instruct decoder, whereas the \(8\)B decoder is a hybrid-thinking model and may require a different data distribution. 
Moreover, the \(4\)B instruct model already outperforms the \(8\)B model in the full-cache setting on several evaluations, which may limit the apparent benefit of scaling the decoder in our current setup.

\begin{table}[!ht]
\centering
\caption{Encoder/decoder scaling at 16$\times$ compression. These models were trained with an earlier version of our SFT data, so the 0.6B/4B row differs from \Cref{app-tab:summary}; LongBench averages all 21 tasks.}
\label{app-tab:scaling}
\small
\setlength{\tabcolsep}{6pt}
\begin{tabular}{lrrrrrr}
\toprule
Enc / Dec & RULER 4K & RULER 8K & RULER 16K & LongBench & LongHealth5 & GSM8K \\
\midrule
0.6B / 4B & \textbf{75.06} & \textbf{70.23} & \textbf{65.91} & 37.33 & \underline{67.50} & \underline{81.05} \\
0.6B / 8B & 71.20 & \underline{67.40} & \underline{62.22} & \underline{37.80} & 64.80 & 77.30 \\
4B / 4B   & \underline{71.30} & 66.90 & 62.00 & \textbf{39.00} & \textbf{69.80} & \textbf{83.20} \\
\bottomrule
\end{tabular}
\end{table}

\subsection{Packed Sequences}
Since our corpus contains many long-context examples with highly variable lengths, conventional batching would waste substantial computation on padding. To improve training throughput, we pack multiple examples into a single sequence. Because we continually pre-train an instruction-tuned model, we prevent cross-example information leakage by resetting the attention mask at example boundaries, yielding a block-diagonal attention pattern. We implement this efficiently using variable-length attention kernels~\citep{dao2023flashattention, dong2024flex}. 
\clearpage

\section{Compute}
\label{app-subsec:compute}
All pre-training-from-scratch experiments are run on 16 nodes of MI300A GPUs, each containing 4 GPUs. Full-scale continual pre-training experiments for 0.6B-4B models are trained on \(32\) nodes of MI300A GPUs, each containing \(4\) GPUs.

All evaluations are conducted on H200 GPUs. 

All scaling experiments (\texttt{Qwen3-Embedding-4B} encoder, \texttt{Qwen3-8B} decoder) are trained on H200 clusters with 32 nodes to alleviate memory issues.
\clearpage

\FloatBarrier
\FloatBarrier
\clearpage

\begin{figure*}[!t]

\section{Benchmarks and Results}
\label{app-sec:benchmarks}

\vspace{0.5em}

{
\centering
\small
\setlength{\tabcolsep}{6pt}
\renewcommand{\arraystretch}{1.12}

\captionof{table}{\textbf{Evaluation benchmarks and what content is compressed.}
We evaluate on long-context suites and on GSM8K, a short and information-dense benchmark. The ``What is Compressed'' column specifies which part of the input is placed inside the compressible memory block, while remaining components, such as instructions, are kept as hard tokens when applicable.}
\label{app-tab:eval-benchmarks}

\resizebox{\linewidth}{!}{
\begin{tabular}{l p{0.38\textwidth} p{0.2\textwidth}}
\toprule
\textbf{Benchmark} & \textbf{Description} & \textbf{What is Compressed} \\
\midrule
\textbf{Long-context} & & \\
\quad RULER & Synthetic long-context stress tests from 4K--16K. & Task-specific context \\
\quad LongBench & Aggregated long-context suite covering single- and multi-document QA, summarization, few-shot learning, synthetic tasks, and code completion. & Task-specific context \\
\quad LongHealth & Clinical-document multiple-choice QA with patient-case context. Five documents are concatenated. No CoT prompting. & Task-specific context \\
\midrule
\textbf{Fine-Grained Compression} & & \\
\quad GSM8K & Short and information-dense grade-school math word problems. & Whole prompt \\
\bottomrule
\end{tabular}
}

\par
}

\vspace{1.25em}

\raggedright
\normalsize
\subsection{GSM8K Performance vs Compression Ratio}

The long-context benchmarks show that \names can compress extended contexts efficiently. We next evaluate on GSM8K~\citep{cobbe2021training} to test whether \names can also handle short, information-dense inputs, where nearly every token may be relevant to the answer. In \Cref{app-fig:gsm}, \names achieve the highest accuracy across all compression ratios on GSM8K, with particularly strong gains over baselines at higher compression ratios. These are aggressive compression settings: compression ratios of \(16\times\) and \(8\times\) remove 93.75\% and 87.5\% of the input tokens, respectively. Strong performance in this setting demonstrates that \names are not specialized only for long-context QA; they can also compress short, dense contexts, making them practical for general-purpose deployment.

\vspace{1.25em}

{
\centering
\includegraphics[width=0.5\linewidth]{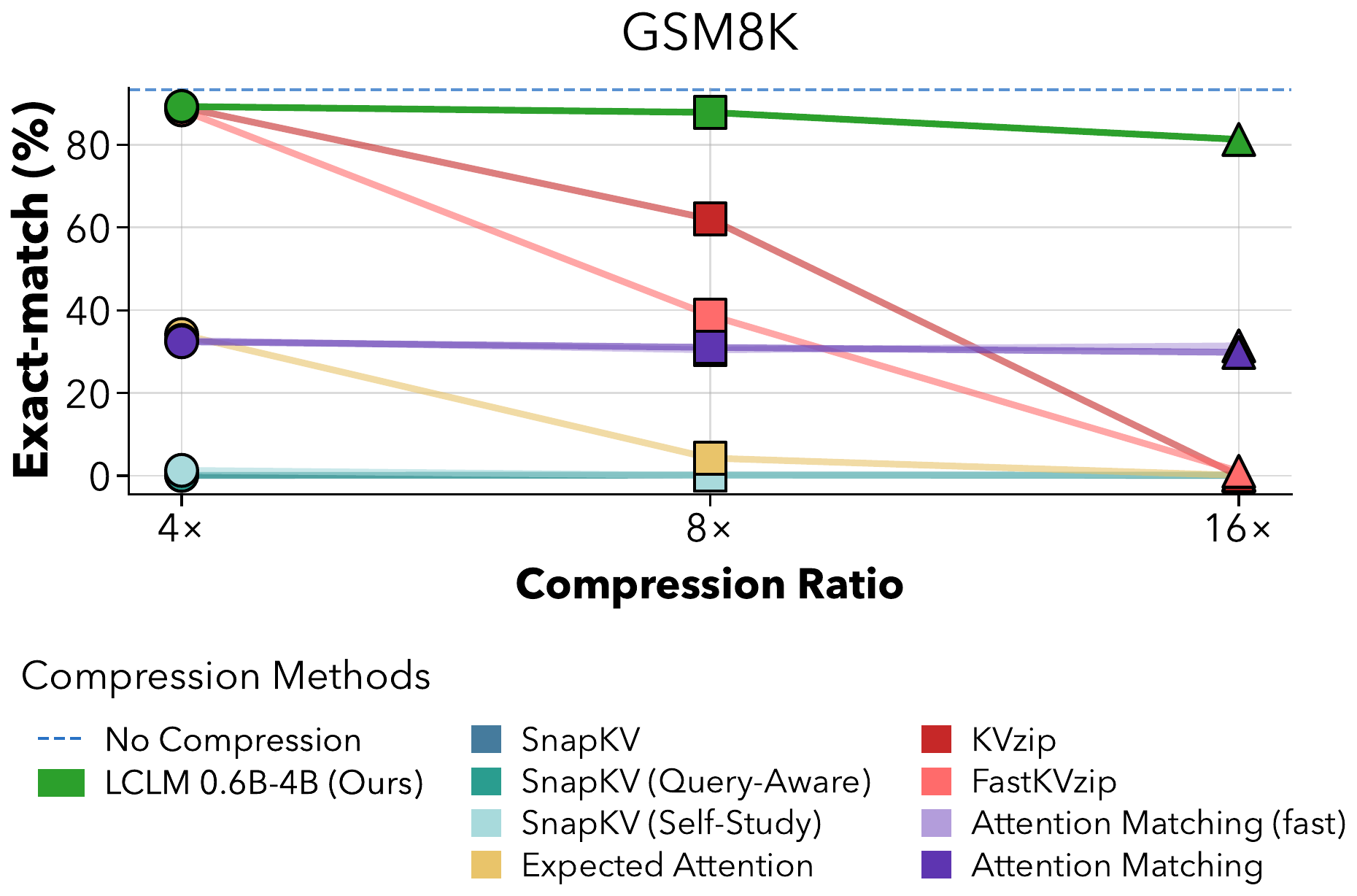}

\caption{\textbf{\fullnames can compress small dense contexts with high accuracy.}
We plot GSM8K accuracy over different compression ratios and find that \names maintain much higher accuracy at larger compression ratios. Full data are provided in \Cref{app-tab:summary}. We do not report compression time for GSM8K because the context is small enough that compression-time measurements are too noisy to support reliable conclusions.}
\label{app-fig:gsm}

\par
}

\end{figure*}

\subsection{Baseline Configurations}\label{app-subsec:baselines}
Query-aware methods, such as SnapKV, are designed to compress a context given a known query; however, the resulting compressed cache may not generalize well to unseen queries or future turns. Query-agnostic methods, such as KVzip and FastKVzip, compress a cache without access to the downstream query, often using reconstruction-style objectives such as repeating the original context. Self-study, as introduced by \citet{cartridges}, performs compression using synthetic queries to expand coverage over potential inference-time queries. These synthetic queries are typically produced by prompting the target decoder to generate questions about the context. In Attention Matching~\citep{zweiger2026fast}, the authors experiment with a mixed query-agnostic and self-study configuration and find that self-study has limited impact on the repeat-prefill prompt. Since self-study introduces additional compression-time overhead, we do not apply it to RULER.

It is also important to distinguish algorithmic and theoretical cache reduction from systems-realizable cache reduction. Some KV cache methods produce an actual smaller cache that can, in principle, reduce decoding memory and latency. Others, such as \citet{kim2025kvzip, kim2026fast}, are evaluated by masking or subsampling entries from a full cache, or by using non-uniform eviction patterns across heads and layers. These approaches are useful for measuring the quality of a compressed cache, but they do not necessarily translate into wall-clock speedups or memory savings in standard inference engines without specialized kernels that have not been released or even implemented.

\subsection{Timing Implementation}\label{app-subsec:timing}
\paragraph{Comparison with KV cache compression.} In \Cref{fig:cover,fig:pareto_optimal}, the decoder of every method, including ours, runs in Hugging Face Transformers with fused attention kernels: FlashAttention-2~\citep{dao2023flashattention} for the baselines and PyTorch scaled dot-product attention for ours. Our encoder additionally uses FlashAttention, \texttt{torch.compile}, and CUDA graphs, with an encoder batch size of 32 windows.

\paragraph{Comparison with uncompressed models.} In \Cref{fig:lm_pareto}, the decoder of every model, including ours, is served with vLLM~\citep{kwon2023efficient} in BF16 on a single H200, with batch size 1, CUDA graphs enabled, and prefix caching disabled. Our encoder uses the same optimized implementation as above. For \names, we time the encoder and the vLLM decoder prefill over a sequence of the same length as the compressed latent sequence, and report their sum. Each point is the fastest of 20 timed runs after 30 warmup runs.

\clearpage
\FloatBarrier
\subsection{Results}

\begin{table}[!ht]
\centering
\caption{Summary across compression ratios: RULER, LongBench, LongHealth5, GSM8K. Rows are grouped by compression ratio (16x, 8x, 4x). Best per column within each ratio block in \textbf{bold}, second-best \underline{underlined}.}
\label{app-tab:summary}
\scriptsize
\setlength{\tabcolsep}{3pt}
\resizebox{\linewidth}{!}{%
\begin{tabular}{lrrrrrrr}
\toprule
Model & RULER 4k & RULER 8k & RULER 16k & LB en16 & LB cn5 & LongHealth5 & GSM8K \\
\midrule
Qwen 4B Instruct (full KV) & 94.49 & 93.44 & 93.53 & 45.30 & 46.54 & 76.75 & 93.25 \\
\midrule
\multicolumn{8}{c}{\textit{16x compression}} \\
\midrule
LCLM 0.6b-4b (Mean) & \textbf{77.01} & \textbf{72.10} & \textbf{67.04} & 38.26 & 31.38 & 66.25 & \textbf{81.27} \\
ExpAttn & 40.97 & 46.93 & 50.67 & 29.44 & 20.51 & 30.25 & 0.08 \\
SnapKV & 14.31 & 14.11 & 18.27 & 26.96 & 18.30 & 10.50 & 0.08 \\
SnapKV-QA & 20.54 & 39.09 & \underline{61.96} & 38.52 & \textbf{42.26} & \textbf{76.25} & 0.08 \\
KVzip & 62.73 & 61.73 & 57.97 & 25.86 & 17.12 & 67.00 & 0.00 \\
\bottomrule
\end{tabular}
}
\end{table}

\begin{table}[!ht]
\centering
\caption{RULER per-task at 4k context. Rows grouped by compression ratio (16x, 8x, 4x).}
\label{app-tab:ruler-4k}
\scriptsize
\setlength{\tabcolsep}{3pt}
\resizebox{\linewidth}{!}{%
\begin{tabular}{lrrrrrrrrrrrrr|r}
\toprule
Model & ns1 & ns2 & ns3 & nm1 & nm2 & nm3 & nmv & nmq & vt & cwe & fwe & qa1 & qa2 & AVG \\
\midrule
Qwen 4B Instruct & 100.00 & 100.00 & 99.80 & 99.80 & 96.80 & 100.00 & 99.60 & 100.00 & 99.60 & 97.72 & 87.60 & 84.00 & 63.40 & 94.49 \\
\midrule
\multicolumn{15}{c}{\textit{16x compression}} \\
\midrule
LCLM 0.6b-4b (Mean) & \underline{96.60} & 90.20 & \underline{58.40} & \textbf{86.00} & 84.20 & \textbf{63.20} & \textbf{74.85} & \textbf{75.55} & 82.92 & \underline{76.04} & \textbf{92.80} & \underline{67.60} & \textbf{52.80} & \textbf{77.01} \\
ExpAttn & \textbf{100.00} & 49.20 & 0.60 & 57.60 & 2.20 & 0.00 & 56.75 & 25.70 & 97.80 & 18.36 & 66.20 & 33.20 & 25.00 & 40.97 \\
SnapKV & 4.00 & 1.60 & 2.40 & 9.80 & 1.80 & 0.00 & 8.55 & 8.55 & 5.48 & 25.30 & 73.13 & 24.80 & 20.60 & 14.31 \\
SnapKV-QA & 16.60 & 7.20 & 2.40 & 10.20 & 4.80 & 0.00 & 9.10 & 10.50 & 15.80 & 25.14 & 57.67 & 60.60 & 47.00 & 20.54 \\
KVzip & \textbf{100.00} & \underline{91.60} & \textbf{96.00} & 62.40 & 69.80 & \underline{41.00} & 32.85 & 43.35 & \textbf{99.96} & 21.80 & 70.13 & 49.00 & 37.60 & 62.73 \\
\bottomrule
\end{tabular}
}
\end{table}

\begin{table}[!ht]
\centering
\caption{RULER per-task at 8k context. Rows grouped by compression ratio (16x, 8x, 4x).}
\label{app-tab:ruler-8k}
\scriptsize
\setlength{\tabcolsep}{3pt}
\resizebox{\linewidth}{!}{%
\begin{tabular}{lrrrrrrrrrrrrr|r}
\toprule
Model & ns1 & ns2 & ns3 & nm1 & nm2 & nm3 & nmv & nmq & vt & cwe & fwe & qa1 & qa2 & AVG \\
\midrule
Qwen 4B Instruct & 100.00 & 100.00 & 99.80 & 99.40 & 96.40 & 99.20 & 98.05 & 100.00 & 99.52 & 94.98 & 87.33 & 81.20 & 58.80 & 93.44 \\
\midrule
\multicolumn{15}{c}{\textit{16x compression}} \\
\midrule
LCLM 0.6b-4b (Mean) & 96.00 & 90.40 & \underline{61.80} & \textbf{84.20} & 74.20 & \underline{44.20} & \underline{71.55} & \textbf{77.70} & 77.28 & 52.38 & \textbf{93.33} & 64.40 & \underline{49.80} & \textbf{72.10} \\
ExpAttn & \textbf{100.00} & 76.60 & 1.00 & \underline{81.20} & 2.40 & 0.00 & 69.30 & 55.85 & 99.08 & 10.52 & 60.73 & 30.40 & 23.00 & 46.93 \\
SnapKV & 8.60 & 1.20 & 2.40 & 10.00 & 0.80 & 0.00 & 9.80 & 9.80 & 5.04 & 24.78 & \underline{78.60} & 15.40 & 17.00 & 14.11 \\
SnapKV-QA & 88.20 & 62.00 & 2.40 & 14.60 & 9.00 & 7.60 & 10.30 & 44.25 & 52.68 & 21.16 & 68.13 & \textbf{74.20} & \textbf{53.60} & 39.09 \\
KVzip & \textbf{100.00} & \textbf{98.00} & \textbf{95.80} & 64.80 & 56.40 & \textbf{45.20} & 31.25 & 51.05 & \textbf{100.00} & 8.16 & 73.80 & 43.20 & 34.80 & 61.73 \\
\bottomrule
\end{tabular}
}
\end{table}

\begin{table}[!ht]
\centering
\caption{RULER per-task at 16k context. Rows grouped by compression ratio (16x, 8x, 4x).}
\label{app-tab:ruler-16k}
\scriptsize
\setlength{\tabcolsep}{3pt}
\resizebox{\linewidth}{!}{%
\begin{tabular}{lrrrrrrrrrrrrr|r}
\toprule
Model & ns1 & ns2 & ns3 & nm1 & nm2 & nm3 & nmv & nmq & vt & cwe & fwe & qa1 & qa2 & AVG \\
\midrule
Qwen 4B Instruct & 100.00 & 100.00 & 99.80 & 99.20 & 94.80 & 99.80 & 98.55 & 99.95 & 99.84 & 87.90 & 98.80 & 78.20 & 59.00 & 93.53 \\
\midrule
\multicolumn{15}{c}{\textit{16x compression}} \\
\midrule
LCLM 0.6b-4b (Mean) & 97.00 & 90.00 & \underline{59.40} & 78.20 & \textbf{65.20} & \underline{34.40} & \underline{71.80} & 70.30 & 72.28 & 21.72 & \textbf{99.20} & 62.60 & \underline{49.40} & \textbf{67.04} \\
ExpAttn & \textbf{100.00} & 83.00 & 1.60 & \underline{86.60} & 1.60 & 0.00 & \textbf{78.30} & \underline{76.35} & 99.32 & 8.52 & 71.07 & 28.40 & 24.00 & 50.67 \\
SnapKV & 41.80 & 4.20 & 2.40 & 11.20 & 2.00 & 0.20 & 10.00 & 10.85 & 17.60 & 16.64 & 92.20 & 12.40 & 16.00 & 18.27 \\
SnapKV-QA & \underline{99.80} & \textbf{99.40} & 2.40 & \textbf{88.40} & 11.60 & 27.00 & 30.75 & \textbf{99.15} & 93.68 & 22.22 & \underline{98.33} & \textbf{74.80} & \textbf{58.00} & \underline{61.96} \\
KVzip & \textbf{100.00} & \underline{92.20} & \textbf{91.60} & 55.20 & 37.60 & 18.40 & 35.70 & 53.10 & \textbf{99.96} & 3.18 & 88.87 & 41.20 & 36.60 & 57.97 \\
\bottomrule
\end{tabular}
}
\end{table}

\begin{table}[!ht]
\centering
\caption{LongBench English (en16) per-subtask. Rows grouped by compression ratio (16x, 8x, 4x).}
\label{app-tab:longbench-en}
\scriptsize
\setlength{\tabcolsep}{3pt}
\resizebox{\linewidth}{!}{%
\begin{tabular}{lrrrrrrrrrrrrrrrr|r}
\toprule
Model & narr & qasp & mfq & hpot & 2wiki & musq & govr & qmsm & mnews & trec & triv & samsm & pcnt & pr & lcc & rbp & AVG \\
\midrule
Qwen 4B Instruct & 30.68 & 44.05 & 47.39 & 50.55 & 36.25 & 16.39 & 30.56 & 22.53 & 23.63 & 76.50 & 85.09 & 40.06 & 3.44 & 94.98 & 62.43 & 60.32 & 45.30 \\
\midrule
\multicolumn{18}{c}{\textit{16x compression}} \\
\midrule
LCLM 0.6b-4b (Mean) & 26.12 & 40.43 & 42.38 & 46.55 & \textbf{37.30} & \underline{24.24} & 28.96 & 23.18 & 23.45 & 54.00 & \textbf{86.58} & 33.80 & 2.66 & \underline{67.25} & 37.76 & 37.54 & 38.26 \\
ExpAttn & 17.67 & 23.16 & 28.88 & 33.76 & 23.45 & 8.78 & 23.06 & 19.86 & 19.54 & 36.00 & 81.02 & 31.80 & \underline{7.36} & 30.32 & 28.24 & \textbf{58.14} & 29.44 \\
SnapKV & 12.97 & 13.68 & 22.14 & 20.93 & 16.13 & 5.61 & 21.61 & 17.66 & 14.70 & 32.75 & 85.87 & 36.57 & 3.25 & 17.22 & \textbf{53.42} & \underline{56.78} & 26.96 \\
SnapKV-QA & \textbf{28.36} & 28.81 & 40.37 & \textbf{53.20} & 29.33 & 17.33 & 21.57 & 22.78 & 14.06 & 35.00 & 83.92 & 37.66 & 2.37 & \textbf{92.50} & \textbf{53.42} & 55.60 & 38.52 \\
KVzip & 16.08 & 16.38 & 32.53 & 10.30 & 8.68 & 4.93 & 24.43 & 21.38 & 17.49 & 43.00 & 78.22 & 31.34 & 6.50 & 14.00 & 31.96 & 56.53 & 25.86 \\
\bottomrule
\end{tabular}
}
\end{table}

\begin{table}[!ht]
\centering
\caption{LongBench Chinese (cn5) per-subtask. Rows grouped by compression ratio (16x, 8x, 4x).}
\label{app-tab:longbench-cn}
\small
\setlength{\tabcolsep}{6pt}
\begin{tabular}{lrrrrr|r}
\toprule
Model & mfq & dure & vcsum & lsht & pr & AVG \\
\midrule
Qwen 4B Instruct & 61.10 & 25.32 & 11.19 & 45.75 & 89.33 & 46.54 \\
\midrule
\multicolumn{7}{c}{\textit{16x compression}} \\
\midrule
LCLM 0.6b-4b (Mean) & 39.94 & 27.00 & 10.91 & 23.98 & \underline{55.09} & 31.38 \\
ExpAttn & 25.52 & 16.37 & 11.22 & 17.75 & 31.67 & 20.51 \\
SnapKV & 20.36 & 16.77 & 10.44 & 35.00 & 8.93 & 18.30 \\
SnapKV-QA & \textbf{48.61} & 21.46 & 10.44 & \textbf{41.50} & \textbf{89.28} & \textbf{42.26} \\
KVzip & 25.95 & 16.30 & 12.50 & 23.75 & 7.12 & 17.12 \\
\bottomrule
\end{tabular}
\end{table}

\FloatBarrier

\FloatBarrier
\subsection{LR Sweep}
\label{app-subsec:lr-sweep}
The tables in this subsection and in \Cref{app-subsec:arch-sweep} report the sweep and ablation models of \Cref{sec:arch,app-sec:training-recipe-sweeps}, which were trained with an earlier version of our SFT data, so their scores differ from those in \Cref{app-tab:summary}.
\subsubsection{Continual Pre-training LR sweep}

\begin{table}[!ht]
\centering
\caption{Continual Pre-training LR sweep summary (16 $\times$ ratio compression, bidirectional mask, mlp adapter, $O = 0$).}
\label{tab:continual-lr-summary}
\small
\setlength{\tabcolsep}{6pt}
\begin{adjustbox}{max width=\linewidth}
\begin{tabular}{lrrrrrrr}
\toprule
Model & RULER 4k & RULER 8k & RULER 16k & LB en16 & LB cn5 & LongHealth5 & GSM8K \\
\midrule
$1.0 \times 10^{-6}$ & 62.33 & 58.75 & 56.65 & \underline{33.29} & \underline{22.88} & 60.25 & 66.94 \\
$3.0 \times 10^{-6}$ & \underline{70.73} & \underline{65.41} & \underline{63.55} & 32.89 & 22.09 & \underline{62.50} & \underline{72.25} \\
$1.0 \times 10^{-5}$ & \textbf{71.62} & \textbf{66.83} & \textbf{63.63} & \textbf{33.52} & \textbf{23.94} & \textbf{62.75} & \textbf{74.53} \\
\bottomrule
\end{tabular}
\end{adjustbox}
\end{table}

\begin{table}[!ht]
\centering
\caption{Continual Pretraining LR sweep -- RULER per-task}
\label{tab:continual-lr-ruler}
\scriptsize
\setlength{\tabcolsep}{3pt}
\resizebox{\linewidth}{!}{%
\begin{tabular}{lrrrrrrrrrrrrr|r}
\toprule
Model & ns1 & ns2 & ns3 & nm1 & nm2 & nm3 & nmv & nmq & vt & cwe & fwe & qa1 & qa2 & AVG \\
\midrule
\multicolumn{15}{c}{\textit{Context length 4096}} \\
\midrule
$1.0 \times 10^{-6}$ & 97.40 & 66.40 & 31.40 & \underline{62.00} & 47.20 & 17.40 & 71.70 & 62.10 & 84.88 & \textbf{69.16} & 88.00 & 62.80 & 49.80 & 62.33 \\
$3.0 \times 10^{-6}$ & \textbf{98.80} & \textbf{90.60} & \textbf{42.60} & \textbf{81.40} & \underline{54.40} & \underline{21.80} & \underline{84.50} & \underline{81.15} & \textbf{91.48} & 65.72 & \underline{88.07} & \underline{65.20} & \textbf{53.80} & \underline{70.73} \\
$1.0 \times 10^{-5}$ & \underline{97.80} & \underline{83.00} & \underline{39.40} & \textbf{81.40} & \textbf{64.60} & \textbf{29.80} & \textbf{85.55} & \textbf{81.30} & \underline{90.96} & \underline{68.24} & \textbf{91.07} & \textbf{66.20} & \underline{51.80} & \textbf{71.62} \\
\midrule
\multicolumn{15}{c}{\textit{Context length 8192}} \\
\midrule
$1.0 \times 10^{-6}$ & 96.20 & 77.80 & 28.00 & 60.80 & 40.00 & 12.40 & 60.40 & 57.95 & 84.84 & \textbf{38.70} & \textbf{96.40} & \textbf{62.20} & 48.00 & 58.75 \\
$3.0 \times 10^{-6}$ & \underline{99.00} & \underline{88.00} & \underline{40.80} & \textbf{75.60} & \underline{44.20} & \underline{14.40} & \textbf{78.55} & \textbf{76.15} & \textbf{91.40} & 34.18 & \underline{95.47} & 61.60 & \textbf{51.00} & \underline{65.41} \\
\bottomrule
\end{tabular}
}
\end{table}

\begin{table}[!ht]
\centering
\caption{Continual Pretraining LR sweep -- LongBench English (en16) per-subtask.}
\label{tab:continual-lr-longbench-en}
\scriptsize
\setlength{\tabcolsep}{3pt}
\resizebox{\linewidth}{!}{%
\begin{tabular}{lrrrrrrrrrrrrrrrr|r}
\toprule
Model & narr & qasp & mfq & hpot & 2wiki & musq & govr & qmsm & mnews & trec & triv & samsm & pcnt & pr & lcc & rbp & AVG \\
\midrule
$1.0 \times 10^{-6}$ & 22.98 & 37.30 & \textbf{42.23} & \underline{37.19} & \textbf{31.51} & \textbf{14.42} & 28.61 & 22.15 & 22.72 & \textbf{31.00} & 85.35 & \textbf{33.31} & \underline{1.50} & \textbf{50.00} & 35.85 & 36.53 & \underline{33.29} \\
$3.0 \times 10^{-6}$ & \textbf{23.67} & \underline{38.81} & 41.70 & \textbf{38.09} & 29.49 & \underline{13.43} & \textbf{29.25} & \textbf{22.98} & \textbf{22.98} & \underline{30.00} & \underline{86.44} & \underline{32.39} & 0.50 & 41.50 & \underline{36.19} & \underline{38.82} & 32.89 \\
$1.0 \times 10^{-5}$ & \underline{23.32} & \textbf{40.05} & \underline{42.18} & 35.36 & \underline{31.18} & 12.64 & \underline{29.12} & \underline{22.95} & \underline{22.85} & 29.50 & \textbf{88.68} & 32.28 & \textbf{4.29} & \underline{44.00} & \textbf{38.07} & \textbf{39.92} & \textbf{33.52} \\
\bottomrule
\end{tabular}
}
\end{table}

\begin{table}[!ht]
\centering
\caption{Continual Pretraining LR sweep -- LongBench Chinese (cn5) per-subtask.}
\label{tab:continual-lr-longbench-cn}
\small
\setlength{\tabcolsep}{6pt}
\begin{tabular}{lrrrrr|r}
\toprule
Model & mfq & dure & vcsum & lsht & pr & AVG \\
\midrule
$1.0 \times 10^{-6}$ & \underline{38.79} & 24.13 & \textbf{11.00} & \textbf{21.00} & \underline{19.50} & \underline{22.88} \\
$3.0 \times 10^{-6}$ & 36.39 & \underline{25.97} & 10.85 & 17.75 & \underline{19.50} & 22.09 \\
$1.0 \times 10^{-5}$ & \textbf{39.65} & \textbf{27.07} & \underline{10.99} & \underline{19.50} & \textbf{22.50} & \textbf{23.94} \\
\bottomrule
\end{tabular}
\end{table}

\FloatBarrier

\subsubsection{Post-training LR sweep}

\begin{table}[!ht]
\centering
\caption{Post-training LR sweep summary (16 $\times$ ratio compression, causal mask, mlp adapter, continual pre-training LR $= 1.0 \times 10^{-5}$).}
\label{tab:posttrain-lr-summary}
\small
\setlength{\tabcolsep}{6pt}
\begin{adjustbox}{max width=\linewidth}
\begin{tabular}{lrrrrrrr}
\toprule
Model & RULER 4k & RULER 8k & RULER 16k & LB en16 & LB cn5 & LongHealth5 & GSM8K \\
\midrule
$2.0 \times 10^{-5}$ & 72.99 & 69.08 & 63.89 & 37.98 & \underline{33.08} & 66.75 & 80.89 \\
$3.0 \times 10^{-5}$ & \underline{75.06} & \underline{70.23} & \underline{65.91} & \textbf{39.08} & 31.74 & \underline{67.50} & \textbf{81.05} \\
$5.0 \times 10^{-5}$ & \textbf{75.17} & \textbf{71.06} & \textbf{66.22} & \underline{38.72} & \textbf{34.16} & \textbf{69.00} & \underline{80.97} \\
\bottomrule
\end{tabular}
\end{adjustbox}
\end{table}

\begin{table}[!ht]
\centering
\caption{Post-training LR sweep -- RULER per-task}
\label{tab:posttrain-lr-ruler}
\scriptsize
\setlength{\tabcolsep}{3pt}
\resizebox{\linewidth}{!}{%
\begin{tabular}{lrrrrrrrrrrrrr|r}
\toprule
Model & ns1 & ns2 & ns3 & nm1 & nm2 & nm3 & nmv & nmq & vt & cwe & fwe & qa1 & qa2 & AVG \\
\midrule
\multicolumn{15}{c}{\textit{Context length 4096}} \\
\midrule
$2.0 \times 10^{-5}$ & 94.60 & 89.20 & 40.60 & 82.40 & 80.20 & 46.00 & \underline{72.20} & \underline{75.55} & \underline{83.56} & 78.94 & \underline{89.40} & \textbf{66.40} & \textbf{49.80} & 72.99 \\
$3.0 \times 10^{-5}$ & \textbf{96.40} & \underline{89.40} & \underline{54.60} & \textbf{85.00} & \underline{80.80} & \underline{48.60} & \textbf{72.45} & \textbf{80.20} & \textbf{83.92} & \textbf{80.54} & 88.27 & \underline{66.00} & \underline{49.60} & \underline{75.06} \\
$5.0 \times 10^{-5}$ & \underline{96.00} & \textbf{90.00} & \textbf{58.20} & \underline{85.00} & \textbf{87.00} & \textbf{50.60} & 71.10 & 74.65 & 82.12 & \underline{79.04} & \textbf{91.93} & 64.20 & 47.40 & \textbf{75.17} \\
\midrule
\multicolumn{15}{c}{\textit{Context length 8192}} \\
\midrule
$2.0 \times 10^{-5}$ & 93.60 & 86.20 & 51.20 & 80.20 & \underline{67.00} & 32.20 & 68.00 & \textbf{77.70} & \textbf{80.76} & 54.82 & \textbf{95.73} & \textbf{63.40} & \textbf{47.20} & 69.08 \\
$3.0 \times 10^{-5}$ & \textbf{96.00} & \textbf{90.20} & \underline{55.20} & \textbf{83.40} & 65.40 & \underline{36.60} & \textbf{70.25} & \underline{77.65} & \underline{80.52} & \underline{55.44} & \underline{95.67} & \underline{61.80} & 44.80 & \underline{70.23} \\
\bottomrule
\end{tabular}
}
\end{table}

\begin{table}[!ht]
\centering
\caption{Post-training LR sweep -- LongBench English (en16) per-subtask.}
\label{tab:posttrain-lr-longbench-en}
\scriptsize
\setlength{\tabcolsep}{3pt}
\resizebox{\linewidth}{!}{%
\begin{tabular}{lrrrrrrrrrrrrrrrr|r}
\toprule
Model & narr & qasp & mfq & hpot & 2wiki & musq & govr & qmsm & mnews & trec & triv & samsm & pcnt & pr & lcc & rbp & AVG \\
\midrule
$2.0 \times 10^{-5}$ & \underline{23.54} & \underline{40.34} & \textbf{48.52} & 46.17 & 36.93 & \underline{21.76} & \textbf{28.59} & \underline{22.21} & 22.76 & \textbf{49.00} & \underline{87.88} & \underline{32.31} & \underline{3.50} & 74.00 & 34.65 & 35.54 & 37.98 \\
$3.0 \times 10^{-5}$ & \textbf{25.57} & 39.80 & \underline{47.98} & \textbf{47.96} & \textbf{38.44} & 21.17 & \underline{28.56} & \textbf{22.42} & \textbf{23.48} & \underline{45.00} & \textbf{89.43} & \textbf{32.50} & 3.00 & \underline{81.50} & \textbf{37.47} & \textbf{41.05} & \textbf{39.08} \\
$5.0 \times 10^{-5}$ & 23.12 & \textbf{42.47} & 47.64 & \underline{46.80} & \underline{37.13} & \textbf{24.03} & 28.27 & 22.04 & \underline{22.80} & 44.00 & 87.10 & 32.15 & \textbf{4.50} & \textbf{85.50} & \underline{35.45} & \underline{36.57} & \underline{38.72} \\
\bottomrule
\end{tabular}
}
\end{table}

\begin{table}[!ht]
\centering
\caption{Post-training LR sweep -- LongBench Chinese (cn5) per-subtask.}
\label{tab:posttrain-lr-longbench-cn}
\small
\setlength{\tabcolsep}{6pt}
\begin{tabular}{lrrrrr|r}
\toprule
Model & mfq & dure & vcsum & lsht & pr & AVG \\
\midrule
$2.0 \times 10^{-5}$ & \underline{47.53} & \textbf{24.26} & \textbf{10.77} & \textbf{20.83} & \underline{62.00} & \underline{33.08} \\
$3.0 \times 10^{-5}$ & 46.83 & \underline{23.63} & \underline{10.55} & 18.67 & 59.00 & 31.74 \\
$5.0 \times 10^{-5}$ & \textbf{47.61} & 21.95 & 10.38 & \underline{18.85} & \textbf{72.00} & \textbf{34.16} \\
\bottomrule
\end{tabular}
\end{table}

\FloatBarrier   
\clearpage
\FloatBarrier
\subsection{Architecture Comparison}
\label{app-subsec:arch-sweep}

\subsubsection{Pooling Operator}
\label{app-subsubsec:pooling-operator}

\begin{table}[!ht]
\centering
\caption{Pooling operator: Mean vs Concat ($W = 1024$, causal, continual pre-training LR = $1.0 \times 10^{-5}$). These models were trained with an earlier version of our SFT data, so their scores differ from those in \Cref{app-tab:summary}.}
\label{app-tab:pool-summary}
\small
\setlength{\tabcolsep}{6pt}
\begin{tabular}{lrrrrrrr}
\toprule
Model & RULER 4k & RULER 8k & RULER 16k & LB en16 & LB cn5 & LongHealth5 & GSM8K \\
\midrule
\multicolumn{8}{c}{\textit{4x compression}} \\
\midrule
Mean   & 91.76 & 91.03 & 89.96 & \textbf{46.04} & 40.92 & 76.25 & \textbf{91.05} \\
Concat & \textbf{92.30} & \textbf{91.50} & \textbf{90.50} & 45.39 & \textbf{41.80} & \textbf{82.50} & 89.90 \\
\midrule
\multicolumn{8}{c}{\textit{8x compression}} \\
\midrule
Mean   & 85.42 & 84.48 & 82.47 & 42.23 & 34.61 & 71.75 & 87.26 \\
Concat & \textbf{87.20} & \textbf{86.80} & \textbf{84.40} & \textbf{42.32} & \textbf{36.46} & \textbf{79.50} & \textbf{87.40} \\
\midrule
\multicolumn{8}{c}{\textit{16x compression}} \\
\midrule
Mean   & \textbf{75.06} & 70.23 & \textbf{65.91} & \textbf{39.08} & \textbf{31.74} & 67.50 & \textbf{81.05} \\
Concat & 74.50 & \textbf{70.30} & 64.90 & 36.24 & 25.88 & \textbf{76.30} & 78.90 \\
\bottomrule
\end{tabular}
\end{table}

\begin{table}[!ht]
\centering
\caption{Pooling operator: RULER per-task at 4k context.}
\label{app-tab:pool-ruler-4k}
\scriptsize
\setlength{\tabcolsep}{3pt}
\resizebox{\linewidth}{!}{%
\begin{tabular}{lrrrrrrrrrrrrr|r}
\toprule
Model & ns1 & ns2 & ns3 & nm1 & nm2 & nm3 & nmv & nmq & vt & cwe & fwe & qa1 & qa2 & AVG \\
\midrule
\multicolumn{15}{c}{\textit{4x compression}} \\
\midrule
Mean   & \textbf{99.40} & 99.60 & 93.40 & 99.00 & 99.40 & 93.80 & 98.25 & \textbf{99.20} & \textbf{94.64} & 91.42 & 86.73 & \textbf{79.20} & 58.80 & 91.76 \\
Concat & 99.20 & \textbf{99.80} & \textbf{94.40} & \textbf{99.40} & \textbf{99.80} & \textbf{97.60} & \textbf{98.80} & \textbf{99.20} & 93.90 & \textbf{92.80} & \textbf{87.60} & 77.20 & \textbf{59.80} & \textbf{92.27} \\
\midrule
\multicolumn{15}{c}{\textit{8x compression}} \\
\midrule
Mean   & \textbf{98.60} & 94.20 & 64.20 & 93.00 & \textbf{98.20} & 82.20 & 92.30 & 90.70 & 87.40 & 90.80 & \textbf{90.60} & 73.20 & \textbf{55.00} & 85.42 \\
Concat & 98.40 & \textbf{97.00} & \textbf{76.60} & \textbf{94.60} & 95.60 & \textbf{86.00} & \textbf{93.80} & \textbf{92.30} & \textbf{90.70} & \textbf{92.80} & 88.40 & \textbf{74.80} & 53.20 & \textbf{87.25} \\
\midrule
\multicolumn{15}{c}{\textit{16x compression}} \\
\midrule
Mean   & \textbf{96.40} & 89.40 & 54.60 & \textbf{85.00} & 80.80 & 48.60 & \textbf{72.45} & \textbf{80.20} & \textbf{83.92} & \textbf{80.54} & 88.27 & \textbf{66.00} & \textbf{49.60} & \textbf{75.06} \\
Concat & \textbf{96.40} & \textbf{91.80} & \textbf{62.40} & 83.80 & \textbf{84.20} & \textbf{54.00} & 59.50 & 77.50 & 79.50 & 74.30 & \textbf{90.10} & 65.40 & 49.00 & 74.45 \\
\bottomrule
\end{tabular}
}
\end{table}

\begin{table}[!ht]
\centering
\caption{Pooling operator: RULER per-task at 8k context.}
\label{app-tab:pool-ruler-8k}
\scriptsize
\setlength{\tabcolsep}{3pt}
\resizebox{\linewidth}{!}{%
\begin{tabular}{lrrrrrrrrrrrrr|r}
\toprule
Model & ns1 & ns2 & ns3 & nm1 & nm2 & nm3 & nmv & nmq & vt & cwe & fwe & qa1 & qa2 & AVG \\
\midrule
\multicolumn{15}{c}{\textit{4x compression}} \\
\midrule
Mean   & \textbf{99.40} & \textbf{99.80} & 95.60 & 98.60 & 98.40 & 91.60 & \textbf{98.80} & 98.95 & 94.64 & 82.04 & \textbf{90.40} & \textbf{76.20} & \textbf{59.00} & 91.03 \\
Concat & 99.20 & 99.60 & \textbf{96.60} & \textbf{99.20} & \textbf{99.60} & \textbf{96.00} & 98.60 & \textbf{99.20} & \textbf{94.90} & \textbf{84.20} & 89.80 & 75.20 & 57.40 & \textbf{91.50} \\
\midrule
\multicolumn{15}{c}{\textit{8x compression}} \\
\midrule
Mean   & \textbf{98.80} & \textbf{96.40} & 71.40 & 90.80 & \textbf{96.80} & 74.60 & 90.40 & 90.95 & 89.40 & 79.82 & \textbf{95.67} & 70.60 & 52.60 & 84.48 \\
Concat & 98.60 & 95.80 & \textbf{82.00} & \textbf{92.00} & 95.40 & \textbf{81.00} & \textbf{93.10} & \textbf{92.30} & \textbf{90.00} & \textbf{88.00} & 94.60 & \textbf{71.60} & \textbf{53.60} & \textbf{86.77} \\
\midrule
\multicolumn{15}{c}{\textit{16x compression}} \\
\midrule
Mean   & 96.00 & 90.20 & 55.20 & \textbf{83.40} & 65.40 & 36.60 & \textbf{70.25} & 77.65 & \textbf{80.52} & \textbf{55.44} & \textbf{95.67} & 61.80 & \textbf{44.80} & 70.23 \\
Concat & \textbf{97.60} & \textbf{93.20} & \textbf{61.80} & 82.80 & \textbf{72.80} & \textbf{41.20} & 62.10 & \textbf{78.00} & 68.10 & 54.60 & 94.70 & \textbf{62.60} & \textbf{44.80} & \textbf{70.33} \\
\bottomrule
\end{tabular}
}
\end{table}

\begin{table}[!ht]
\centering
\caption{Pooling operator: RULER per-task at 16k context.}
\label{app-tab:pool-ruler-16k}
\scriptsize
\setlength{\tabcolsep}{3pt}
\resizebox{\linewidth}{!}{%
\begin{tabular}{lrrrrrrrrrrrrr|r}
\toprule
Model & ns1 & ns2 & ns3 & nm1 & nm2 & nm3 & nmv & nmq & vt & cwe & fwe & qa1 & qa2 & AVG \\
\midrule
\multicolumn{15}{c}{\textit{4x compression}} \\
\midrule
Mean   & \textbf{99.60} & \textbf{99.60} & 94.80 & 99.20 & 98.20 & 90.40 & 98.15 & \textbf{98.80} & 91.40 & 67.34 & \textbf{99.53} & \textbf{75.60} & 56.80 & 89.96 \\
Concat & 99.20 & 99.20 & \textbf{95.60} & \textbf{99.60} & \textbf{99.00} & \textbf{92.40} & \textbf{98.30} & 98.70 & \textbf{92.80} & \textbf{69.90} & 98.50 & \textbf{75.60} & \textbf{57.60} & \textbf{90.49} \\
\midrule
\multicolumn{15}{c}{\textit{8x compression}} \\
\midrule
Mean   & \textbf{99.20} & 95.00 & 72.60 & \textbf{88.80} & \textbf{94.20} & 69.20 & 88.10 & 90.10 & 89.04 & 67.34 & 97.73 & 68.40 & \textbf{52.40} & 82.47 \\
Concat & 98.80 & \textbf{95.80} & \textbf{82.00} & 88.20 & 90.20 & \textbf{73.20} & \textbf{91.00} & \textbf{90.90} & \textbf{90.00} & \textbf{77.50} & \textbf{98.50} & \textbf{69.40} & 52.20 & \textbf{84.44} \\
\midrule
\multicolumn{15}{c}{\textit{16x compression}} \\
\midrule
Mean   & 95.80 & 88.00 & 57.40 & \textbf{75.60} & 62.20 & \textbf{30.00} & \textbf{70.70} & \textbf{73.05} & \textbf{77.32} & 24.78 & 98.73 & 59.40 & \textbf{43.80} & \textbf{65.91} \\
Concat & \textbf{96.80} & \textbf{90.60} & \textbf{65.00} & \textbf{75.60} & \textbf{65.00} & 28.80 & 63.10 & 70.30 & 58.20 & \textbf{26.60} & \textbf{98.90} & \textbf{60.80} & 43.60 & 64.87 \\
\bottomrule
\end{tabular}
}
\end{table}

\begin{table}[!ht]
\centering
\caption{Pooling operator: LongBench English (en16) per-subtask.}
\label{app-tab:pool-longbench-en}
\scriptsize
\setlength{\tabcolsep}{3pt}
\resizebox{\linewidth}{!}{%
\begin{tabular}{lrrrrrrrrrrrrrrrr|r}
\toprule
Model & narr & qasp & mfq & hpot & 2wiki & musq & govr & qmsm & mnews & trec & triv & samsm & pcnt & pr & lcc & rbp & AVG \\
\midrule
\multicolumn{18}{c}{\textit{4x compression}} \\
\midrule
Mean   & \textbf{30.74} & \textbf{46.76} & \textbf{49.76} & 52.90 & \textbf{42.75} & 28.61 & \textbf{30.12} & \textbf{22.89} & 23.83 & \textbf{68.50} & 87.90 & 34.10 & 6.67 & \textbf{100.00} & \textbf{52.01} & \textbf{59.17} & \textbf{46.04} \\
Concat & 28.30 & 46.00 & 48.70 & \textbf{56.80} & 42.40 & \textbf{29.70} & 29.80 & 22.80 & \textbf{24.00} & 66.00 & \textbf{88.50} & \textbf{34.20} & \textbf{7.50} & 98.00 & 49.20 & 54.30 & 45.39 \\
\midrule
\multicolumn{18}{c}{\textit{8x compression}} \\
\midrule
Mean   & 28.60 & \textbf{43.92} & \textbf{51.54} & \textbf{54.02} & \textbf{45.36} & 28.05 & 29.05 & 22.85 & 23.56 & 62.00 & 86.39 & \textbf{33.17} & \textbf{5.45} & \textbf{71.50} & 43.57 & 46.63 & 42.23 \\
Concat & \textbf{30.30} & 42.70 & 50.70 & 50.80 & 44.30 & \textbf{28.20} & \textbf{29.20} & \textbf{22.90} & \textbf{23.90} & \textbf{62.50} & \textbf{89.00} & 33.10 & 4.60 & 70.50 & \textbf{45.20} & \textbf{49.20} & \textbf{42.32} \\
\midrule
\multicolumn{18}{c}{\textit{16x compression}} \\
\midrule
Mean   & \textbf{25.57} & \textbf{39.80} & 47.98 & 47.96 & \textbf{38.44} & 21.17 & 28.56 & \textbf{22.42} & \textbf{23.48} & 45.00 & \textbf{89.43} & 32.50 & 3.00 & \textbf{81.50} & 37.47 & \textbf{41.05} & \textbf{39.08} \\
Concat & 23.30 & 36.80 & \textbf{48.00} & \textbf{49.00} & 38.30 & \textbf{22.60} & \textbf{28.60} & 22.40 & 23.10 & \textbf{46.50} & 88.20 & \textbf{32.70} & \textbf{3.50} & 37.00 & \textbf{39.90} & 39.90 & 36.24 \\
\bottomrule
\end{tabular}
}
\end{table}

\begin{table}[!ht]
\centering
\caption{Pooling operator: LongBench Chinese (cn5) per-subtask.}
\label{app-tab:pool-longbench-cn}
\scriptsize
\setlength{\tabcolsep}{3pt}
\begin{tabular}{lrrrrr|r}
\toprule
Model & mfq & dure & vcsum & lsht & pr & AVG \\
\midrule
\multicolumn{7}{c}{\textit{4x compression}} \\
\midrule
Mean   & 45.52 & 26.07 & \textbf{11.00} & 25.00 & \textbf{97.00} & 40.92 \\
Concat & \textbf{46.20} & \textbf{26.80} & \textbf{11.00} & \textbf{30.00} & 95.00 & \textbf{41.80} \\
\midrule
\multicolumn{7}{c}{\textit{8x compression}} \\
\midrule
Mean   & 51.27 & \textbf{24.83} & 10.79 & 25.65 & 60.50 & 34.61 \\
Concat & \textbf{53.10} & 24.10 & \textbf{10.90} & \textbf{26.70} & \textbf{67.50} & \textbf{36.46} \\
\midrule
\multicolumn{7}{c}{\textit{16x compression}} \\
\midrule
Mean   & 46.83 & \textbf{23.63} & \textbf{10.55} & 18.67 & \textbf{59.00} & \textbf{31.74} \\
Concat & \textbf{51.00} & 23.50 & 10.10 & \textbf{19.30} & 25.50 & 25.88 \\
\bottomrule
\end{tabular}
\end{table}

\FloatBarrier

\subsubsection{Adapter \& Overlap}
\label{app-subsubsec:adapter-overlap}
\begin{table}[!ht]
\centering
\caption{Adapter \& overlap comparison (16 $\times$ compression, mean pooling, $W = 1024$, continual pre-training LR $= 1.0 \times 10^{-5}$).}
\label{tab:adapter-summary}
\small
\setlength{\tabcolsep}{6pt}
\begin{adjustbox}{max width=\linewidth}
\begin{tabular}{lrrrrrrr}
\toprule
Model & RULER 4k & RULER 8k & RULER 16k & LB en16 & LB cn5 & LongHealth5 & GSM8K \\
\midrule
Causal-MLP-O0 & \textbf{75.06} & \textbf{70.23} & \textbf{65.91} & \textbf{39.08} & \textbf{31.74} & \textbf{67.50} & \textbf{81.05} \\
Bidirectional-MLP-O0 & 71.62 & 66.83 & 63.63 & 33.52 & 23.94 & 62.75 & 74.53 \\
Bidirectional-ATTN-MLP-O256 & \underline{71.91} & \underline{68.11} & \underline{64.96} & 36.28 & 26.54 & \underline{64.00} & 73.77 \\
Bidirectional-MLP-O256 & 62.03 & 58.66 & 58.24 & \underline{36.97} & \underline{26.63} & 63.25 & \underline{74.98} \\
\bottomrule
\end{tabular}
\end{adjustbox}
\end{table}

\begin{table}[!ht]
\centering
\caption{Adapter \& overlap comparison -- RULER per-task}
\label{tab:arch-ruler}
\scriptsize
\setlength{\tabcolsep}{3pt}
\resizebox{\linewidth}{!}{%
\begin{tabular}{lrrrrrrrrrrrrr|r}
\toprule
Model & ns1 & ns2 & ns3 & nm1 & nm2 & nm3 & nmv & nmq & vt & cwe & fwe & qa1 & qa2 & AVG \\
\midrule
\multicolumn{15}{c}{\textit{Context length 4096}} \\
\midrule
Causal-MLP-O0 & 96.40 & \textbf{89.40} & \textbf{54.60} & \textbf{85.00} & \textbf{80.80} & \textbf{48.60} & 72.45 & 80.20 & 83.92 & \textbf{80.54} & \underline{88.27} & \underline{66.00} & 49.60 & \textbf{75.06} \\
Bidirectional-MLP-O0 & 97.80 & \underline{83.00} & 39.40 & \underline{81.40} & \underline{64.60} & \underline{29.80} & \underline{85.55} & \underline{81.30} & 90.96 & 68.24 & \textbf{91.07} & \textbf{66.20} & \textbf{51.80} & 71.62 \\
Bidirectional-ATTN-MLP-O256 & \textbf{99.80} & 80.40 & \underline{50.20} & 78.60 & 62.20 & 26.20 & \textbf{85.70} & \textbf{81.35} & \underline{95.00} & \underline{80.00} & 87.00 & 61.80 & 46.60 & \underline{71.91} \\
Bidirectional-MLP-O256 & \underline{99.00} & 60.60 & 21.60 & 61.80 & 55.40 & 17.60 & 69.90 & 61.60 & \textbf{97.12} & 57.66 & 87.27 & \underline{66.00} & \underline{50.80} & 62.03 \\
\midrule
\multicolumn{15}{c}{\textit{Context length 8192}} \\
\midrule
Causal-MLP-O0 & 96.00 & \textbf{90.20} & \underline{55.20} & \textbf{83.40} & \textbf{65.40} & \textbf{36.60} & 70.25 & \textbf{77.65} & 80.52 & \textbf{55.44} & \textbf{95.67} & 61.80 & 44.80 & \textbf{70.23} \\
\bottomrule
\end{tabular}
}
\end{table}

\begin{table}[!ht]
\centering
\caption{Adapter \& overlap  comparison -- LongBench English (en16) per-subtask.}
\label{tab:arch-longbench-en}
\scriptsize
\setlength{\tabcolsep}{3pt}
\resizebox{\linewidth}{!}{%
\begin{tabular}{lrrrrrrrrrrrrrrrr|r}
\toprule
Model & narr & qasp & mfq & hpot & 2wiki & musq & govr & qmsm & mnews & trec & triv & samsm & pcnt & pr & lcc & rbp & AVG \\
\midrule
Causal-MLP-O0 & \underline{25.57} & \underline{39.80} & \textbf{47.98} & \underline{47.96} & \textbf{38.44} & \textbf{21.17} & 28.56 & 22.42 & \textbf{23.48} & \textbf{45.00} & \textbf{89.43} & \textbf{32.50} & 3.00 & \textbf{81.50} & 37.47 & 41.05 & \textbf{39.08} \\
Bidirectional-MLP-O0 & 23.32 & \textbf{40.05} & 42.18 & 35.36 & 31.18 & 12.64 & \textbf{29.12} & \textbf{22.95} & 22.85 & 29.50 & \underline{88.68} & \underline{32.28} & 4.29 & 44.00 & 38.07 & 39.92 & 33.52 \\
Bidirectional-ATTN-MLP-O256 & \textbf{27.45} & 38.87 & 42.56 & \textbf{52.31} & \underline{36.98} & \underline{20.22} & 27.98 & 22.44 & \underline{23.01} & 27.50 & 87.23 & 32.16 & \textbf{6.00} & 51.50 & \underline{41.22} & \underline{43.00} & 36.28 \\
Bidirectional-MLP-O256 & 23.20 & 39.67 & \underline{42.77} & 46.56 & 31.94 & 16.41 & \underline{28.64} & \underline{22.50} & 22.65 & \underline{40.00} & 86.67 & 31.80 & \underline{4.50} & \underline{67.50} & \textbf{41.76} & \textbf{44.89} & \underline{36.97} \\
\bottomrule
\end{tabular}
}
\end{table}

\begin{table}[!ht]
\centering
\caption{Adapter \& overlap comparison -- LongBench Chinese (cn5) per-subtask.}
\label{tab:arch-longbench-cn}
\scriptsize
\setlength{\tabcolsep}{3pt}
\begin{tabular}{lrrrrr|r}
\toprule
Model & mfq & dure & vcsum & lsht & pr & AVG \\
\midrule
Causal-MLP-O0 & \textbf{46.83} & 23.63 & 10.55 & 18.67 & \textbf{59.00} & \textbf{31.74} \\
Bidirectional-MLP-O0 & 39.65 & \textbf{27.07} & \underline{10.99} & \underline{19.50} & 22.50 & 23.94 \\
Bidirectional-ATTN-MLP-O256 & \underline{44.35} & 23.26 & 10.75 & 18.83 & 35.50 & 26.54 \\
Bidirectional-MLP-O256 & 38.06 & \underline{24.02} & \textbf{11.05} & \textbf{22.50} & \underline{37.50} & \underline{26.63} \\
\bottomrule
\end{tabular}
\end{table}

\FloatBarrier

\FloatBarrier
\subsubsection{Mask-type Comparison}
\begin{table}[!ht]
\centering
\caption{Mask-type ablation  (16 $\times$ compression, mean pooling, W=1024, continual pre-training LR $= 1.0 \times 10^{-5}$).}
\label{tab:abl-mask}
\small
\setlength{\tabcolsep}{6pt}
\begin{adjustbox}{max width=\linewidth}
\begin{tabular}{lrrrrrrr}
\toprule
Mask & RULER 4k & RULER 8k & RULER 16k & LB en16 & LB cn5 & LongHealth5 & GSM8K \\
\midrule
Bidirectional & \underline{71.62} & \underline{66.83} & \underline{63.63} & \underline{33.52} & \underline{23.94} & \underline{62.75} & \underline{74.53} \\
Causal & \textbf{75.06} & \textbf{70.23} & \textbf{65.91} & \textbf{39.08} & \textbf{31.74} & \textbf{67.50} & \textbf{81.05} \\
\bottomrule
\end{tabular}
\end{adjustbox}
\end{table}

\FloatBarrier

\FloatBarrier

\FloatBarrier
\subsubsection{Window-size Comparison}
\begin{table}[!ht]
\centering
\caption{Window-size ablation (16 $\times$ compression, mean pooling, continual pre-training LR $= 1.0 \times 10^{-6}$). W=16 and W=256 use causal attention; W=1024 uses bidirectional attention. We do not run W=1024 with causal attention at LR $= 1.0 \times 10^{-6}$. However, \Cref{tab:abl-mask} shows that causal masking is strictly better than bidirectional masking, and W=1024 (bidirectional) already outperforms W=256 (causal) on RULER tasks while also achieving lower pre-training loss. We therefore conclude that W=1024 is the better setting.}
\label{tab:abl-window}
\small
\setlength{\tabcolsep}{6pt}
\begin{adjustbox}{max width=\linewidth}
\begin{tabular}{lrrrrrrr}
\toprule
Model & RULER 4k & RULER 8k & RULER 16k & LB en16 & LB cn5 & LongHealth5 & GSM8K \\
\midrule
Mean W16  (causal) & 24.85 & 23.87 & 22.54 & 29.62 & 22.62 & 56.50 & 46.55 \\
Mean W256 (causal) & \underline{58.37} & \underline{54.39} & \underline{50.18} & \textbf{34.15} & \textbf{25.97} & \textbf{62.25} & \textbf{76.80} \\
Mean W1024 (bidir) & \textbf{62.33} & \textbf{58.75} & \textbf{56.65} & \underline{33.29} & \underline{22.88} & \underline{60.25} & \underline{66.94} \\
\bottomrule
\end{tabular}
\end{adjustbox}
\end{table}

\FloatBarrier

\FloatBarrier
\subsubsection{Encoder-representation Comparison}
\begin{table}[!ht]
\centering
\caption{Encoder-representation ablation (16 $\times$ compression, EOS pooling, causal, continual pre-training LR $= 1.0 \times 10^{-6}$).}
\label{tab:abl-encoder}
\small
\setlength{\tabcolsep}{6pt}
\begin{adjustbox}{max width=\linewidth}
\begin{tabular}{lrrrrrrr}
\toprule
Encoder init & RULER 4k & RULER 8k & RULER 16k & LB en16 & LB cn5 & LongHealth5 & GSM8K \\
\midrule
LLM (Instruct) & \underline{38.29} & \underline{34.39} & \underline{33.50} & \underline{30.48} & \underline{22.73} & \textbf{58.25} & \underline{54.51} \\
Embed (Qwen3-Emb) & \textbf{42.00} & \textbf{38.76} & \textbf{38.12} & \textbf{31.72} & \textbf{25.32} & \underline{58.00} & \textbf{59.36} \\
\bottomrule
\end{tabular}
\end{adjustbox}
\end{table}

\FloatBarrier
  
\clearpage
\FloatBarrier
\subsection{NIAH Agent Results}

\begin{table}[!ht]
\centering
\caption{RULER NIAH (retrieval) per-task results for our LCLM 0.6B-4B model at 16x compression, with and without the agentic expand-or-answer loop, across context lengths. \textit{$\Delta$ agent} is the cell-wise gain over the non-agentic baseline. Each cell averages 50 questions per task and context length, matching \Cref{fig:agentic_bar_plot}.}
\label{app-tab:niah-agent}
\small
\setlength{\tabcolsep}{6pt}
\begin{adjustbox}{max width=\linewidth}
\begin{tabular}{lrrrrrrrr|r}
\toprule
Model & ns1 & ns2 & ns3 & nm1 & nm2 & nm3 & nmv & nmq & AVG \\
\midrule
\multicolumn{10}{c}{\textit{4k context}} \\
\midrule
LCLM 0.6B-4B & 100.00 & 88.00 & 70.00 & 90.00 & 86.00 & 56.00 & 72.00 & 85.50 & 80.94 \\
LCLM 0.6B-4B agent & \textbf{100.00} & \textbf{100.00} & \textbf{100.00} & \textbf{100.00} & \textbf{100.00} & \textbf{100.00} & \textbf{95.50} & \textbf{100.00} & \textbf{99.44} \\
\textit{$\Delta$ agent} & \textit{+0.00} & \textit{+12.00} & \textit{+30.00} & \textit{+10.00} & \textit{+14.00} & \textit{+44.00} & \textit{+23.50} & \textit{+14.50} & \textit{+18.50} \\
\midrule
\multicolumn{10}{c}{\textit{8k context}} \\
\midrule
LCLM 0.6B-4B & 96.00 & 90.00 & 58.00 & 82.00 & 88.00 & 44.00 & 70.00 & 79.00 & 75.88 \\
LCLM 0.6B-4B agent & \textbf{100.00} & \textbf{100.00} & \textbf{100.00} & \textbf{100.00} & \textbf{98.00} & \textbf{100.00} & \textbf{95.00} & \textbf{99.50} & \textbf{99.06} \\
\textit{$\Delta$ agent} & \textit{+4.00} & \textit{+10.00} & \textit{+42.00} & \textit{+18.00} & \textit{+10.00} & \textit{+56.00} & \textit{+25.00} & \textit{+20.50} & \textit{+23.19} \\
\midrule
\multicolumn{10}{c}{\textit{16k context}} \\
\midrule
LCLM 0.6B-4B & 92.00 & 82.00 & 60.00 & 86.00 & 60.00 & 26.00 & 68.00 & 82.50 & 69.56 \\
LCLM 0.6B-4B agent & \textbf{100.00} & \textbf{100.00} & \textbf{100.00} & \textbf{98.00} & \textbf{86.00} & \textbf{86.00} & \textbf{93.00} & \textbf{97.50} & \textbf{95.06} \\
\textit{$\Delta$ agent} & \textit{+8.00} & \textit{+18.00} & \textit{+40.00} & \textit{+12.00} & \textit{+26.00} & \textit{+60.00} & \textit{+25.00} & \textit{+15.00} & \textit{+25.50} \\
\bottomrule
\end{tabular}
\end{adjustbox}
\end{table}

\end{document}